\documentclass[journal]{IEEEtran}

\usepackage{url}
\usepackage{graphicx}
\usepackage{subfigure}
\usepackage{caption}
\usepackage{cite}
\usepackage[ruled]{algorithm2e}

\usepackage{amsmath}
\usepackage{amssymb}
\usepackage{booktabs}
\usepackage{colortbl}
\usepackage{multirow}
\usepackage{amsthm}

\newtheorem{definition}{Definition}

\begin{document}

\title{Towards Fast and Accurate Federated Learning with non-IID Data for Cloud-Based IoT Applications}

\author{Tian~Liu,~\IEEEmembership{~Student Member,~IEEE}, Jiahao~Ding,~\IEEEmembership{~Student Member,~IEEE}, 
Ting~Wang,~\IEEEmembership{~Member,~IEEE}, 
Miao~Pan,~\IEEEmembership{~Senior Member,~IEEE},
and Mingsong~Chen,~\IEEEmembership{~Senior Member,~IEEE}

\IEEEcompsocitemizethanks{
\IEEEcompsocthanksitem Tian Liu, Ting Wang and Mingsong Chen are with the 
MoE Engineering Research Center of Software/Hardware Co-design Technology and Application,
East China Normal University,
Shanghai, 200062, China (email: liutian2534@qq.com, \{twang, mschen\}@sei.ecnu.edu.cn). 
Tian Liu is also with the Department of Information Science and Engineering,
Zaozhuang University, Zaozhuang 277160, China.
Jiahao Ding and Miao Pan are with the Department of Electrical and Computer
Engineering, University of Houston, Houston, TX 77204 USA (e-mail: \{jding7, mpan2\}@uh.edu).
}}


\maketitle

\begin{abstract}
As a promising method of central model training on decentralized device data while securing user privacy, Federated Learning (FL) is becoming popular in Internet of Things (IoT) design. 
However, when the data collected by IoT devices are highly skewed in a non-independent and identically distributed (non-IID) manner, the accuracy of vanilla FL method cannot be guaranteed. 
Although there exist various solutions that try to address the bottleneck of FL with non-IID data, most of them suffer from extra intolerable communication overhead and low model accuracy.
To enable fast and accurate FL, this paper proposes a novel data-based device grouping approach that can effectively reduce the disadvantages of weight divergence during the training of non-IID data. 
However, since our grouping method is based on the similarity of extracted feature maps from IoT devices, it may incur additional risks of privacy exposure.
To solve this problem, we propose an improved version by exploiting similarity information using the Locality-Sensitive Hashing (LSH) algorithm without exposing extracted feature maps.  
Comprehensive experimental results on well-known benchmarks show that our approach can not only accelerate the convergence rate, but also improve the prediction accuracy for FL with non-IID data.
\end{abstract}


\IEEEpeerreviewmaketitle

\section{Introduction}\label{sec:intro}

\IEEEPARstart{A}{long} with the prosperity of 5G techniques, Artificial Intelligence (AI) and cloud computing,
Federated Learning (FL)~\cite{communication,largescale} is becoming a popular distributed machine learning paradigm for the design of cloud-based Internet of Things (IoT) applications such as commercial surveillance~\cite{deep,object}, industrial control~\cite{enhanced,artificial,6g}, and autonomous driving~\cite{asynchronous}.
Rather than bring data to the centralized model, FL dispatches models to data sources (i.e., IoT devices) to assist the model training.
In FL, the cloud manages a global model for the fusion of local models hosted by IoT devices. 
During the training process, devices only send their gradients to the cloud server for weight aggregation, thus the data privacy protection and low communication overhead can be achieved.

Although FL is promising in enabling collaborative learning among IoT devices, it greatly suffers from the problems caused by data distribution diversity~\cite{reinforcement,flwithnoniid,agnostic}.
When dealing with independent and identically distributed (IID) data, vanilla FL shows excellent performance, since all the stochastic gradients obtained from local data are unbiased estimates of the full gradient~\cite{largescale,flwithnoniid}.
However, when dealing with non-IID, the convergence directions of local models during the training phase are highly inconsistent.
Due to the weight divergence~\cite{flwithnoniid}, the accuracy of vanilla FL may be greatly reduced~\cite{flwithnoniid,agnostic}.
Moreover, non-IID FL requires much more communication rounds than IID FL to achieve the global model convergence, which in turn poses intolerable pressure on the limited network bandwidth of IoT devices~\cite{wireless1,wireless2}.

In order to improve the accuracy of non-IID FL, various model optimization techniques~\cite{feddane,heterogeneous} and model comparison methods \cite{multicenter,hierarchical} have been proposed to mitigate the weight divergence problem.
However, most non-IID FL methods have their own innegligible side-effects, since they will result in significant communication overhead. 
Moreover, the exposure of local models to the cloud for comparison may easily violate the FL privacy requirement.
All these issues hinder the wide deployment of FL techniques on non-IID IoT applications. 
As a result, how to design a fast and accurate non-IID FL without exposing model privacy information is becoming a major bottleneck in IoT design.

To address the above challenge, this paper presents a novel device grouping-based FL method named FLDG that can effectively reduce the disadvantages of weight divergence during the training of non-IID data.
Unlike vanilla FL that randomly selects IoT devices for model gradient aggregation, our proposed non-IID FL method clusters devices based on the similarities of their raw data and selects only one device from a group for the model gradient aggregation. 
In this way, our approach can stabilize the convergence of the global model, since all kinds of model divergences are involved once in each FL training round without unpredictable perturbations caused by random device selection. 
This paper makes the following three major contributions:

\begin{itemize}
    \item We propose an efficient device grouping method based on extracted feature maps, which can stabilize the optimization direction of the global model, thus accelerating the model convergence rate and reducing the communication overhead of non-IID FL training drastically.  
    \item To prevent the privacy exposure caused by attacks such as model inversion attack or membership inference attack, we present an improved version of FLDG named FLDG-L, which encodes extracted feature maps using the Locality-Sensitive Hashing (LSH) algorithm for device comparison. With this method, the chance of the raw data restore becomes much lower, while the comparison accuracy is mostly remained.
    \item We conduct comprehensive experiments on three well-known benchmarks. Experimental results show that our proposed FLDG and FLDG-L approaches can speed up the convergence rate of the training phase and achieve better performance than state-of-the-art methods.
\end{itemize}

The rest of this paper is organized as follows. 
Section~\ref{sec:related} presents the related work on non-IID FL involving performance enhancement and data privacy protection.
Section~\ref{sec:problem} introduces FL preliminaries and problem statements.
Section~\ref{sec:approach} details our grouping-based FL approaches (i.e., FLDG and FLDG-L).
Section~\ref{sec:experiment} presents experimental results and Section~\ref{sec:conclusion} concludes the paper.

\section{Related Work}\label{sec:related}

Although FL has lower communication overhead than other distributed machine learning methods, it suffers from the diversity of data distributions on IoT devices~\cite{measuring}, especially for non-IID scenarios.
To address this problem, various strategies have been investigated.
For example, Sattler et al.~\cite{robust} dispatched public data and pre-trained model based on the public data to all the devices, and formed a system that only consists of 10 clients in total. 
Nonetheless, the settings they used are unrealistic, since practical FL would typically involve plenty of clients and much more complex data distributions.
Based on reinforcement learning, Wang et al.~\cite{reinforcement} introduced an effective FL method, which selects devices with the highest rewards to participate in each round of interactions. 
However, their approach requires Principle Component Analysis (PCA)- compressed model weights of all the devices to construct the state space for training reinforcement learning strategies, which introduces a large amount of additional communication and computation overhead, thus making its deployment on IoT devices incapable.
Aiming at improving the performance of FL in non-IID scenarios, device grouping methods \cite{reinforcement,multicenter,hierarchical} have been studied to address the problem of weight divergence.
For example, Xie et al.~\cite{multicenter} and Briggs et al.~\cite{hierarchical} replaced the unique global model with multiple global models according to the similarity of local models.
However, these approaches weaken the generalization capability of FL, i.e., IoT devices from different groups cannot learn from each other effectively. 
Based on the K-center method~\cite{active}, Wang et al.~\cite{reinforcement} presented a baseline method that groups devices based on the similarity of local models. 
Their approach selects one representative device from each group for local model aggregation in each round. 
Since the method is based on the model comparison every 10 rounds, it may easily cause the intolerable communication overhead. 
Meanwhile, model exposure may result in serious data privacy leak problems~\cite{industrial,location}.

To address the issues of data privacy in FL-based scenarios, various protection mechanisms have been proposed.
Typically, they can be classified into two categories, i.e., data-based privacy protection and feature map-based privacy protection.
For example, as a kind of data-based privacy protection, Bonawitz et al.~\cite{smc} presented a privacy-preserving protocol to support the secure aggregation in FL by exploiting secure multiparty computation.
Lu et al.~\cite{blockchain} designed a blockchain empowered secure data sharing architecture and shared the data model rather than revealing the actual data to maintain the privacy of data.
For the feature map-based privacy protection, Zhang et al.~\cite{yet} provided privacy assurance for multiparty deep learning with the homomorphic encryption scheme.
Although the above approaches are promising in protecting data privacy, they do not consider the similarity of data distribution among IoT devices.
Moreover, most of them introduce significant memory usage and computation costs.
Therefore, it is hard to directly apply them on IoT applications, which are resource-constrained.

Although FL has been widely used as a distributed machine learning approach in IoT design~\cite{communication,largescale},
most FL methods  focus on adversarial attack issues~\cite{survey,lens} rather than the performance in non-IID scenarios.
To the best of our knowledge, our work is the first attempt that utilizes the
data-based device grouping and LSH algorithm~\cite{similarity,recommendation} to enable a fast and accurate FL for non-IID scenarios without exposing data privacy.

\section{FL Preliminaries and Problem Statements}\label{sec:problem}

\begin{figure*}[htbp]
	\centering
	\subfigure[MNIST]{
    \label{fig:mnist} 
	\includegraphics[width=0.32\linewidth]{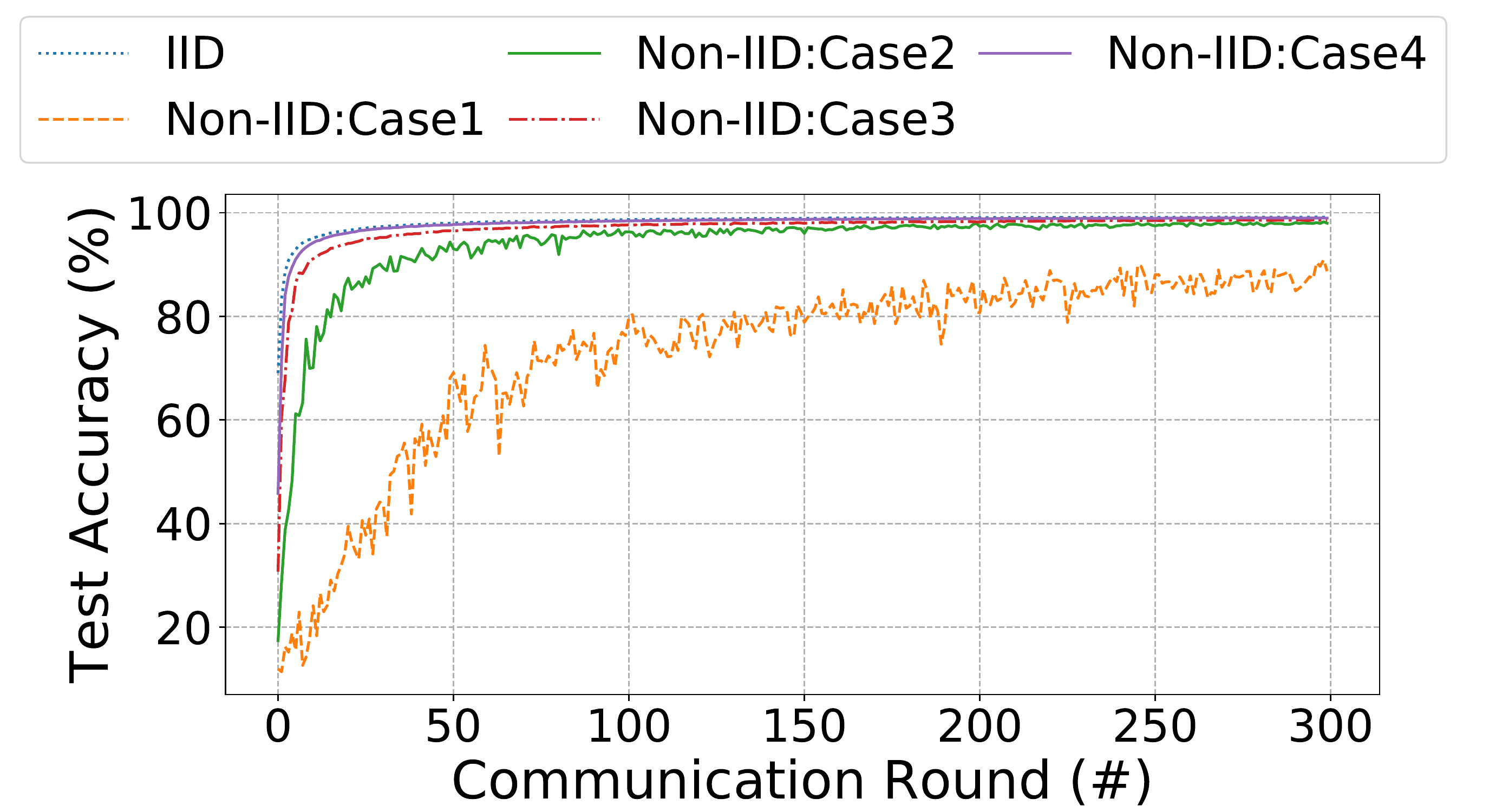}}
	\subfigure[Fashion-MNIST]{
    \label{fig:fashionmnist} 
	\includegraphics[width=0.32\linewidth]{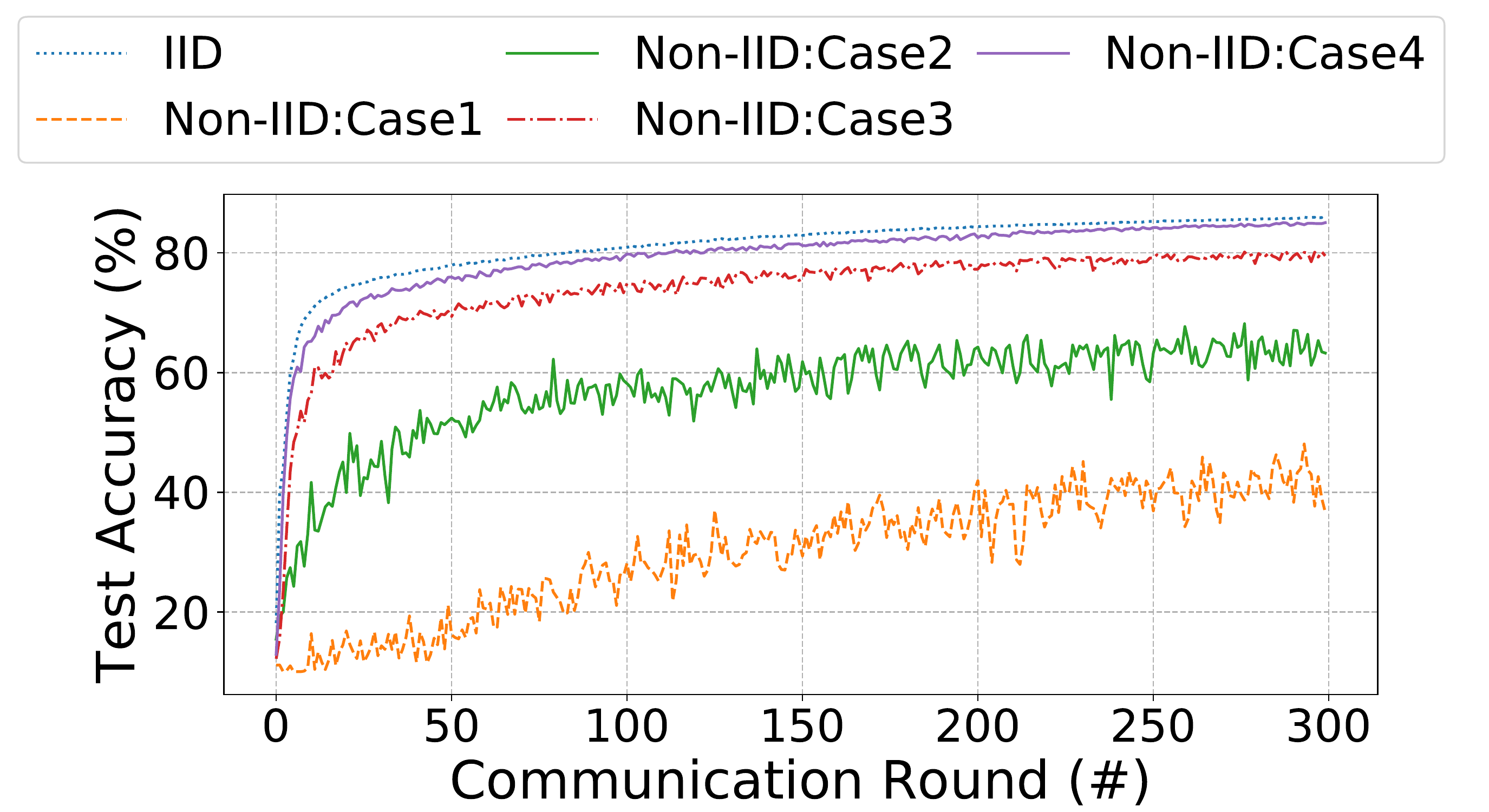}}
	\subfigure[CIFAR-10]{
    \label{fig:cifar} 
	\includegraphics[width=0.32\linewidth]{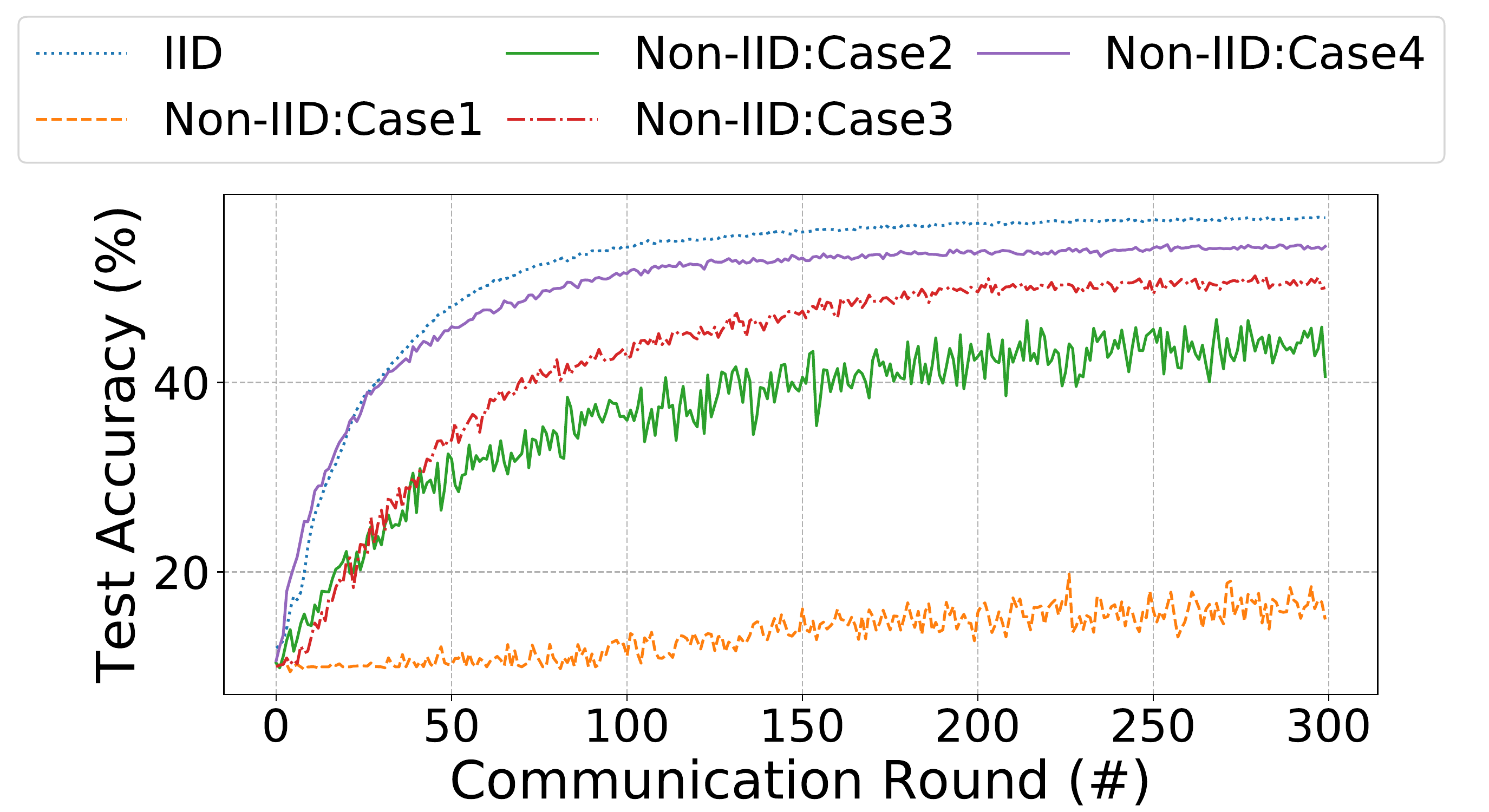}}
	\caption{Test accuracy of vanilla FL with 100 IoT devices for both IID and non-IID scenarios. For MNIST and Fashion-MNIST, each device has 600 training samples. For  CIFAR-10, each device has 500 training samples. All the MNIST, Fashion-MNIST and CIFAR-10 have 10000 test samples, respectively. 
    }
	\label{fig:problem}
\end{figure*}

Based on the concept of distributed machine learning, FL trains a shared global model by iteratively aggregating local gradients from multiple independent devices~\cite{blockchain,towards,privacyenhanced}.
Due to the unstable network connection and limited network bandwidth in real IoT scenarios,
to save the network resources, vanilla FL randomly selects a subset of IoT devices instead of all the devices during each round of gradient aggregation~\cite{communication}.
Suppose that there are a total of $N$ devices and $K$ devices are randomly selected at the beginning of the $t^{th}$ round, where the $i^{th}$ of $K$ randomly selected devices has $D^{i}$ data samples. 
After the local training in round $t$, each local device model will be updated as follows:
\begin{equation}
    w_{t+1}^i = w_{t}^i - {\eta \nabla \ell}_t^i \left( w_{t}^i \right)
\end{equation}
where $w_{t}^i$ and $w_{t+1}^i$ represent the current global model and the updated global model of the $i^{th}$ device, $\eta$ is the learning rate and ${\nabla \ell}_t^{i}$ is the gradient of the $i^{th}$ device model in round $t$. 
To improve communication efficiency, each selected device will upload its weight difference (i.e., $\Delta_{t+1}^i$) rather than the updated model to the cloud for gradient aggregation at the end of each round. 
\begin{equation}
    \Delta_{t+1}^i = w_{t+1}^i - w_{t}^i
\end{equation}
Based on the weight difference information collected from the $K$ devices, the cloud updates the global model using the federated average algorithm shown as follows:
\begin{equation}
    w_{t+1} \leftarrow  \ w_{t} +  ({\sum_{i=1}^{K} D^i \Delta_{t+1}^i})/{\sum_{i=1}^{K} D^i}
\label{eq:3}
\end{equation}
where $({\sum_{i=1}^{K} D^i \Delta_{t+1}^i})/{\sum_{i=1}^{K} D^i}$ denotes the weighted average gradient of $K$ randomly selected devices in round $t$, $w_{t}$ and $w_{t+1}$ represent the current global model and the updated global model, respectively.

Although vanilla FL (i.e., FedAvg) has shown its excellent performance on collaborative learning with IID data, it cannot be directly applied to IoT scenarios.
This is because the data distributed on IoT devices is mostly non-IID due to various factors such as user preferences and device locations. 
When dealing with non-IID scenarios, the performance of vanilla FL degrades dramatically.
Figure~\ref{fig:problem} shows a performance comparison among IID and four non-IID scenarios for FedAvg,  which follow the distributions defined in Table~\ref{tab:dist}.

From Figure~\ref{fig:problem},  we can find that the FL model accuracy decreases as device data distribution becomes skewed.
As an example, for CIFAR-10 benchmark~\cite{data}, FedAvg converges within around 300 rounds for the IID scenario, and the accuracy can reach 57.37\%.

However, for the four cases (i.e., Cases 4, Case 3, Case 2, and Case 1) of non-IID scenarios, the accuracy at the $300^{th}$ round is  54.53\%, 50.62\%, 44.15\%, and 16.84\%, respectively.
Moreover, Case 1 of the non-IID FL model is still far from convergence at round 300. 
We can observe a similar trend for the Fashion-MNIST and the MNIST benchmarks~\cite{data}.
 The  trends of low accuracy and slow convergence rate for non-IID scenarios are mainly due to the weight divergence of device models \cite{flwithnoniid} during FL training. 
For  IID scenarios, gradient optimization directions of IoT devices are similar due to the same device data distributions. 
However, for non-IID scenarios, due to the random device selection and highly skewed device data distributions, the optimization directions of IoT devices in each FL round vary significantly.
This will strongly affect the optimization direction of the global model which is generated based on the federated averaging operation on local models. 
Since it is hard to find a global model that is optimized for all the IoT devices in non-IID scenarios, the model accuracy for non-IID scenarios is worse than the one for the IID scenario. 
Moreover, the variable optimization directions of local models result in unstable optimization direction of the global model, thus slowing down the convergence rate. 
In other words, more communication rounds are required between the cloud and IoT devices during the FL training, which is a waste of training time and network bandwidth.

\section{Our Grouping-Based FL Approaches}\label{sec:approach}

\subsection{An Overview of Our Grouping-based FL}\label{sec:overview}

In this paper, we assume that IoT devices scattered in different places have non-IID data. 
Therefore, we propose a novel Federated Learning with Device Grouping (FLDG) method to enable efficient FL for non-IID scenarios. 
Our FLDG approach divides the FL process into two phases: 
i) the preprocessing phase, which groups all the IoT devices based on the similarity of extracted feature maps from raw device data;  
and ii) the training phase, which iteratively trains the shared global model by aggregating the gradient information sent by randomly selected devices from the  groups obtained in preprocessing phase. 
Note that our approach groups IoT devices based on the extracted feature maps from raw device data rather than device models, which requires much less comparison efforts and communication cost.

\begin{figure}[htbp]
\centering
\subfigure[The preprocessing phase  of FLDG and FLDG-L]{
\label{fig:preprocessing} 
\includegraphics[width=0.9\linewidth]{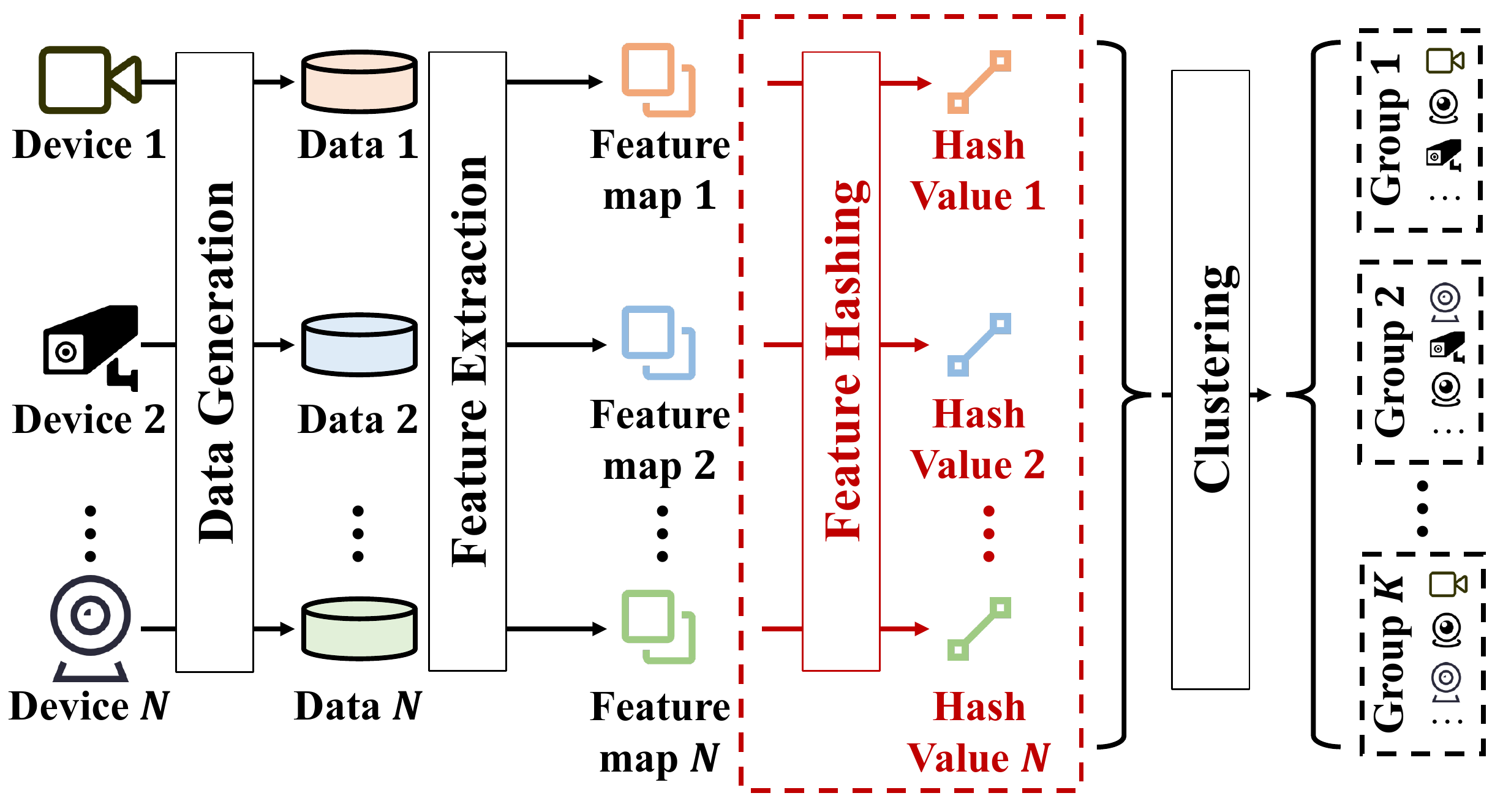}}

\subfigure[ The training phase of FLDG and FLDG-L]{
\label{fig:training}
\includegraphics[width=0.9\linewidth]{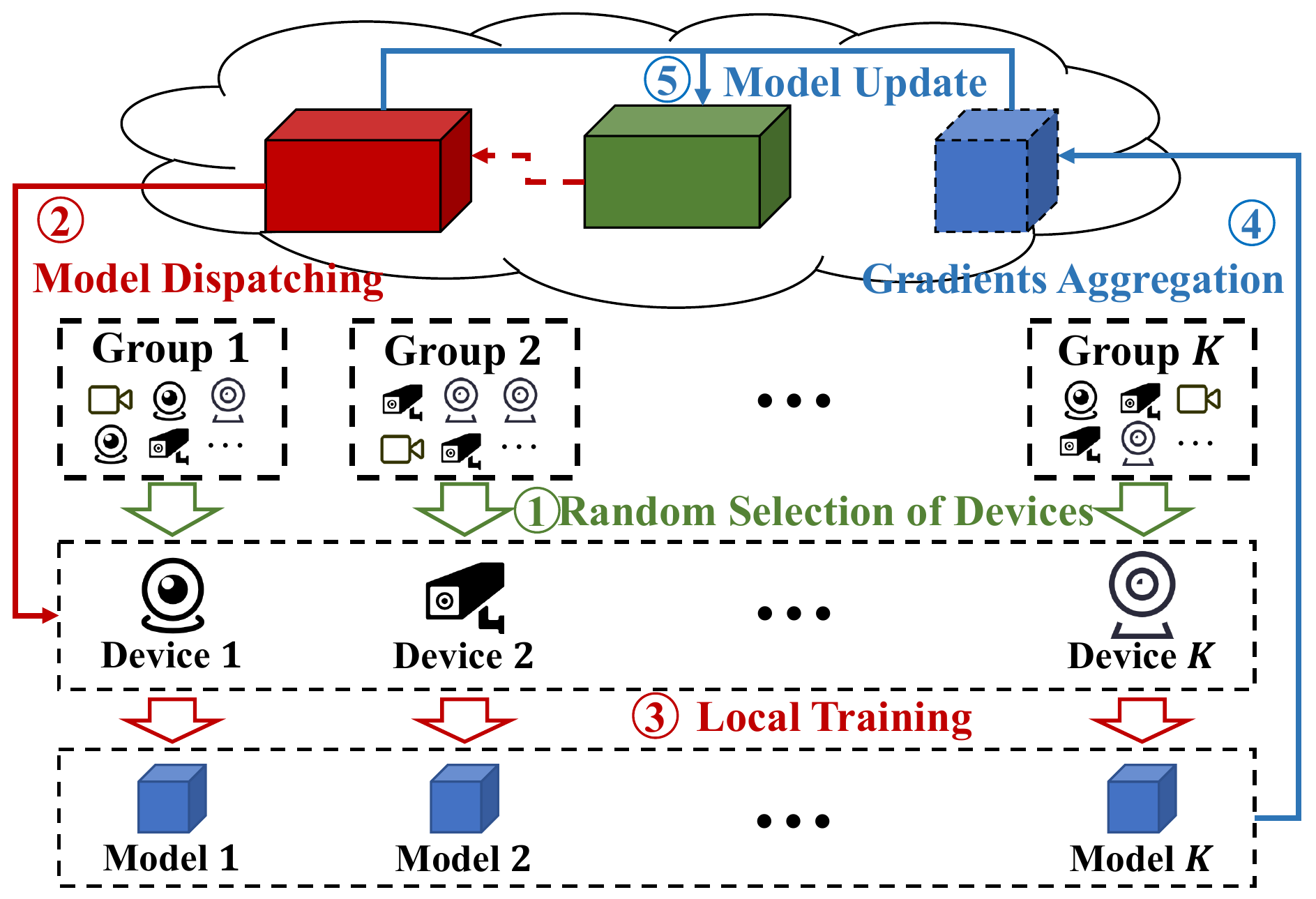}}
\caption{The workflows of FLDG and FLDG-L}
\label{fig:framework}
\end{figure}

Figure \ref{fig:framework} illustrates the workflows of our proposed FL methods, i.e.,  FLDG and FLDG-L, which are based on device grouping. 
Note that the workflow of FLDG-L is almost the same as that of FLDG.
The only difference is that FLDG clusters IoT devices based on feature maps while FLDG-L clusters IoT devices based on the LSH values of feature maps, thus the data privacy during the preprocessing phase can be further secured by using FLDG-L.  
Without taking the step shown in the red dashed box into account, Figure~\ref{fig:preprocessing} shows the preprocessing workflow for FLDG. 
Otherwise, the step shown in the red dashed box encodes the feature maps with the p-stable LSH functions, which makes the restore from LSH values to raw device data much more difficult. 
The details of the feature hashing using p-stable LSH functions will be introduced in the next
section. 
Note that for both FLDG and FLDG-L, the preprocessing phase only needs to be conducted once to train a new global model if the distributions of device data keep unchanged.
Therefore, the device grouping results remain unchanged throughout the whole training process after the preprocessing. 
Since the complexity of LSH and the K-Means algorithm is O(n), the overhead caused by both operations in preprocessing phase is much smaller than the overall workload of the entire FL framework.
Unlike vanilla FL that randomly selects IoT devices, both FLDG and FLDG-L only randomly select one IoT device from each group for the aggregation. 
Figure~\ref{fig:training} shows the training phase of such grouping-based FL methods.

\subsection{FLDG: FL with Device Grouping}\label{sec:fldg}

In our grouping-based FL, the preprocessing phase mainly deals with device grouping. 
It consists of three steps, i.e., data generation, feature extraction, and device clustering.
For the data generation step, IoT devices collect their labeled non-IID data samples and save them in memory for the device grouping and local model training.
 However, due to the large volume of collected data samples, it is not economical 
to directly conduct device clustering based on the similarity of these raw data.
Moreover, comparing raw data at the pixel level may be infeasible.
This is because, for the samples of the same category, they may have totally different representations.
Due to the lack of semantic information, comparing samples pixel by pixel may easily lead to incorrect device clustering results.
As an alternative, our approach conducts the device grouping based on the feature maps of raw device data, which can be extracted by some pre-trained model from libraries like Keras~\cite{keras} and Torchvision~\cite{torchvision}.
By using feature maps, the characteristics of raw data on an IoT device can be effectively and accurately expressed while the  size of comparison objects can be remarkably reduced.

\begin{algorithm}[htbp] 
	{\bf Inputs:}{\\
		i) $N$, \# of IoT devices\;
  ii) $D=\{D_1,$ $\cdots,D_N\}$, set of device datasets\; 
   iii)  $K$, \# of groups to be clustered\;
     iv) $E$, \# of local epochs; v) $\eta$, learning rate\;
     vi) $R$, \# of training rounds\; 
     vii) $G$, device groups;
	}\\
	       {\bf Preprocess($N, D, K, G$)}   \ \ {//Preprocessing}\\
		{\bf 1.}  $\overline{F}\leftarrow \{\}, G \leftarrow \{\}$ \\
		\For{ $i\leftarrow$ 1 to $N$}{
			{\bf 2.} $F \leftarrow \{\}$\\	
			\For{$j\leftarrow$  1 to $|D_i|$}{
				{\bf 3.} $F \leftarrow F\bigcup \{FeatureExtract(D_i^j)\}$
			}
			{\bf 4.}  $\overline{F} \leftarrow \overline{F} \bigcup \{AVG(F)\}$ \\ 
		}
		{\bf 5.} $G \leftarrow Kmeans(\overline{F}, K)$ \\
                {\bf 6.} {\bf return} $G$ \\

	{\bf CloudTrain($R, G, K$)}  \ \ \ \ //{Cloud training phase}\\
		{\bf 1.} $Initialize(w_0)$\\
		\For{$t\leftarrow$  1 to $R$}{
                        {\bf 2.} $S \leftarrow \{\}$\\
			\For{$j\leftarrow$  1 to $K$}{
				{\bf 3.} $S  \leftarrow S \bigcup RandomSelect(G_j)$ \\ 
			}
			{\bf 4.} $Dispatch(w_t,S)$\\
			\For{$i\leftarrow$  1 to $K$}{
				{\bf 4.} $\Delta_{t+1}^i   \leftarrow \text{LocalTrain}(Idx(S_i),   w_{t})$\\
			}
			{\bf 5.} {$  w_{t+1} \leftarrow \ w_t +  \frac{\sum_{i=1}^{K} |D_{Idx(S_i)}| \times \Delta_{t+1}^i}{{\sum_{i=1}^{K}} |D_{Idx(S_i)}|}$}\\
		}

	{\bf LocalTrain($d, w$)}    \ //Local training on the $d^{th}$ device\\
		{\bf 1.} $temp \leftarrow w$, $w^d \leftarrow w$\\
		\For{$i\leftarrow$  1 to $E$}{
			{\bf 2.} $w^d \leftarrow  w^d - {\eta \nabla \ell}^d \left( w^d \right)$\\
		}
		{\bf 3.} {\bf return} $  w^d - temp$\\
	\caption{Implementation of FLDG} 
	\label{alg:1}
\end{algorithm}

The training phase of FLDG is similar to that of vanilla FL except for the device selection step (i.e., step 1 in 
Figure~\ref{fig:training}). To reduce the communication overhead, vanilla FL 
does not take all the IoT devices in one FL training round. 
Instead, only a subset of IoT devices is randomly selected for the gradient aggregation.
 As aforementioned, this will cause the optimization direction divergence of the global model, since model optimization directions in different devices are usually different in non-IID scenarios. 
However, devices in the same group usually have similar model optimization directions.
To improve the performance of FL in non-IID scenarios, our approach randomly selects one IoT device from each group during each round of model training.
In this way, we can make the optimization direction of the global model in each round consistent with its final optimization direction, thus accelerating the model convergence rate of non-IID FL training.
Note that the more groups we clustered in the preprocessing phase, the more IoT devices will be selected in the training phase to participate in the gradient aggregation.

Algorithm~\ref{alg:1} shows the key steps involved in our FLDG algorithm.
The procedure {\it Preprocess} shows the implementation of the preprocessing phase.
In this algorithm, step 1 initializes the data structures, where $\overline{F}$
and $G$ are used for storing the device averaged feature maps and device groups, respectively. 
Steps 2-4 iteratively figure out the feature maps of the $N$ devices. 
In step 3, we use $D_i^j$ to denote the $j^{th}$ sample on the 
$i^{th}$ device. 
Step 4 calculates the average feature map value of all the samples on the $i^{th}$ device. 
After collecting averaged feature maps from all the IoT devices, step 5 clusters the IoT devices based on the similarity of the feature maps using the classical K-Means algorithm.
Finally, step 6 reports the clustered $K$ device groups. 
The procedure {\it CloudTrain} presents the details of the training process of the global model 
on cloud. Step 1 initializes the global model with $w_0$. 
Steps 2-5 show the implementation of each training round, where $S$ is used to save the selected devices.
At the beginning of each round, steps 2-3 randomly select one IoT device from each device group using the function {\it RandomSelect}.
Then step 4 dispatches  the global model to all the selected devices in $S$. 
In step 5, the cloud waits for the gradient information from all the selected devices for the model aggregation.
Note that only when all the gradients are received from the selected devices as shown in step 5, the cloud can start the aggregation operation. 
The procedure {\it LocalTrain} shows the local training process for a specific IoT device based on the definition in Equation (1).

\subsection{FLDG-L: Improved FLDG with LSH}\label{sec:fldgl}

In FLDG, our approach  groups IoT devices based on the similarity between extracted feature maps.
Therefore, all the selected devices should send their averaged feature maps to the cloud, which may lead to high risk of privacy exposure. 
For example, by using model inversion attack~\cite{inversion,collaborative} or membership inference attack~\cite{membership}, raw data can be restored from the feature maps, resulting serious privacy leaking problems. 
To address this problem, we propose an improved version of FLDG named FLDG-L, which
encodes feature maps based on the hash functions from the p-stable LSH family~\cite{datar2004locality}.
The hash function $F_{a,b}$ is formulated as $F_{a,b} = \lfloor \frac{a \cdot v + b}{r}\rfloor$, where $a$ is a vector with entries chosen independently from a p-stable distribution, $b$ is a real number chosen uniformly from the range $\left[0,r\right]$, $r$ is a positive real number which represents the window size.
In FLDG-L, selected devices only send the hash values of averaged feature maps to the cloud, thus the risk of privacy exposure can be greatly reduced.  
The following subsections will introduce the implementation details of FLDG-L and analyze its capability of privacy protection.

\subsubsection{Implementation of FLDG-L}

FLDG-L adopts the similar workflow as FLDG as shown in Figure~\ref{fig:framework}.
The only difference is that FLDG-L adds a step of extracted feature map hashing before the device grouping in the cloud as shown in Figure~\ref{fig:preprocessing}.
Since the FLDG-L training procedures for the global model and local device models 
are the same as the one of FLGD, we omit the introduction to these parts in this subsection.

\begin{algorithm}[htp] 
	{\bf Inputs:}{\\
		i) $N$, \# of IoT devices\;
  ii) $D=\{D_1,$ $\cdots,D_N\}$, set of device datasets\; 
   iii)  $K$, \# of groups to be clustered\;
     iv) $G$, device groups\; 
     v) $LSHF$, p-stable LSH family
	}\\

{\bf Preprocess($N, D, K, G, LSHF$)}  \ \   {//Preprocessing}\\
		{\bf 1.} $G\leftarrow, \{\}$,  $\overline{F}\leftarrow \{\}$\\
                {\bf 2.} $HV\leftarrow <hv_1,\cdots,hv_{|LSHF|}> $\\
		\For{$i\leftarrow$ 1 to $N$}{
			{\bf 3.}   $F \leftarrow \{\}$\\
			\For{$j\leftarrow$ 1 to $|D_i|$}{
				{\bf 4.}  $F\leftarrow F\bigcup \{FeatureExtract(D_i^j)\}$\\ %
			}
			{\bf 5.}  $avg \leftarrow AVG(F)$ \\

			\For{$j\leftarrow $ 1 to $|LSHF|$}{
				{\bf 6.}  $ hv_j \leftarrow LSHF_j(avg)$ \\
			}	
			{\bf 7.}  $ \overline{F} \leftarrow \overline{F} \bigcup \{HV\}$ \\
		}
		{\bf 8.} $G \leftarrow Kmeans(\overline{F}, K)$ \\
		{\bf 9.} {\bf return } $G$\\
	\caption{Implementation of FLDG-L} 
	\label{alg:2}
\end{algorithm}

Algorithm~\ref{alg:2} presents the detailed preprocessing procedure of FLDG-L. 
Unlike FLDG, FLDG-L maintains a family of p-stable LSH functions~\cite{similarity}. At the beginning of FLDG-L, all the IoT devices need to download these LSH functions to enable the following encoding of device feature maps. 
In step 1, the algorithm initializes the data structures, where $G$ and $\overline{F}$  are used to keep the grouping results and the hash values of all the device averaged feature maps, respectively. 
Step 2 constructs a vector data structure with an output dimension of $|LSHF|$ to hold all the hash values for the given p-stable LSH family.
Steps 3-7 iteratively calculate the device averaged feature maps and corresponding 
hash values for all the IoT devices. 
For each IoT device, steps 3-4 try to extract the figure maps for all of its samples, and step 5 figures out their average value. 
Step 6 applies each LSH function of the given p-stable LSH family on the device averaged feature map, and step 7 saves the hash value vector of feature maps for the $i^{th}$ device. 
Based on the classical K-Means method, all the IoT devices can be clustered into $K$ groups based on the hash values of device feature maps. Note that to ensure that there are $K$ groups to be clustered, the p-stable LSH family should contain at least  $\lceil {\log_r}^K \rceil$ LSH  functions.

\subsubsection{Discussions on LSH-based Privacy Protection}

The following definition formalizes the p-stable LSH family, where LSH functions should have the following properties:
i) If two original instances are similar, their hash values using an LSH function are similar with a high probability.
ii) If two original instances are different, their hash values using an LSH function are different with a high probability.

\begin{definition}
An LSH function family $LSHF = \lbrace h : S \rightarrow U \rbrace$ is called $(r_1, \ r_2, \ p_1, \ p_2)$-sensitive if  for any two instances $x,  y  \in  S$ and for any $h  \in LSH$ the following conditions hold:
    \begin{itemize}
    \item If $d(x,  y) \leq  r_1$, then $\Pr[h(x) = h(y)]  \geq p_1$,
    \item If $d(x,  y) \geq  r_2$, then $\Pr[h(x) = h(y)]  \leq p_2$,
    \end{itemize}
where $d(x,y)$ denotes the Euclidean distance between $x$ and $y$, $\left\{r_1,r_2,p_1,p_2\right\}$ are a set of constants, and $\Pr[h(x) = h(y)]$ indicates the probability that the  hash values of $x$ and $y$ are equal. The output dimension of the LSH function family is the same as the cardinality of $LSHF$, i.e., $|LSHF|$.

\end{definition}

According to the properties of traditional hash functions, for a given hash value, it is extremely hard to restore its original data even if the hash function is known.
Therefore, by using hash values of feature maps, the data privacy of IoT devices can be guaranteed. 
Moreover, based on the properties of p-stable LSH family in definition 1, the hash values still partially keep the data characteristics (e.g., distributions) of the original data. Moreover, although the size of hash values is much smaller than original data, they still contain enough information to enable 
the similarity comparison. Therefore, by using p-stable LSH functions, the data privacy and the similarity comparison accuracy can be simultaneously guaranteed.  
Meanwhile, due to the small-size of hash values, the network communication overhead will be drastically reduced between the cloud and IoT devices. 
Note that when the output dimension of a p-stable LSH family is small, it is difficult to form a sufficient number of groups, which may reduce the inference 
performance of FL.
However, if more p-stable LSH functions are used in FLDG-L, the computation and communication overhead will become larger. 
Therefore, there should be a trade-off between model accuracy and the output dimension of p-stable LSH functions.

\section{Experimental Results}\label{sec:experiment}

To evaluate the performance of FLDG and FLDG-L, we performed experiments on a workstation with Intel i7-9700k CPU, 16GB memory, NVIDIA GeForce GTX 1080Ti GPU, and Ubuntu operating system (version 18.04).
We conducted three experiments on benchmarks MNIST, Fashion-MNIST, and CIFAR-10~\cite{data}, respectively, and constructed corresponding FL models as shown in Table \ref{tab:model} using PyTorch (version 1.3.1).
For each convolution layer (i.e., {\it Conv2D}), we provide the information of input and output dimensions, kernel size, stride and padding. 
For the fully connected layer (i.e., {\it FC}), we provide the information about input and output dimensions. 
For the dropout layer (i.e., {\it DropOut}), we provide the probability of an element to be zeroed.
To enable device grouping, we used the pre-trained MobileNetV2 model from Keras (version 2.3.1) to extract feature maps (with a size of $1280\times 1$) from device raw data.
Since FLDG and FLDG-L involve random operations (e.g., device selection from each group), we ran each experiment ten times and used the mean value for fair comparison.

\begin{table}[htp]
\centering
\caption{Model structure for the three datasets.}
\resizebox{\linewidth}{!}{
\begin{tabular}{c|l}
\toprule[2pt]
\multicolumn{1}{c|}{Benchmark} & \multicolumn{1}{c}{Model structure configuration}   \\ \hline
\multirow{4}*{MNIST}                           & Conv2D(1, 20, 5, 1, 2), MaxPool, ReLU                \\  
                                & Conv2D(20, 50, 5, 1, 2), DropOut(0.5), MaxPool, ReLU \\ 
                                & FC(800, 50), ReLU                                    \\  
                                & DropOut(0.5), FC(50, 10)                             \\  \hline
\multirow{4}*{\shortstack{Fashion\\ -MNIST}}                 & Conv2D(1, 16, 5, 1, 2), MaxPool, ReLU                \\  
                                & Conv2D(16, 32, 5, 1, 2), DropOut(0.5), MaxPool, ReLU \\  
                                & FC(512, 50), ReLU                                    \\  
                                & DropOut(0.5), FC(50, 10)                             \\ \hline
\multirow{5}*{CIFAR-10}  & Conv2D(3, 6, 5, 1, 2), ReLU, MaxPool                 \\ 
                                & Conv2D(6, 16, 5, 1, 2), ReLU, MaxPool                \\  
                                & FC(400, 120), ReLU                                   \\  
                                & FC(120, 84), ReLU                                    \\  
                                & FC(84, 10)                                           \\  \bottomrule[2pt]
\end{tabular}
}
\label{tab:model}
\end{table}

In the experiments, we considered an IoT application with 100 IoT devices. For FL training, we set the default number of groups and learning rate $\eta$ to 10 and 0.01, respectively. 
For each IoT device, we set the batch size and epoch of local training to 50 and 5, 
respectively.
For the LSH functions, we choose the window size $r = 3.0$.
According to \cite{measuring}, we synthesized four non-IID data distributions shown in Table \ref{tab:dist} based on Dirichlet distribution to generate FL training data.
For each experiment, we checked the test accuracy of obtained global models based on all the benchmark test data. 
The following sub-sections firstly investigate the impacts of output dimension and the number of groups on FLDG and FLDG-L, and then compare the performance of FLDG and FLDG-L with state-of-the-art methods.

\begin{table}[h]
\centering
\caption{Four cases of non-IID  data distributions}
\resizebox{\linewidth}{!}{
\begin{tabular}{c|c|c|c}
\toprule[2pt]
Dataset            & MNIST                                   & Fashion-MNIST                                  & CIFAR-10                                \\ \hline
Sample \# in total   & 60000                                   & 60000                                          & 50000                                   \\ \hline
Sample \# per IoT device & 600                                     & 600                                            & 500                                     \\ \midrule[1pt]
Case \#           & \multicolumn{3}{c}{Data Distributions}                                                                                             \\ \hline
1                  & \multicolumn{3}{c}{100\% belong to one label}                                                                                     \\ \hline
2                  & \multicolumn{3}{c}{100\% evenly belong to two labels}                                                                             \\ \hline
3                  & \multicolumn{3}{c}{\begin{tabular}[c]{@{}c@{}}80\% belong to one label,\\ the remaining 20\% belong to other labels\end{tabular}} \\ \hline
4                  & \multicolumn{3}{c}{\begin{tabular}[c]{@{}c@{}}50\% belong to one label,\\ the remaining 50\% belong to other labels\end{tabular}} \\ \bottomrule[2pt]
\end{tabular}
}
\label{tab:dist}
\end{table}

\begin{figure*}[htp]
\centering
\subfigure[Case 1 of MNIST]{
\includegraphics[width=0.23\linewidth]{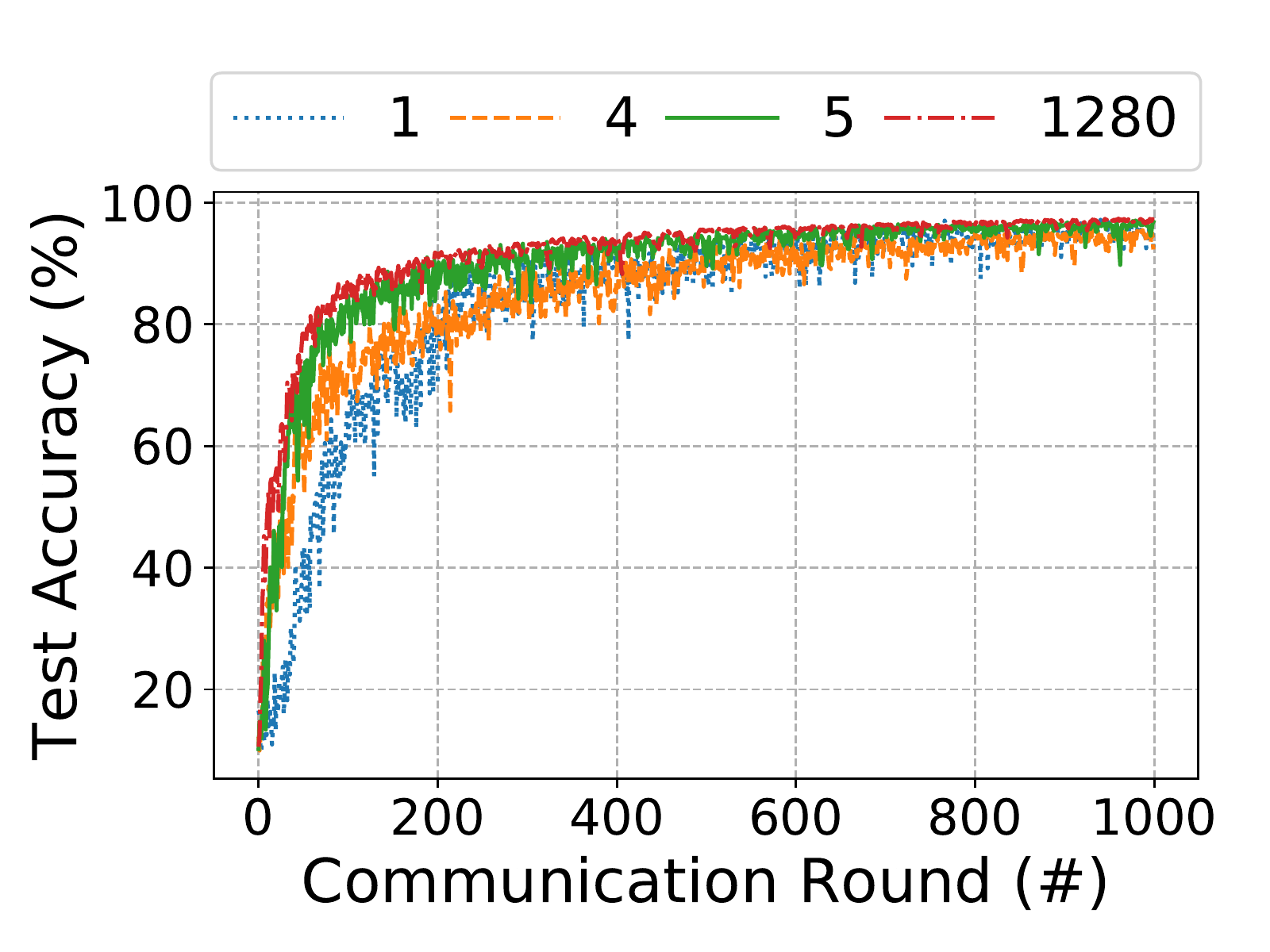}}
\subfigure[Case 2 of MNIST]{
\includegraphics[width=0.23\linewidth]{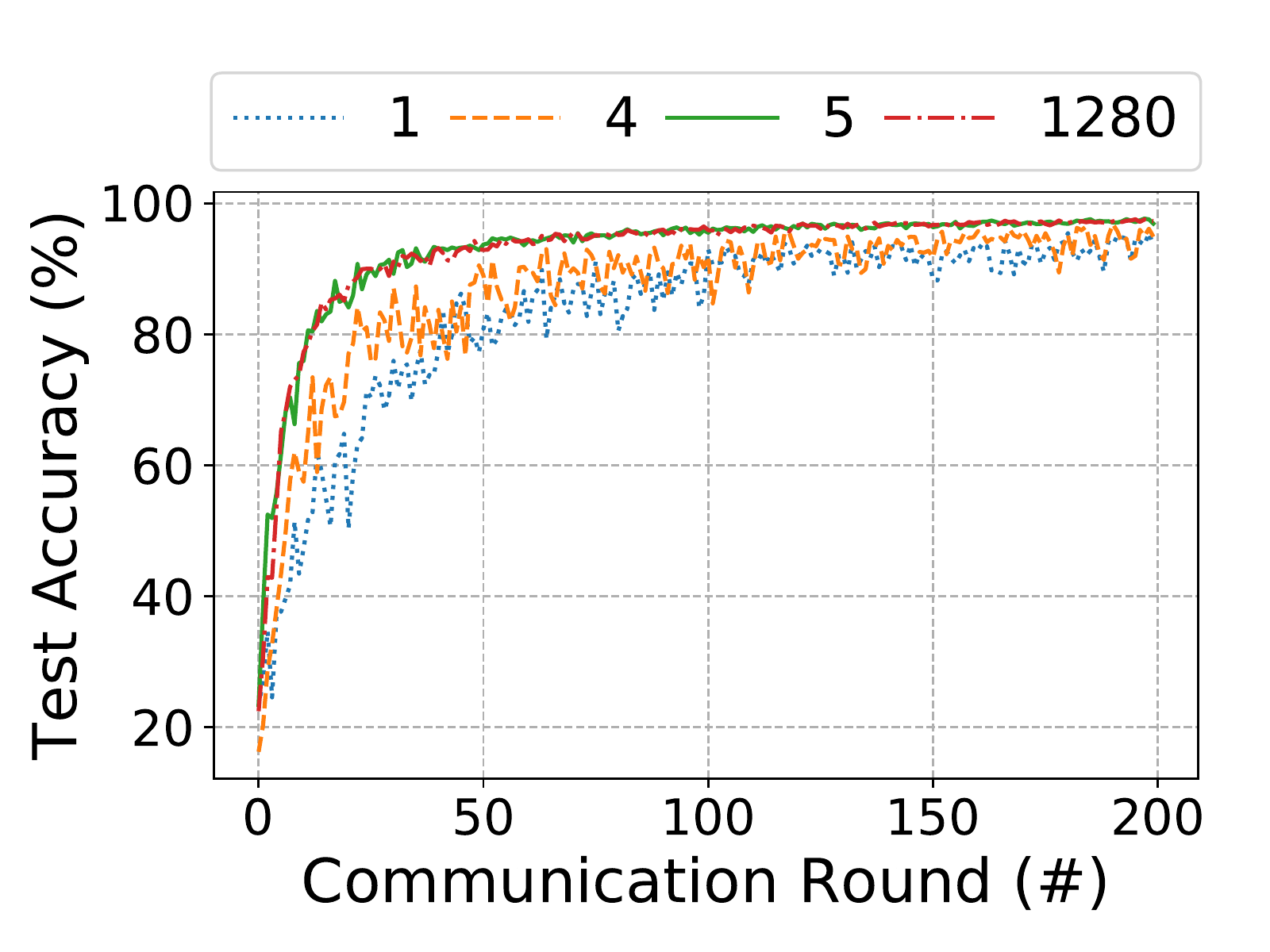}}
\subfigure[Case 3 of MNIST]{
\includegraphics[width=0.23\linewidth]{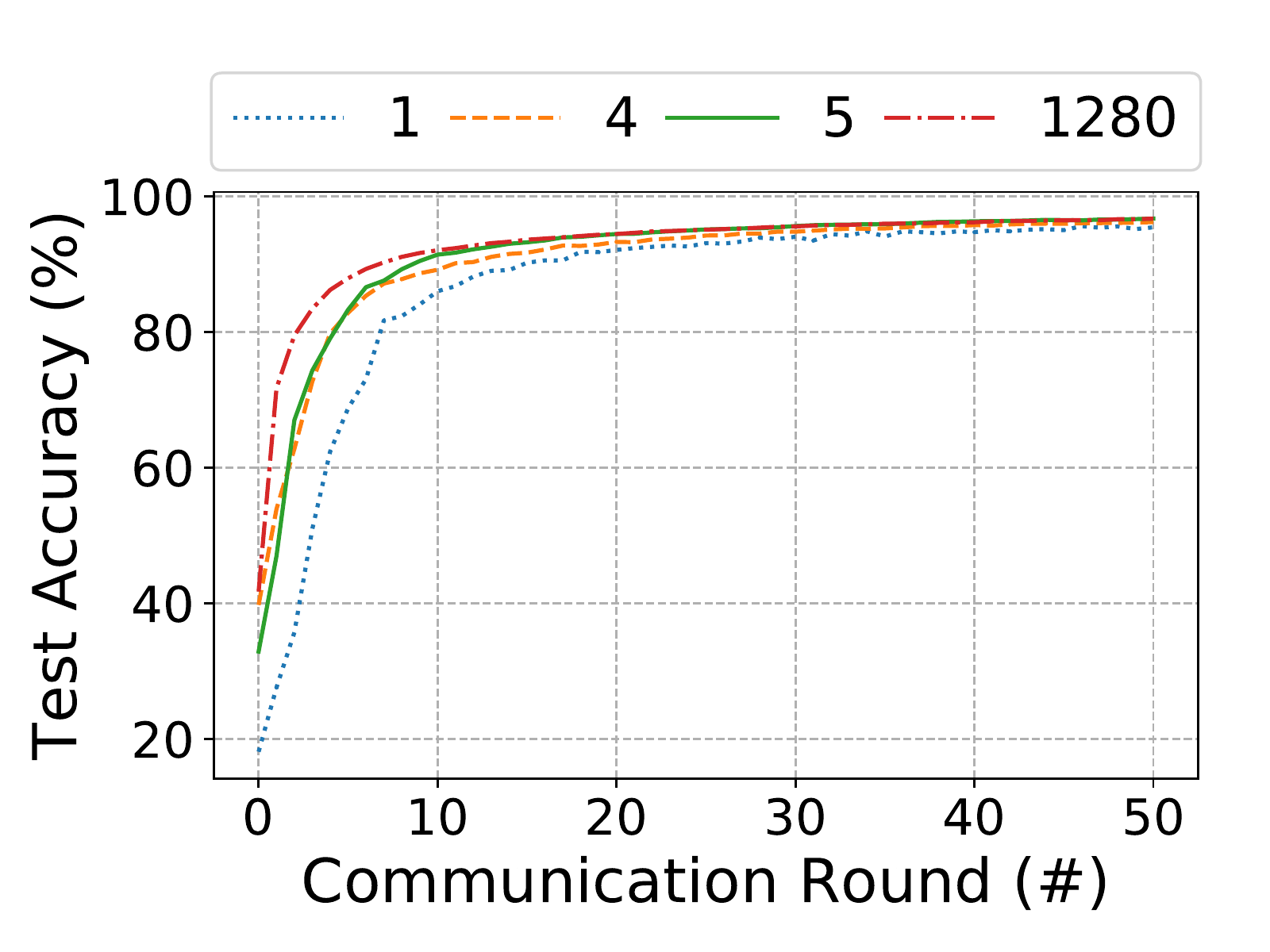}}
\subfigure[Case 4 of MNIST]{
\includegraphics[width=0.23\linewidth]{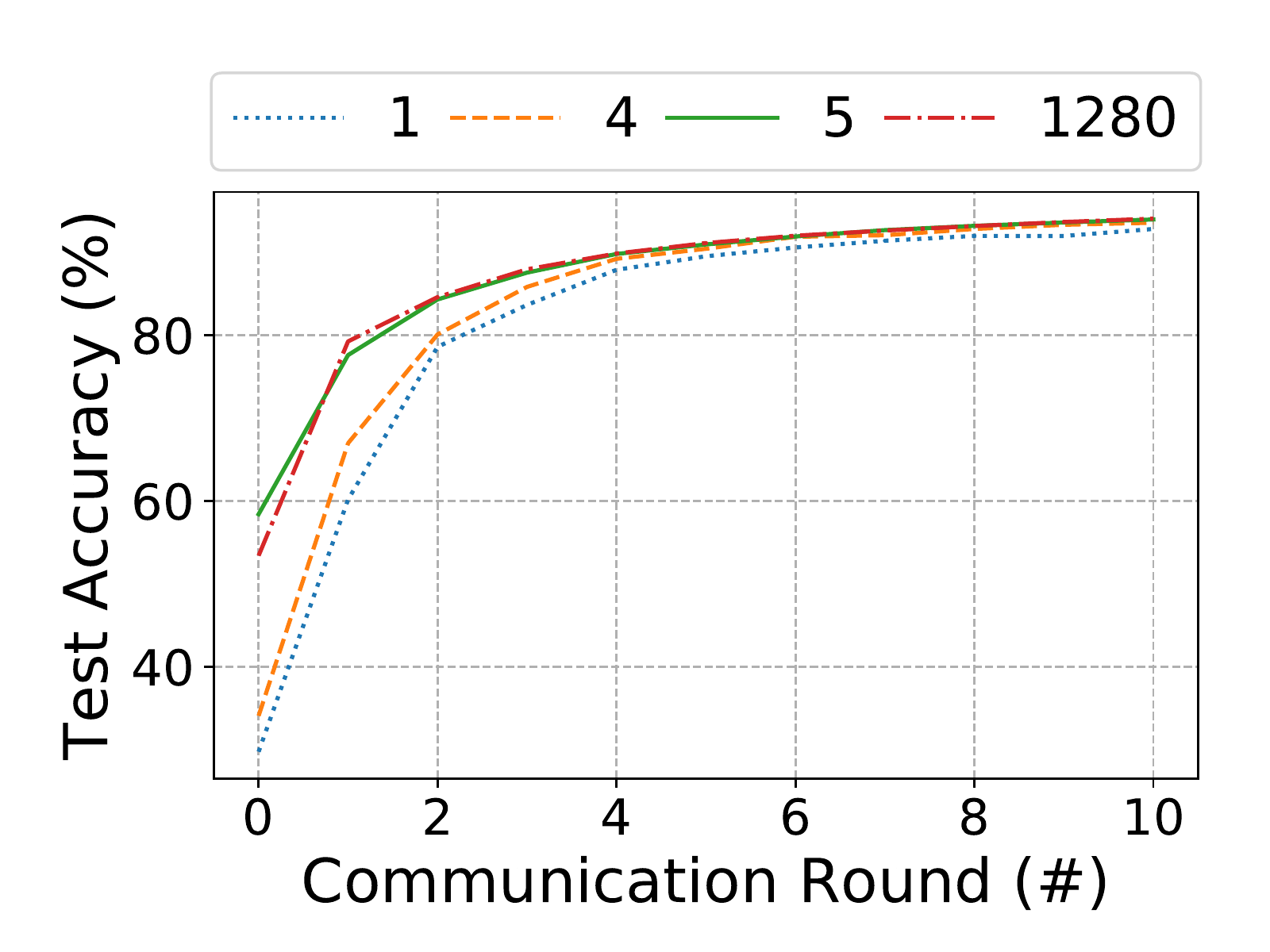}}
 \vspace{-0.1in}
\subfigure[Case 1 of Fashion-MNIST]{
\includegraphics[width=0.23\linewidth]{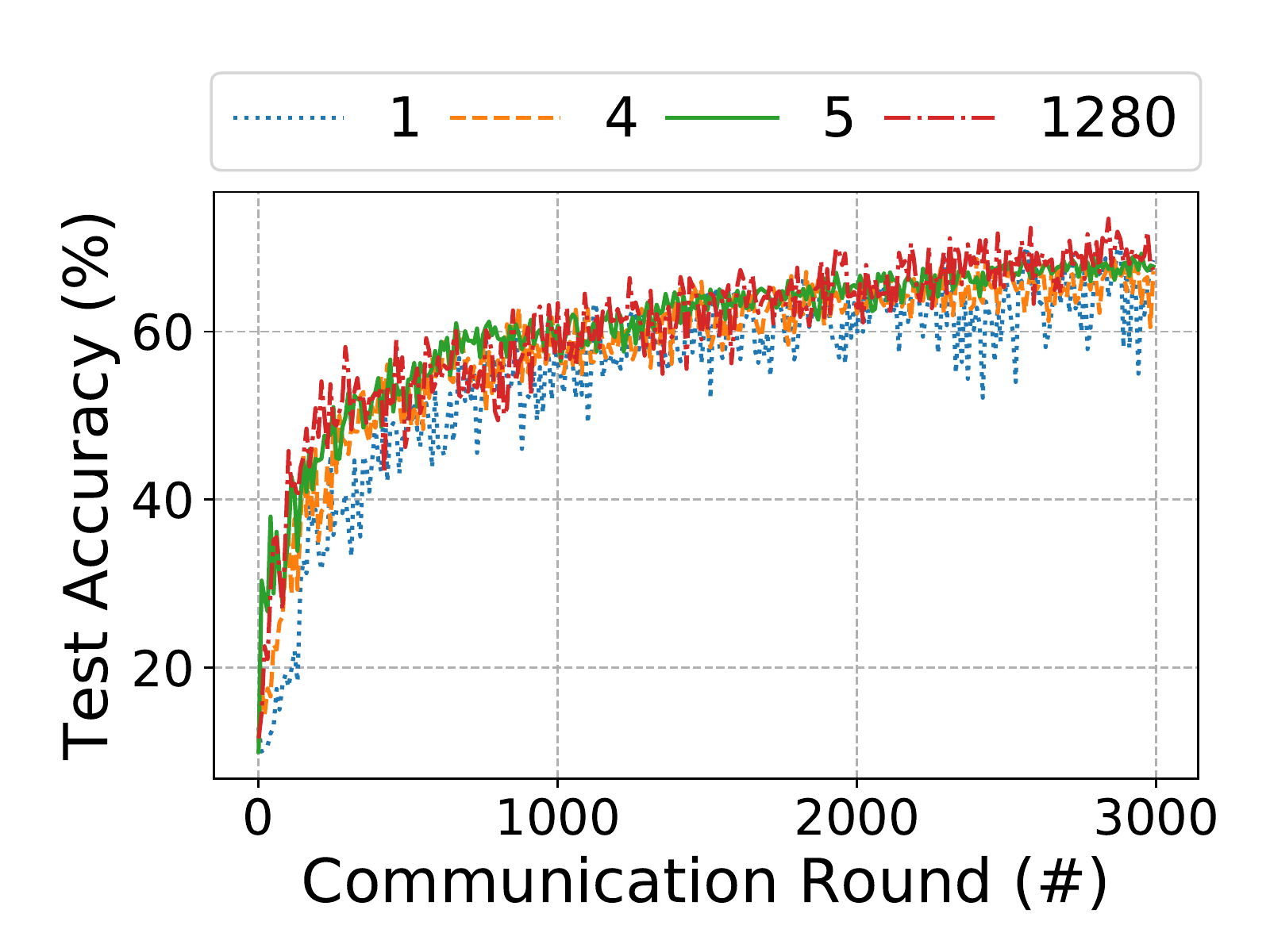}}
\subfigure[Case 2 of Fashion-MNIST]{
\includegraphics[width=0.23\linewidth]{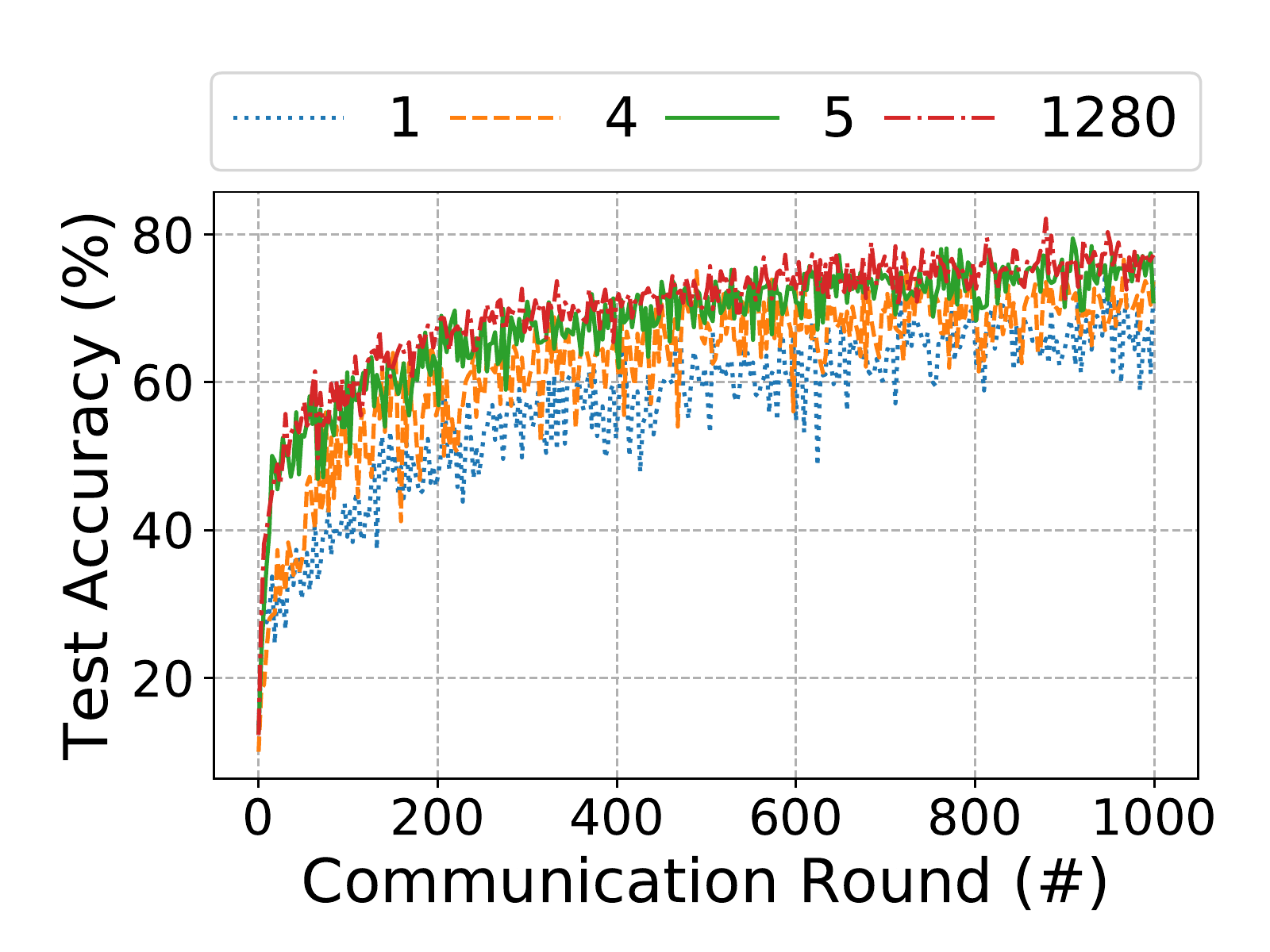}}
\subfigure[Case 3 of Fashion-MNIST]{
\includegraphics[width=0.23\linewidth]{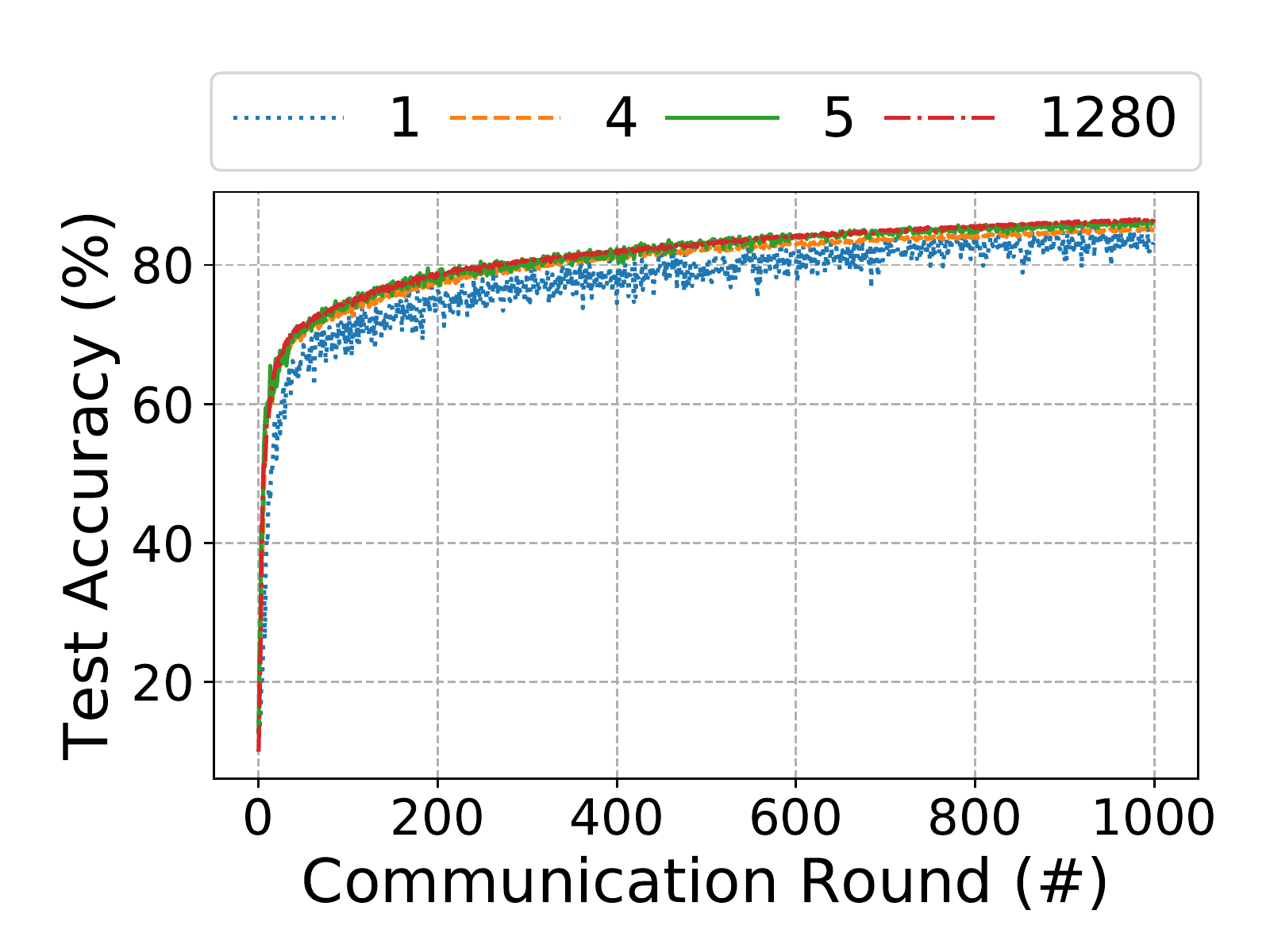}}
\subfigure[Case 4 of Fashion-MNIST]{
\includegraphics[width=0.23\linewidth]{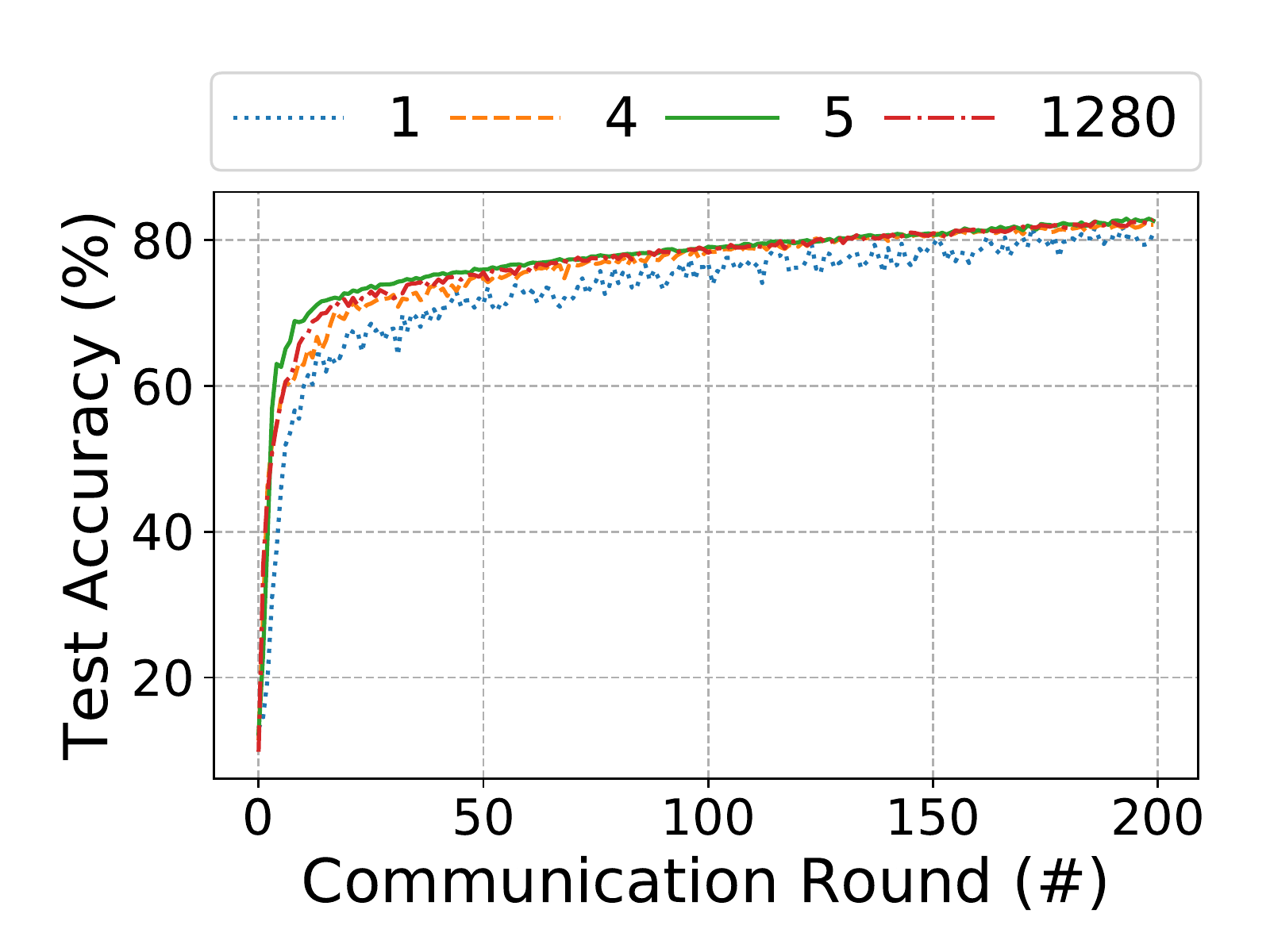}}
 \vspace{-0.1in}
\subfigure[Case 1 of CIFAR-10]{
\includegraphics[width=0.23\linewidth]{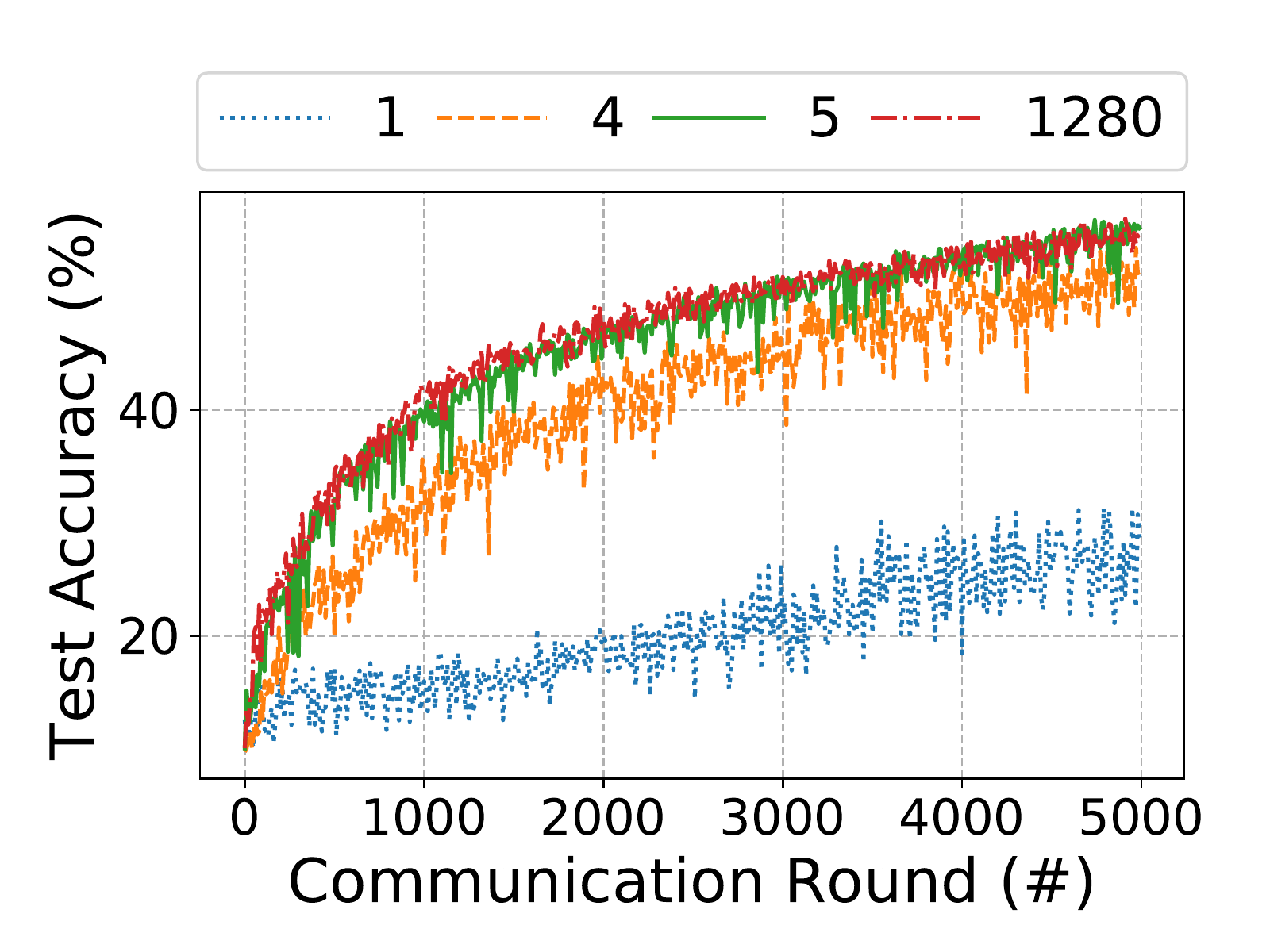}}
\subfigure[Case 2 of CIFAR-10]{
\includegraphics[width=0.23\linewidth]{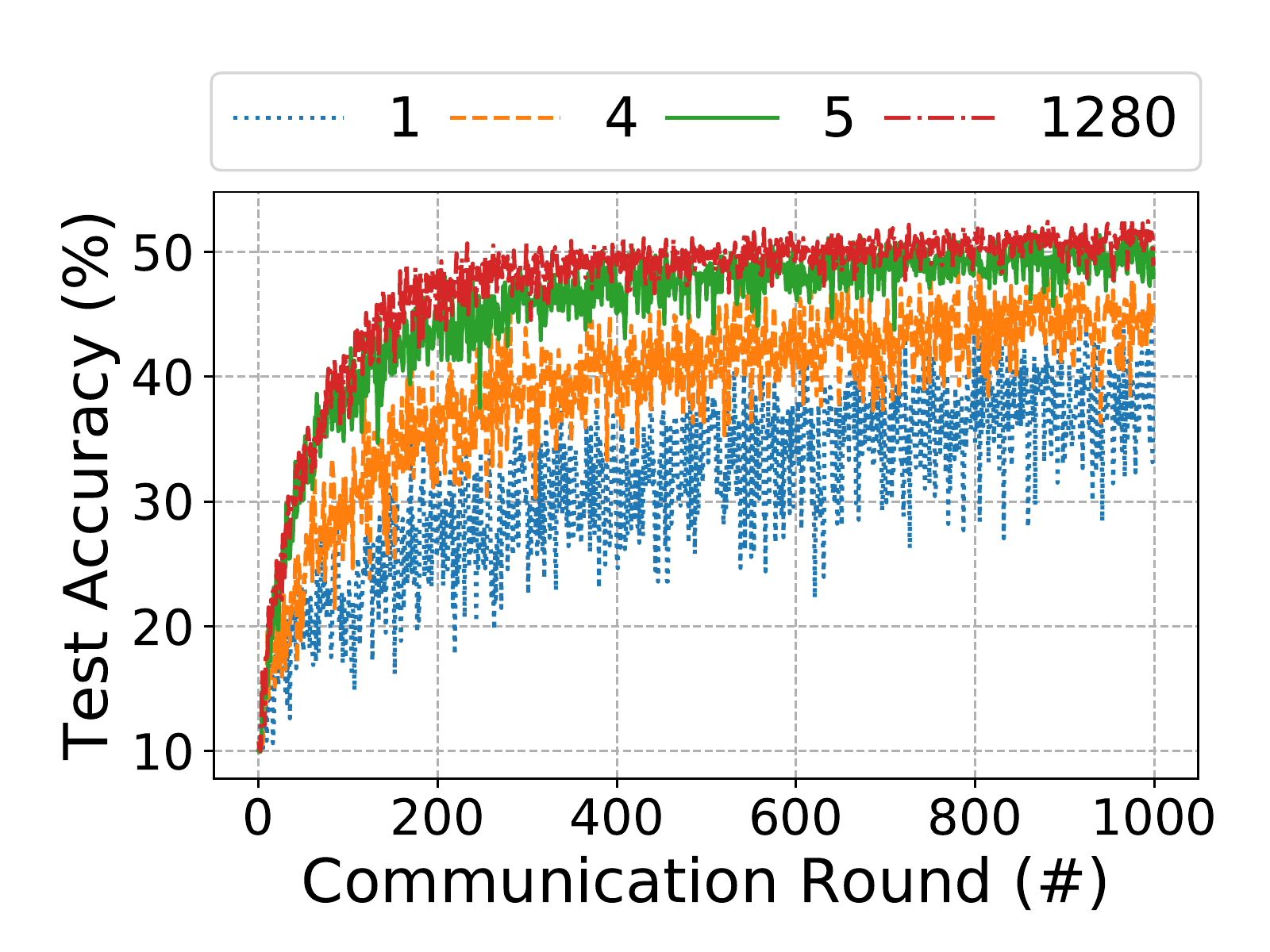}}
\subfigure[Case 3 of CIFAR-10]{
\includegraphics[width=0.23\linewidth]{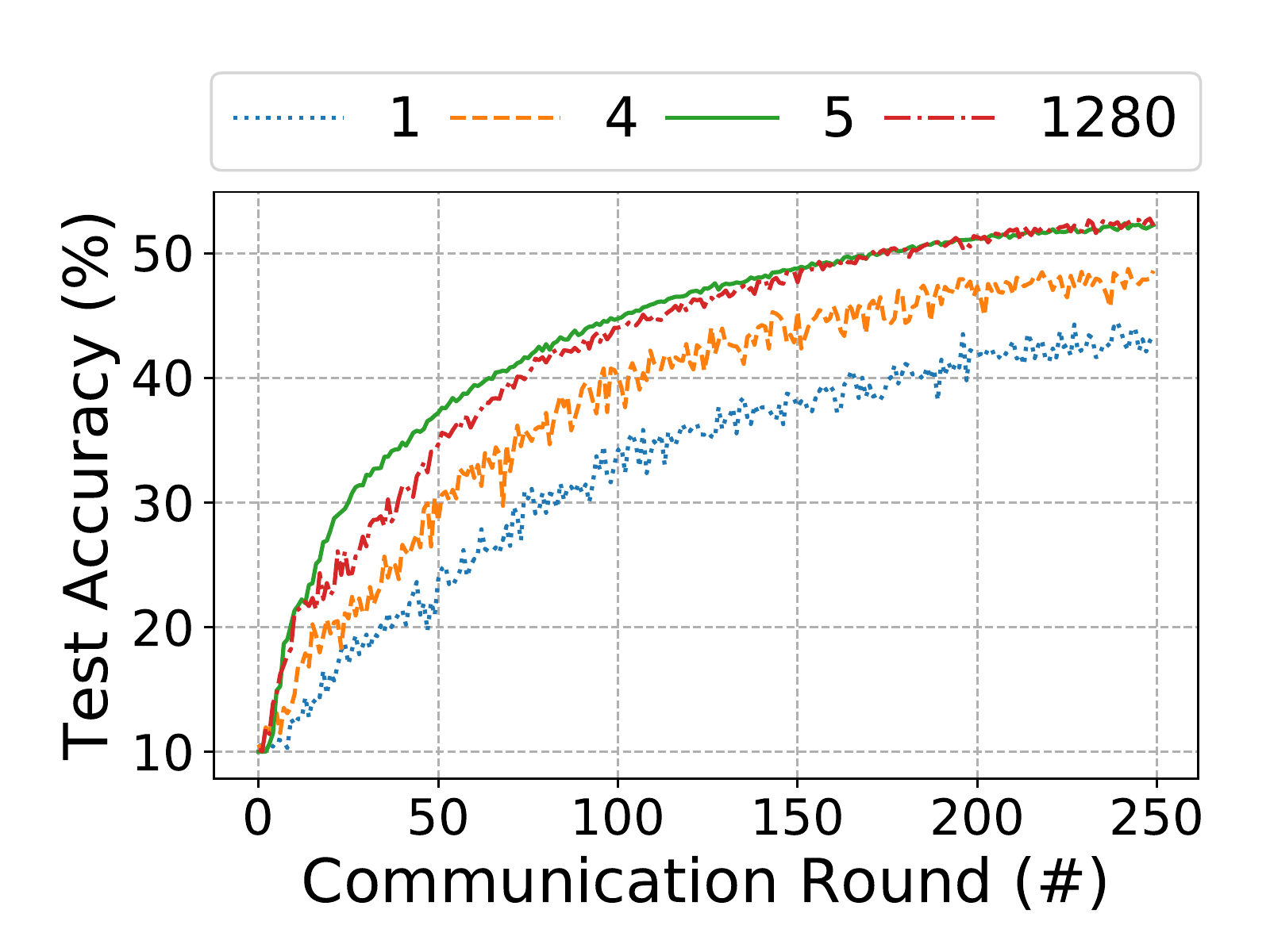}}
\subfigure[Case 4 of CIFAR-10]{
\includegraphics[width=0.23\linewidth]{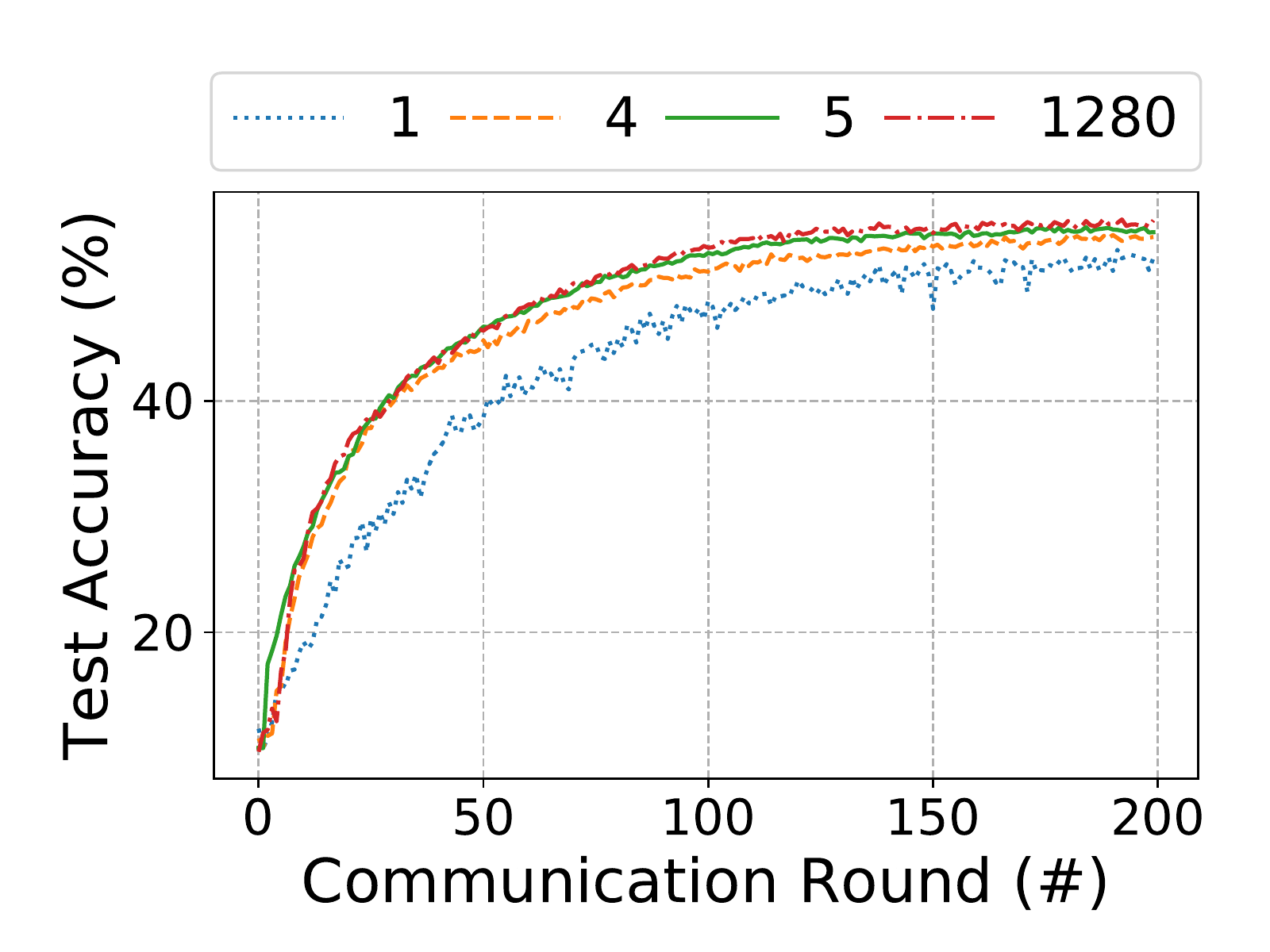}}
\caption{Test accuracy comparison w.r.t. different output dimensions using FLDG-L}
\vspace{-0.1in}
\label{fig:code}
\end{figure*}

\begin{figure*}[htp]
\centering
\subfigure[Case 1 of MNIST]{
\includegraphics[width=0.23\linewidth]{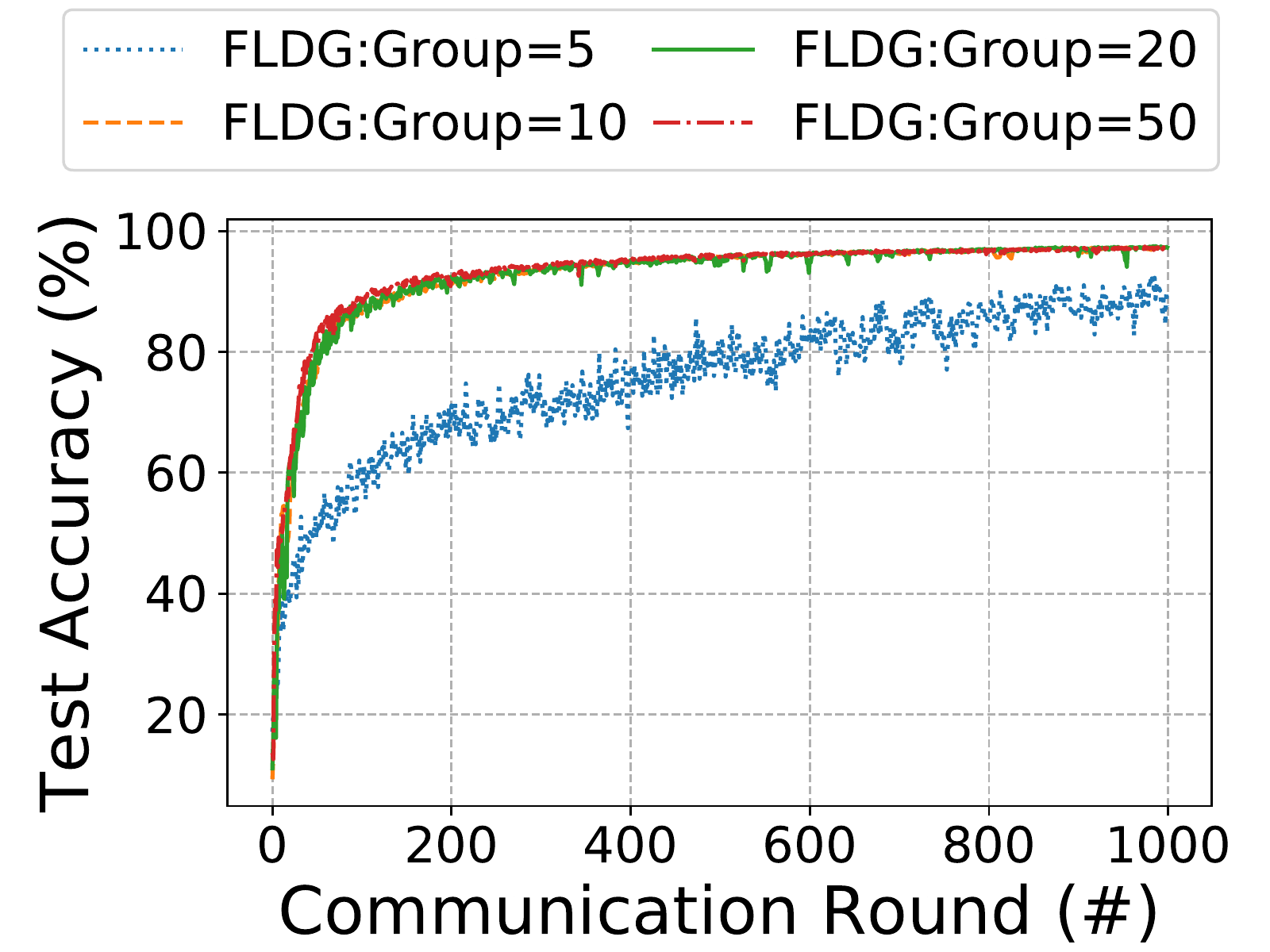}}
\subfigure[Case 2 of MNIST]{
\includegraphics[width=0.23\linewidth]{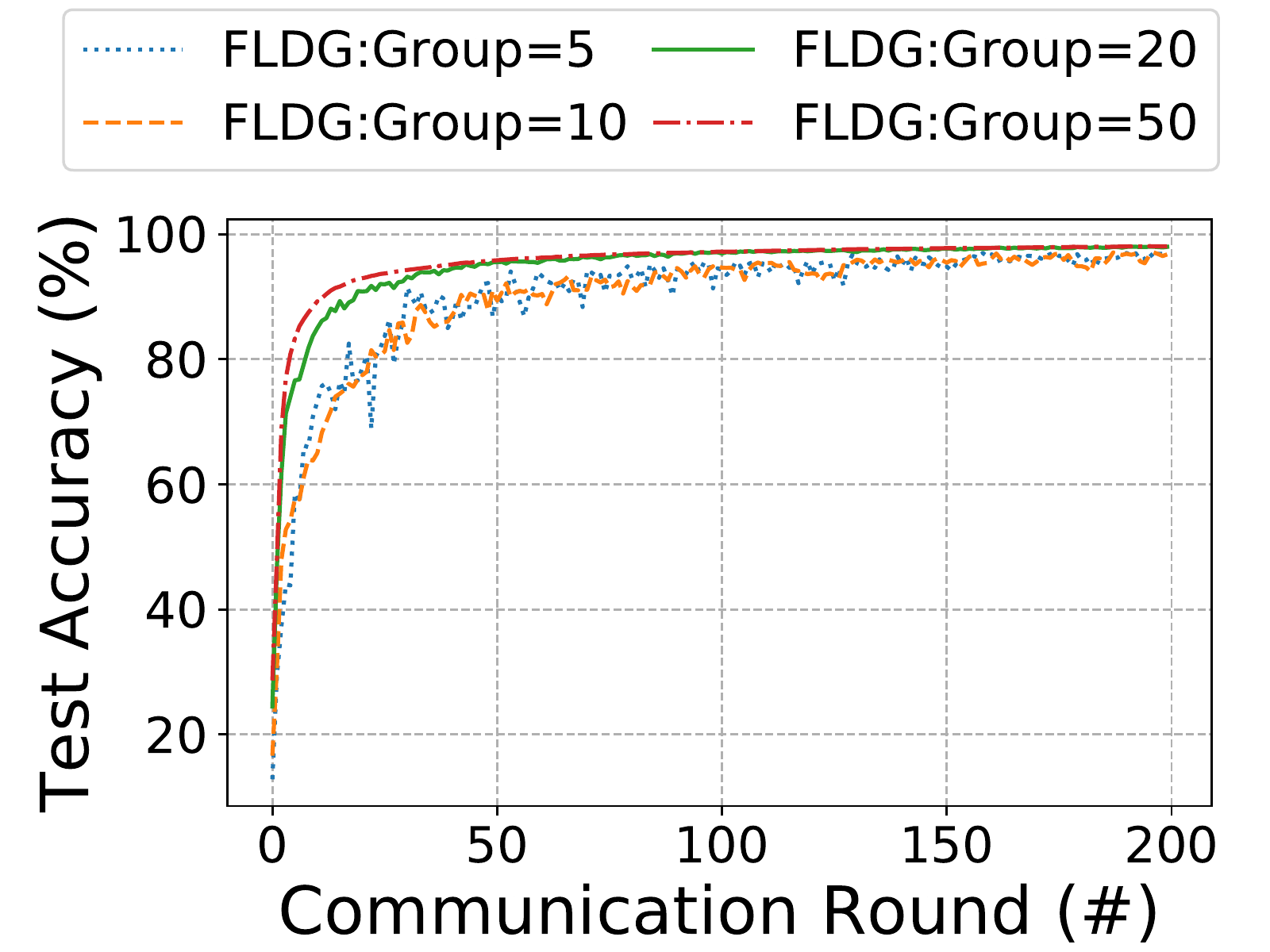}}
\subfigure[Case 3 of MNIST]{
\includegraphics[width=0.23\linewidth]{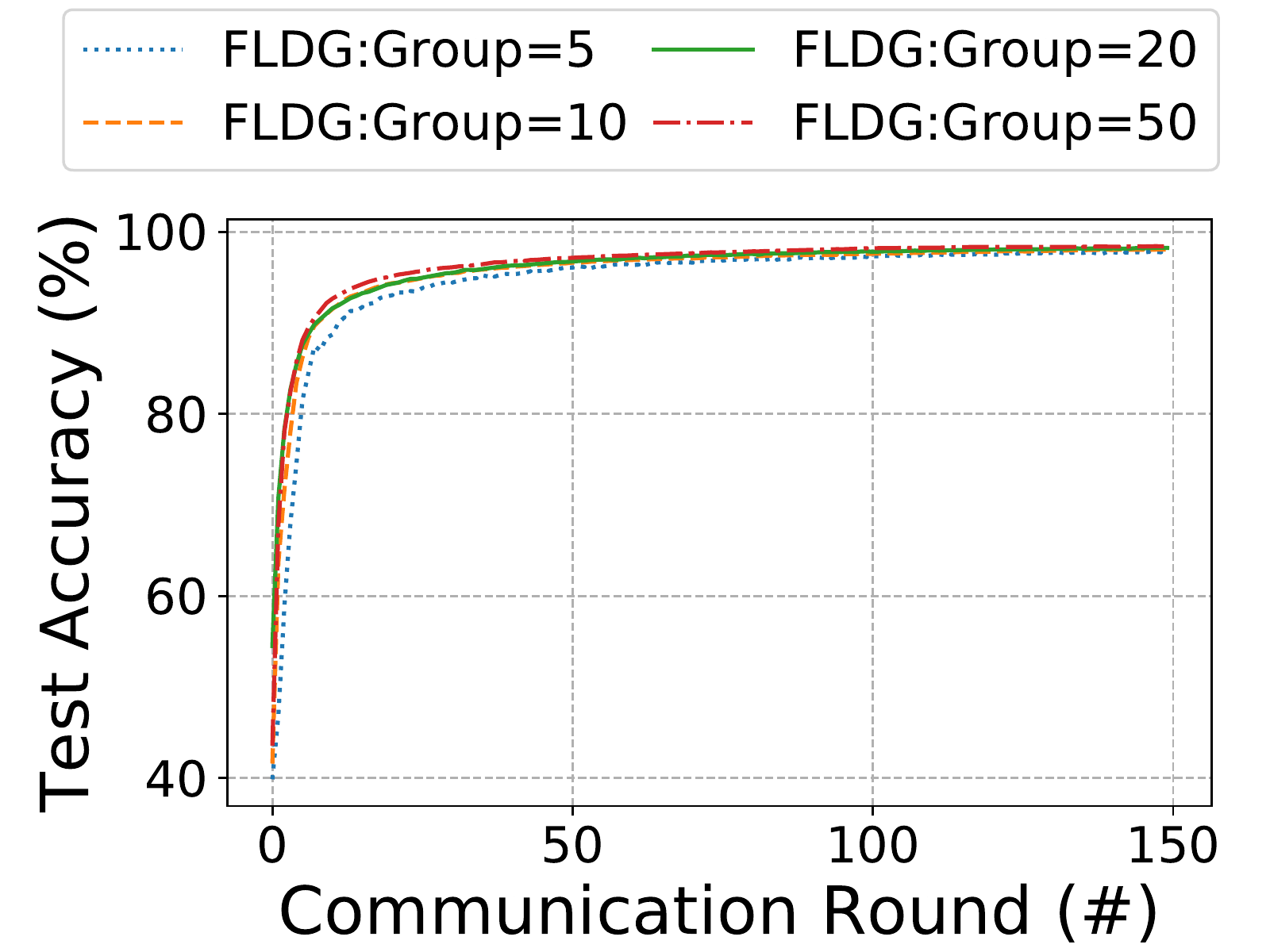}}
\subfigure[Case 4 of MNIST]{
\includegraphics[width=0.23\linewidth]{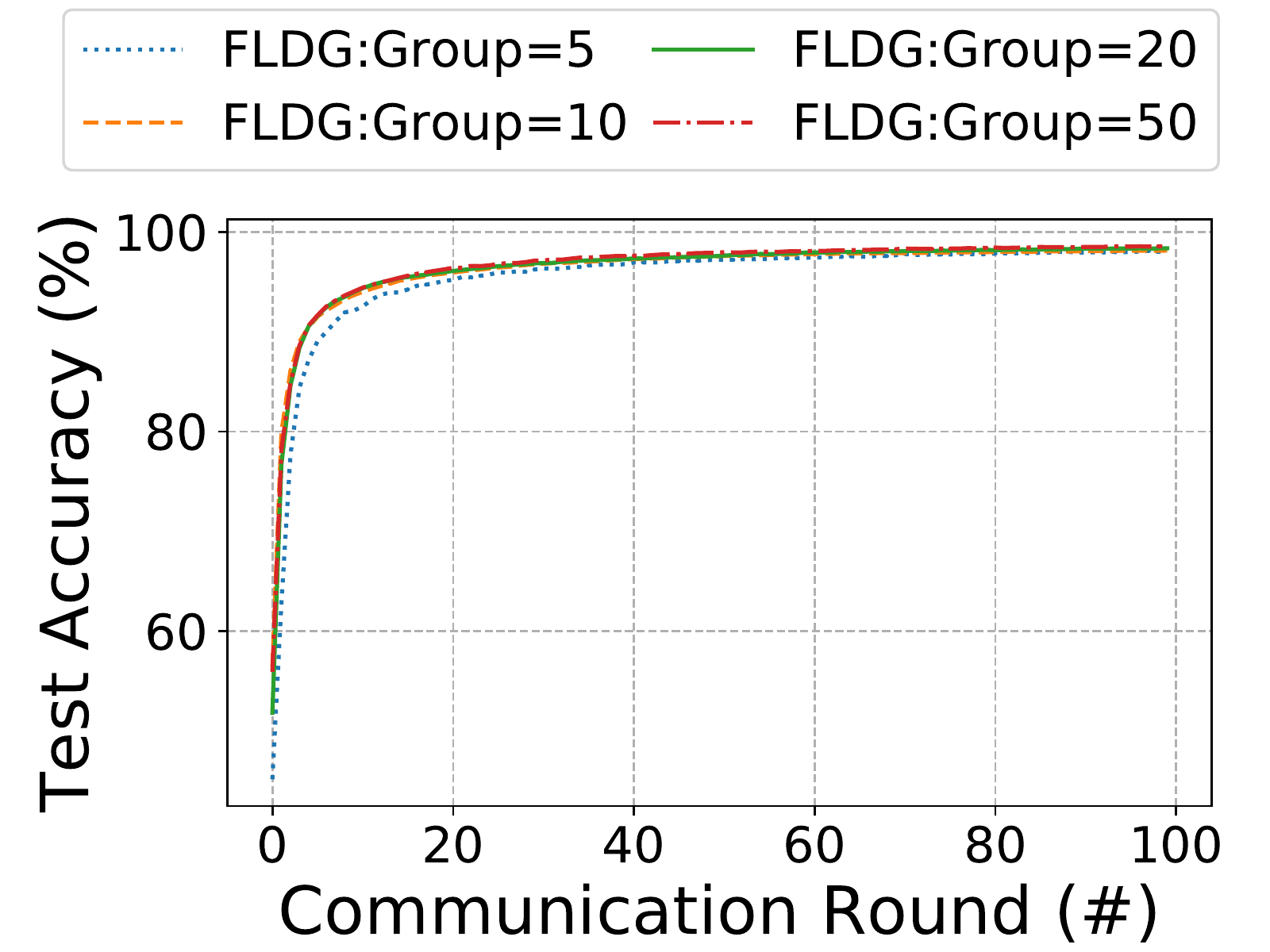}}
\vspace{-0.1in}
\subfigure[Case 1 of Fashion-MNIST]{
\includegraphics[width=0.23\linewidth]{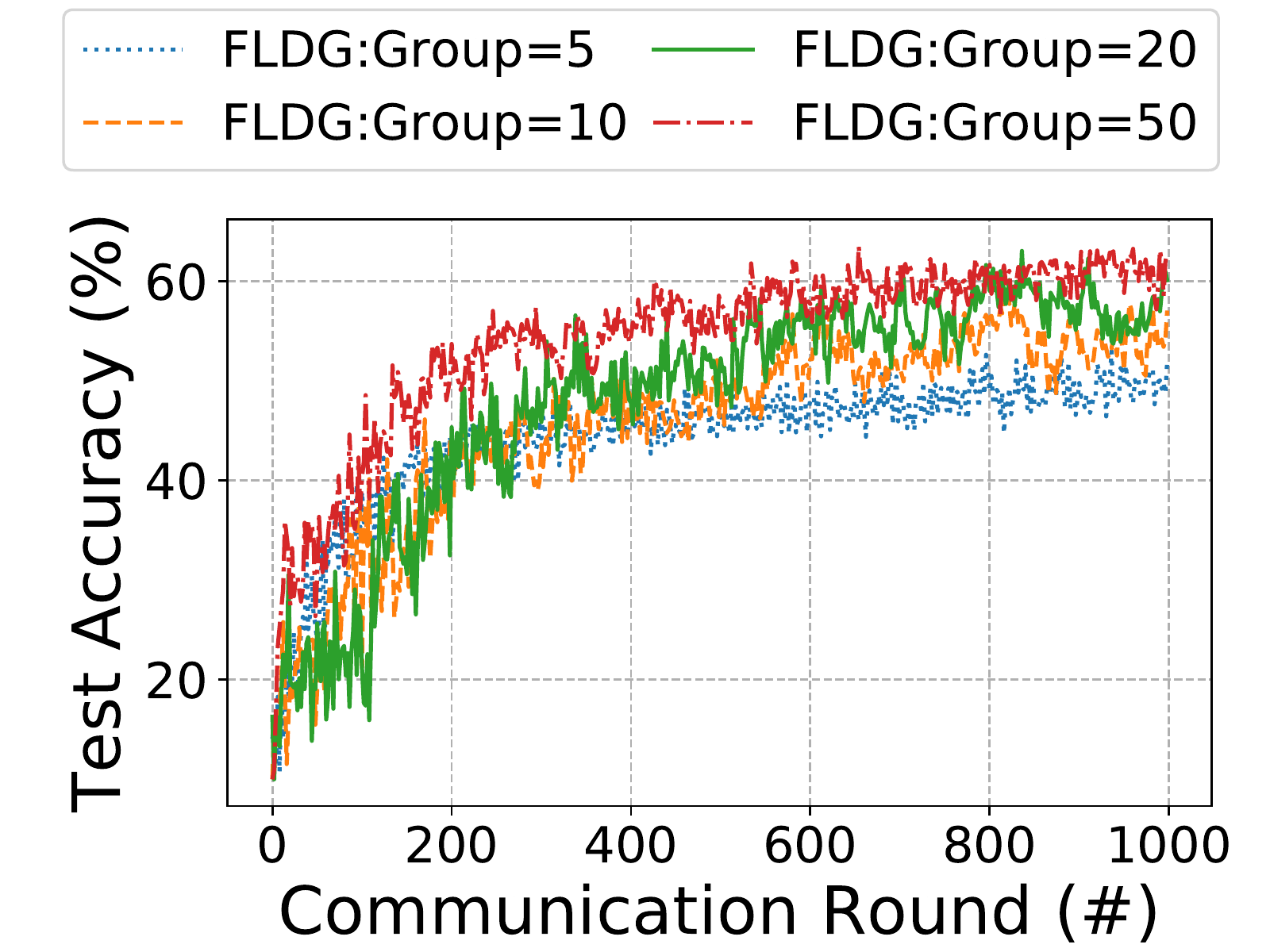}}
\subfigure[Case 2 of Fashion-MNIST]{
\includegraphics[width=0.23\linewidth]{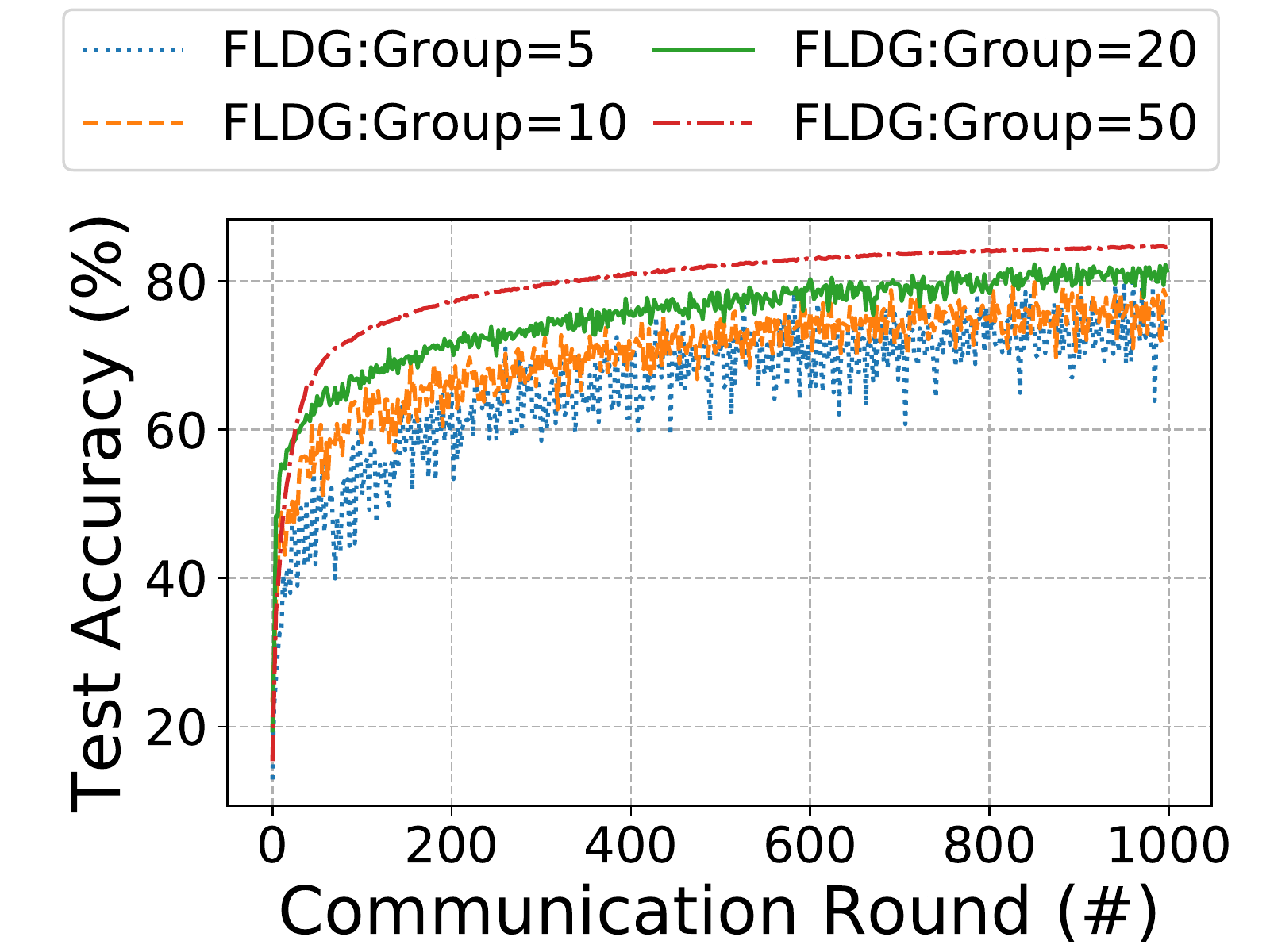}}
\subfigure[Case 3 of Fashion-MNIST]{
\includegraphics[width=0.23\linewidth]{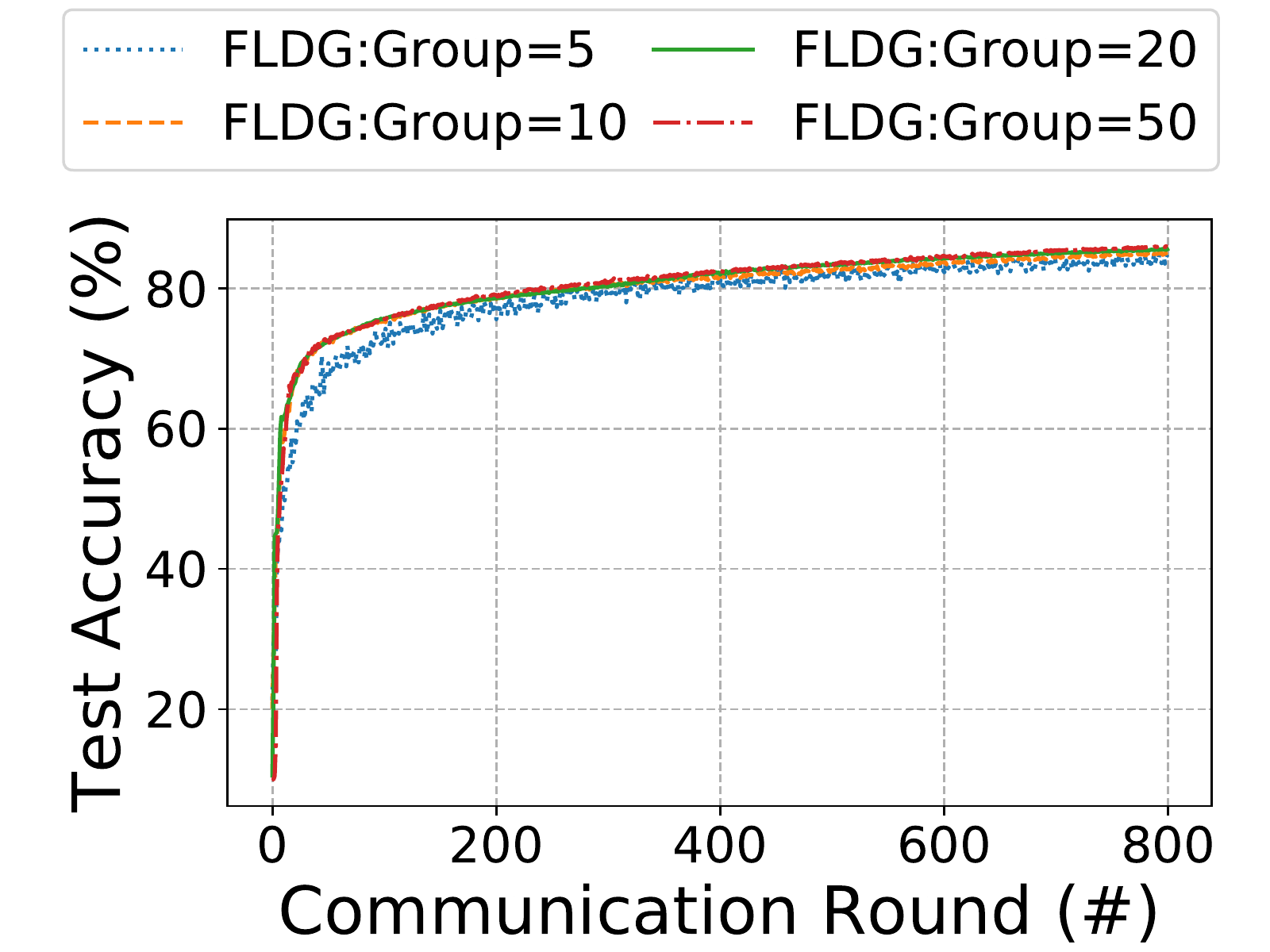}}
\subfigure[Case 4 of Fashion-MNIST]{
\includegraphics[width=0.23\linewidth]{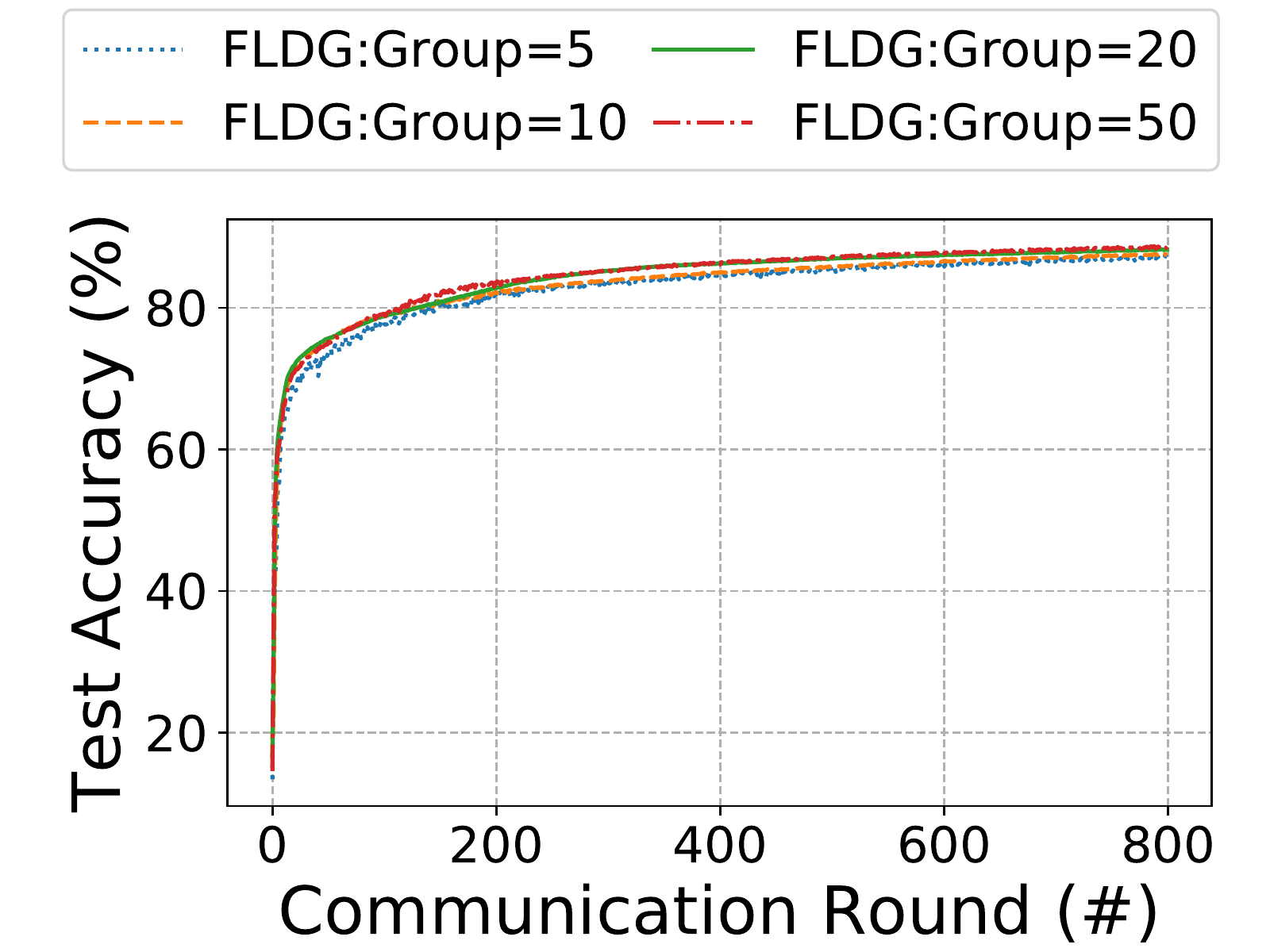}}
\vspace{-0.1in}
\subfigure[Case 1 of CIFAR-10]{
\includegraphics[width=0.23\linewidth]{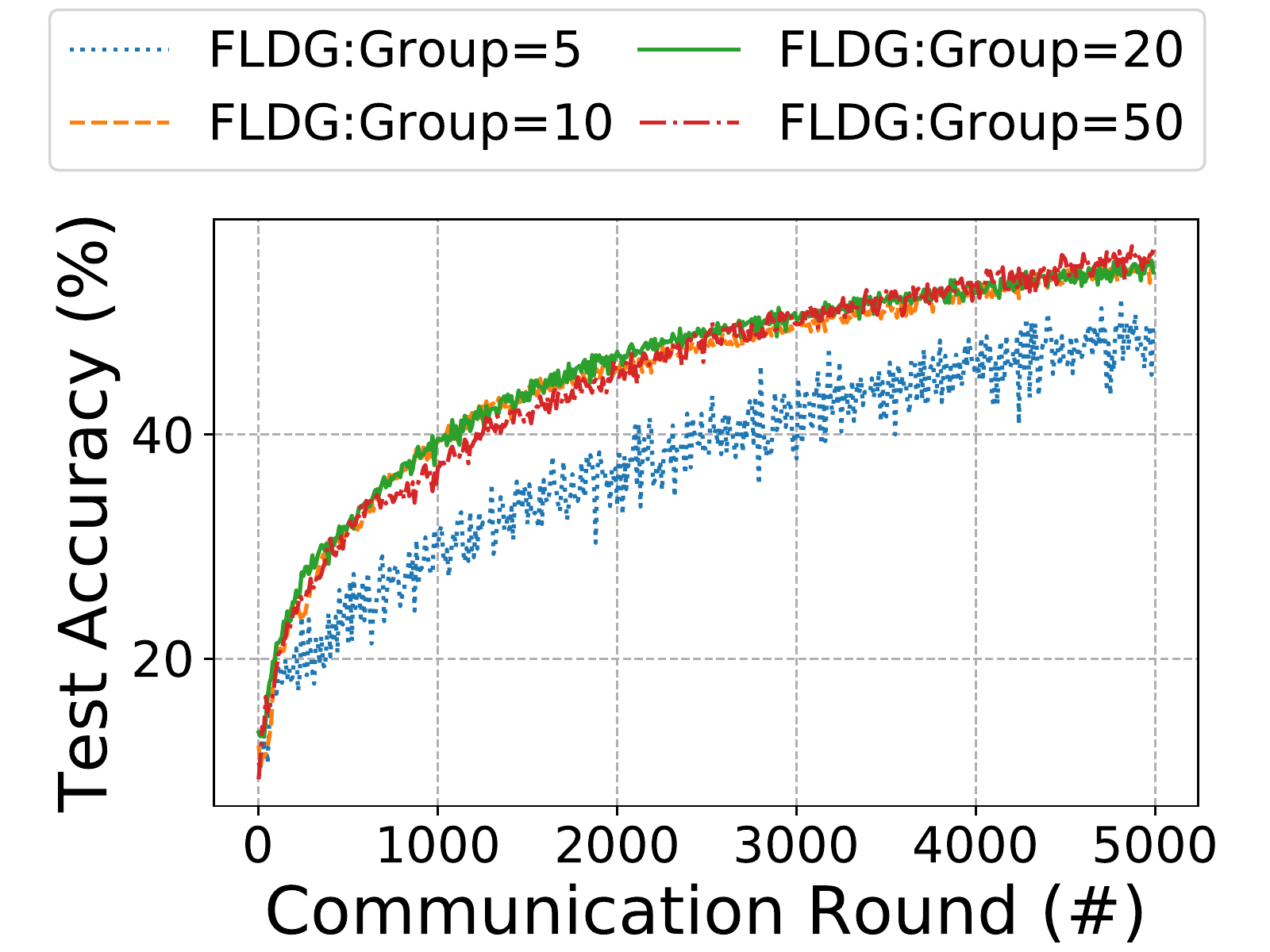}}
\subfigure[Case 2 of CIFAR-10]{
\includegraphics[width=0.23\linewidth]{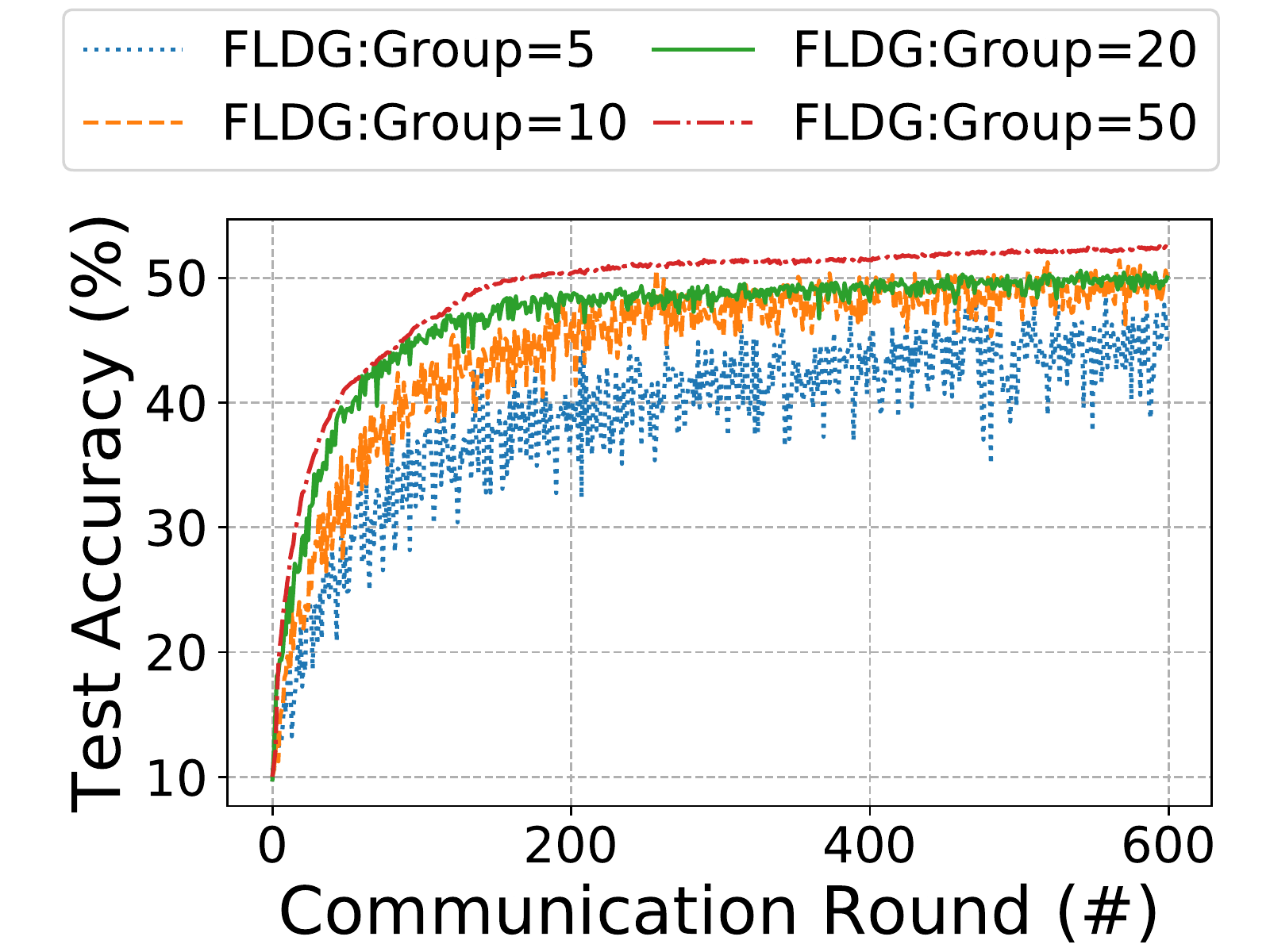}}
\subfigure[Case 3 of CIFAR-10]{
\includegraphics[width=0.23\linewidth]{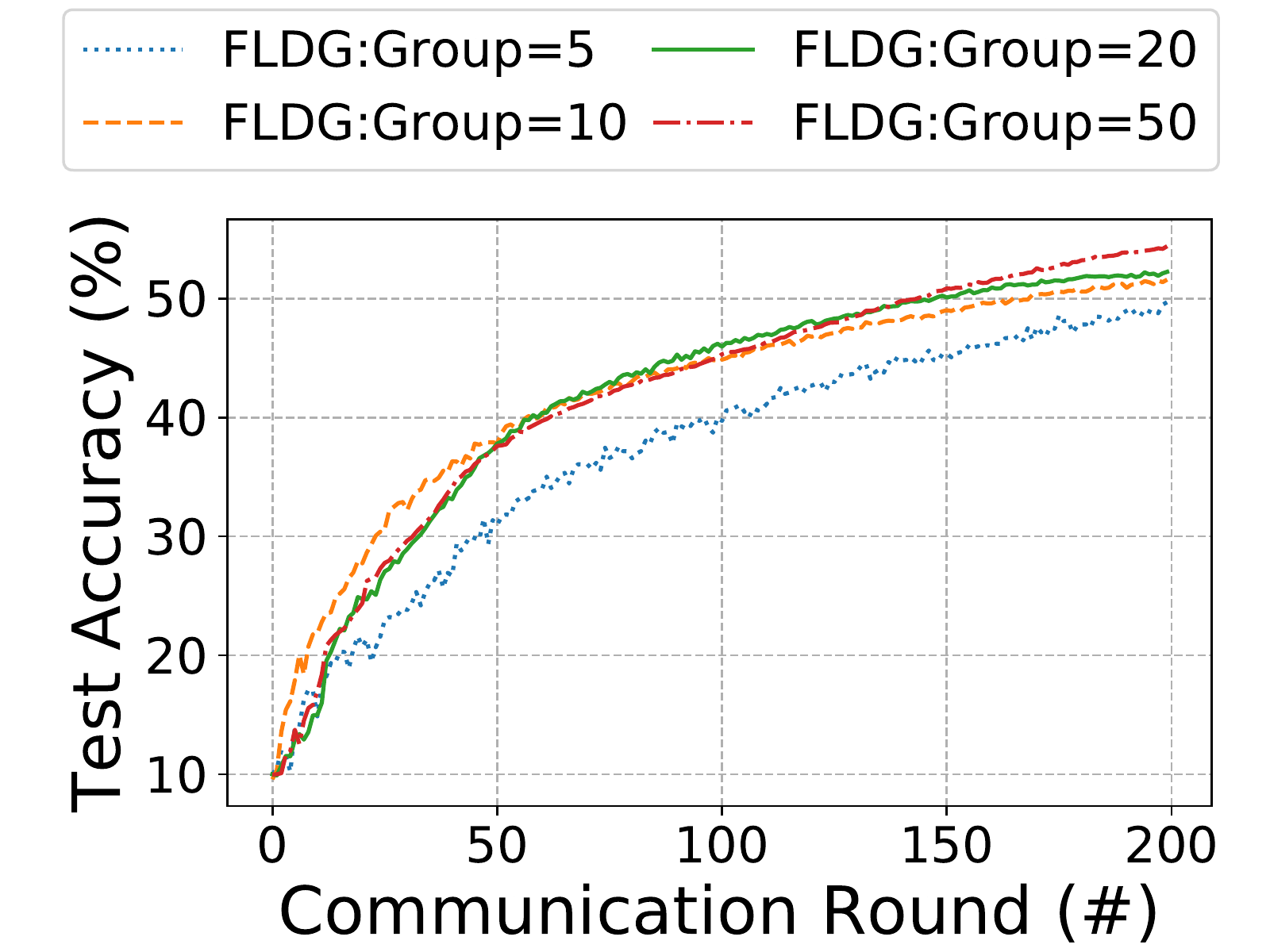}}
\subfigure[Case 4 of CIFAR-10]{
\includegraphics[width=0.23\linewidth]{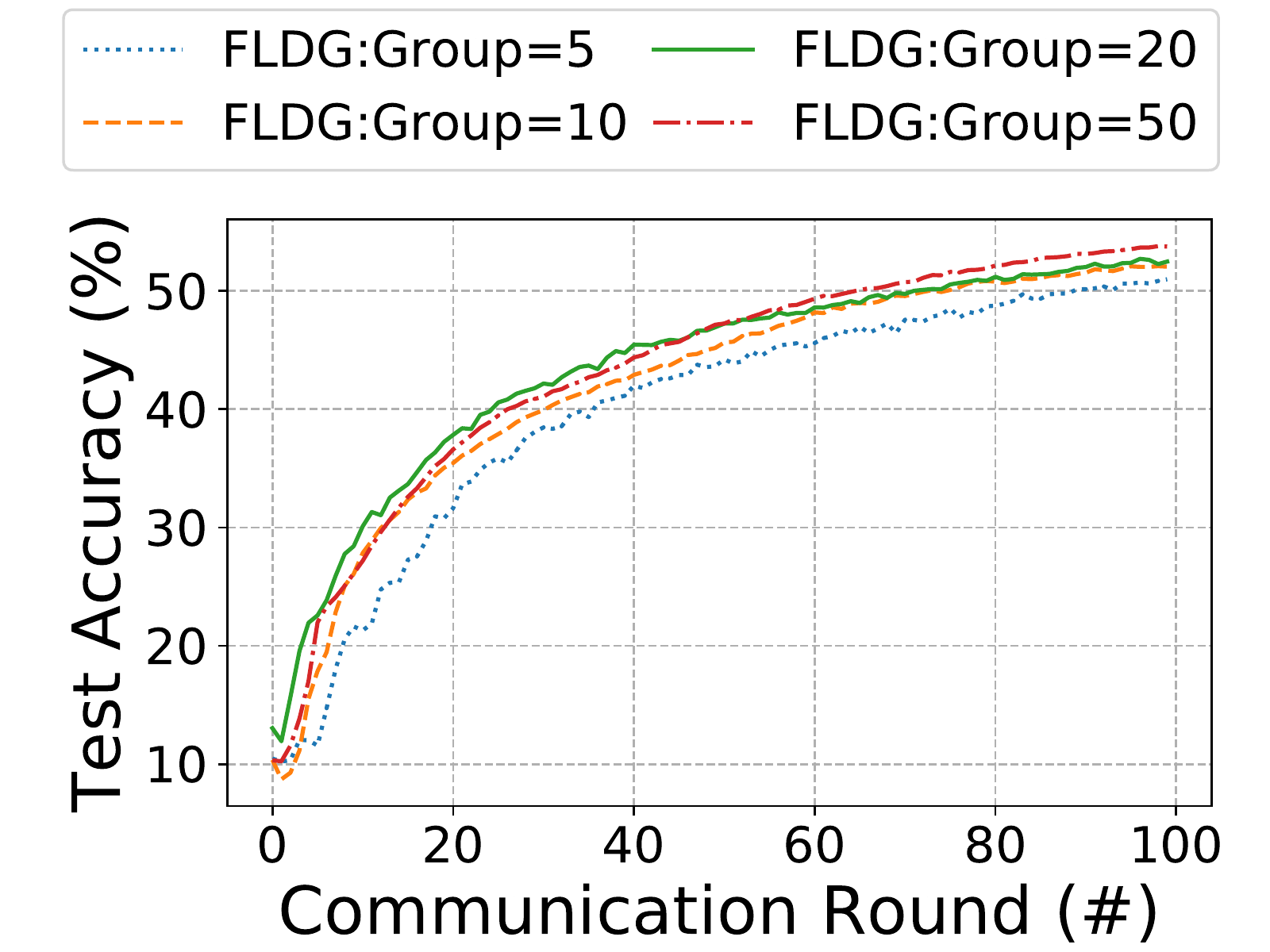}}
\caption{Test accuracy comparison w.r.t. different number of groups using FLDG}
\label{fig:num}
\vspace{-0.1in}
\end{figure*}

\begin{figure*}[h]
\centering
\subfigure[Case 1 of MNIST]{
\includegraphics[width=0.23\linewidth]{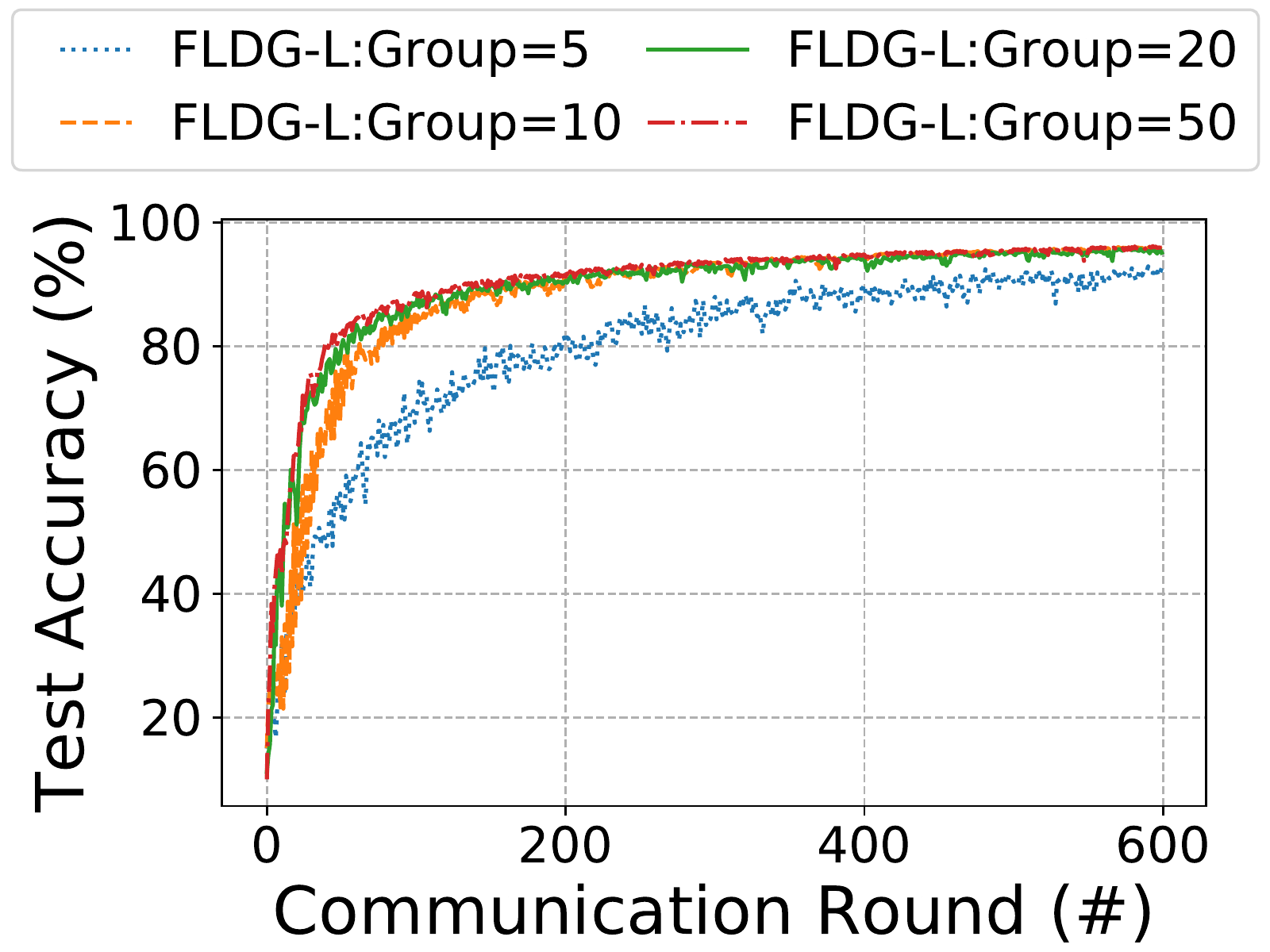}}
\subfigure[Case 2 of MNIST]{
\includegraphics[width=0.23\linewidth]{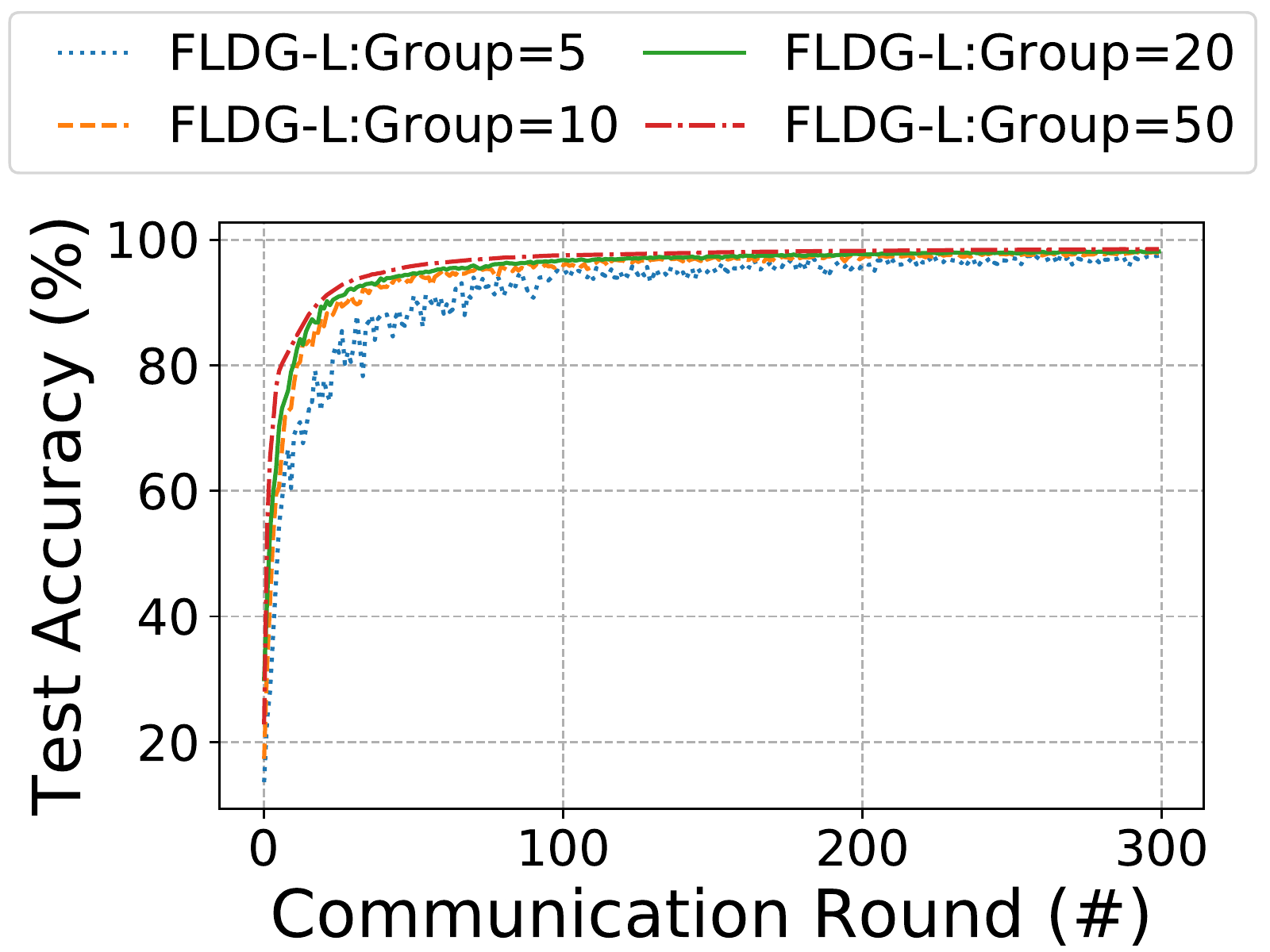}}
\subfigure[Case 3 of MNIST]{
\includegraphics[width=0.23\linewidth]{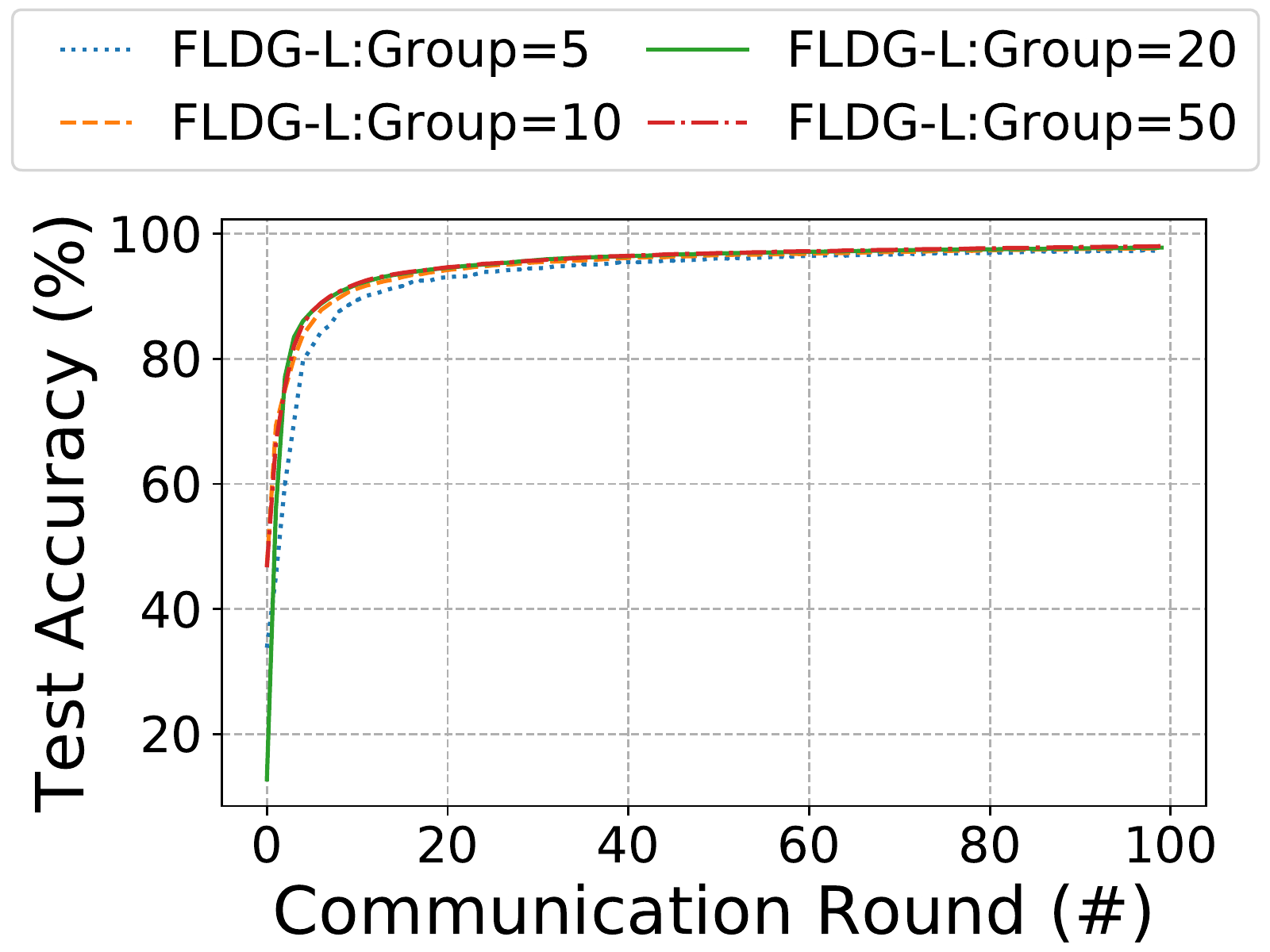}}
\subfigure[Case 4 of MNIST]{
\includegraphics[width=0.23\linewidth]{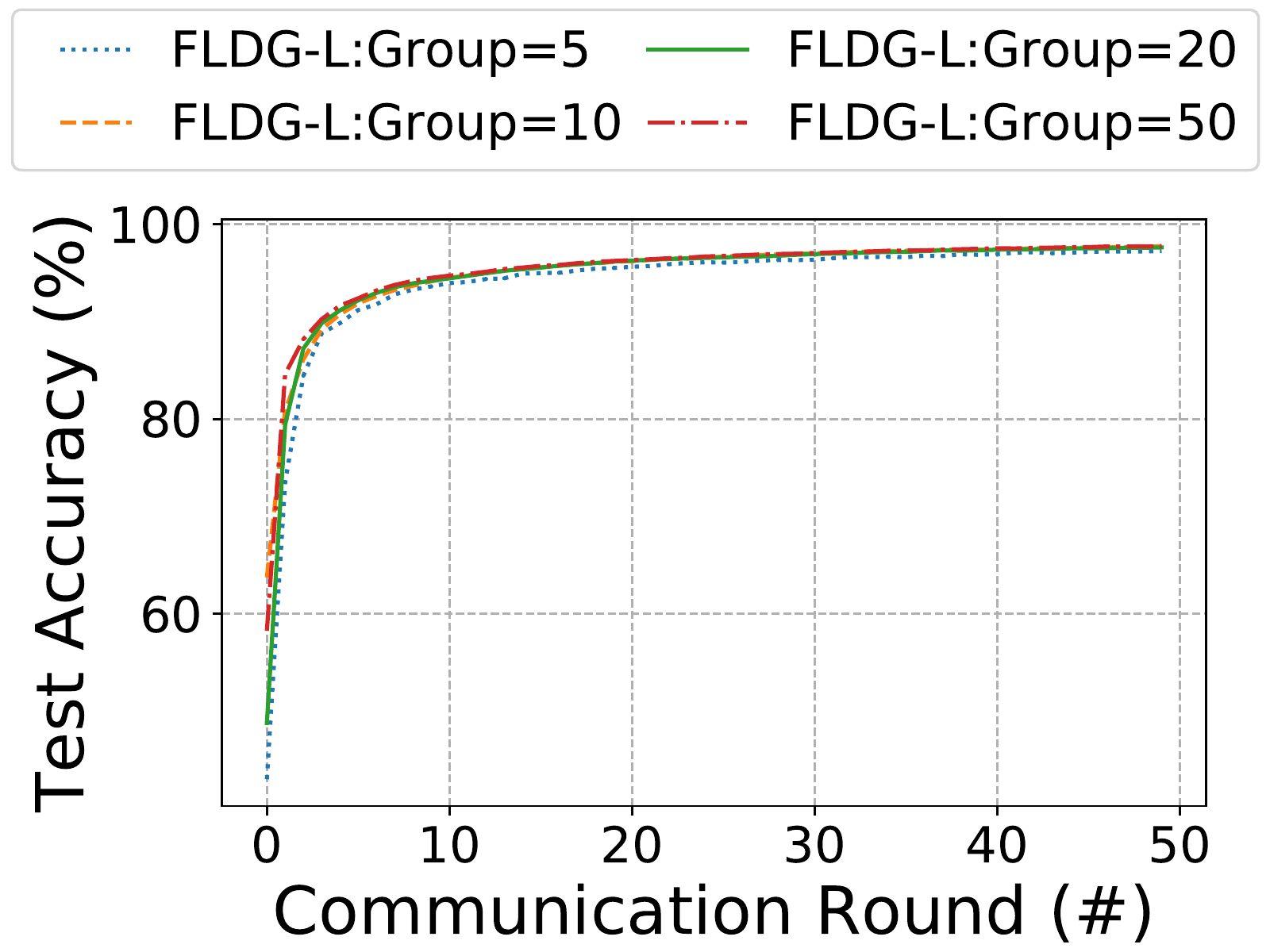}}

\subfigure[Case 1 of Fashion-MNIST]{
\includegraphics[width=0.23\linewidth]{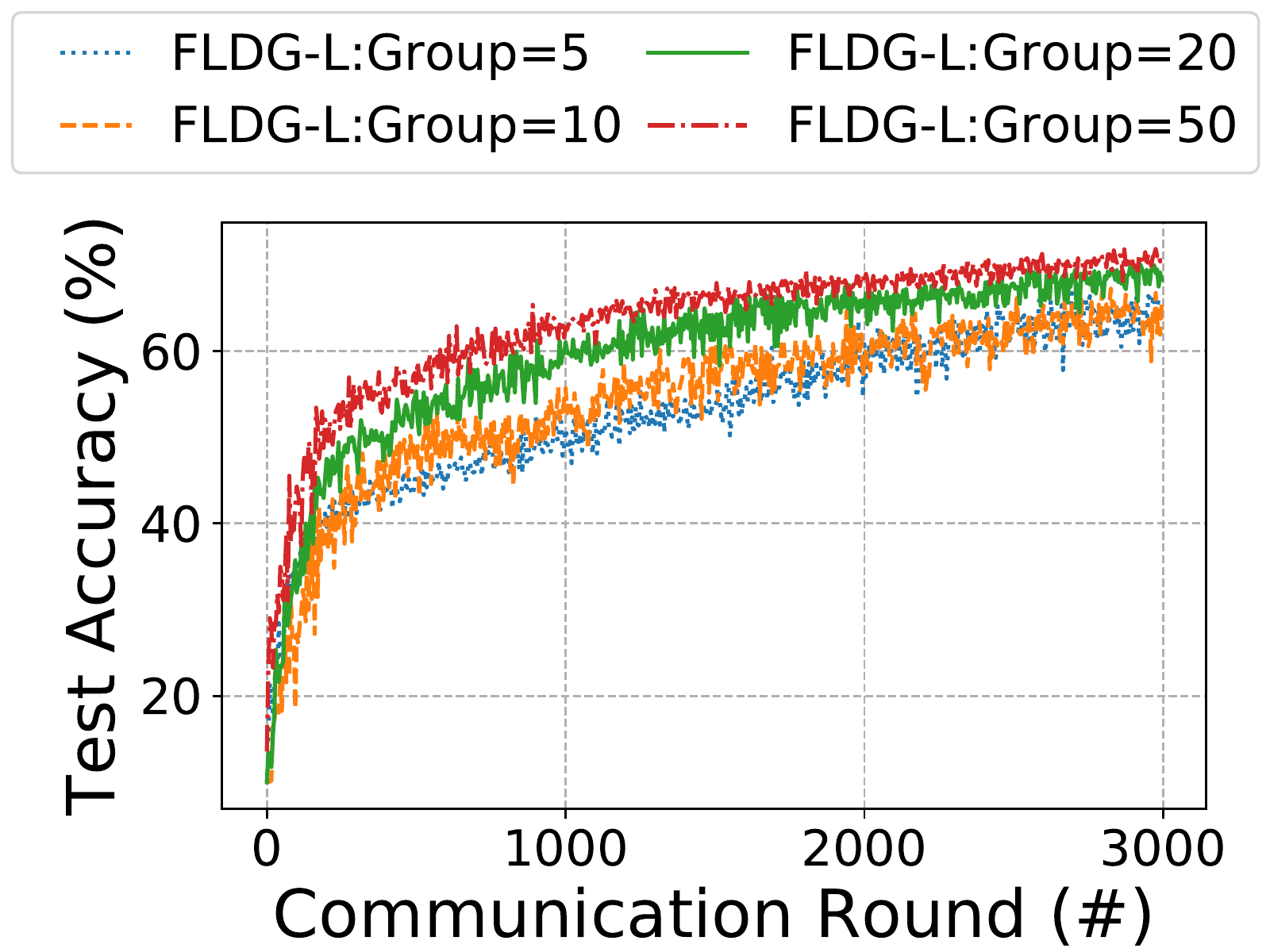}}
\subfigure[Case 2 of Fashion-MNIST]{
\includegraphics[width=0.23\linewidth]{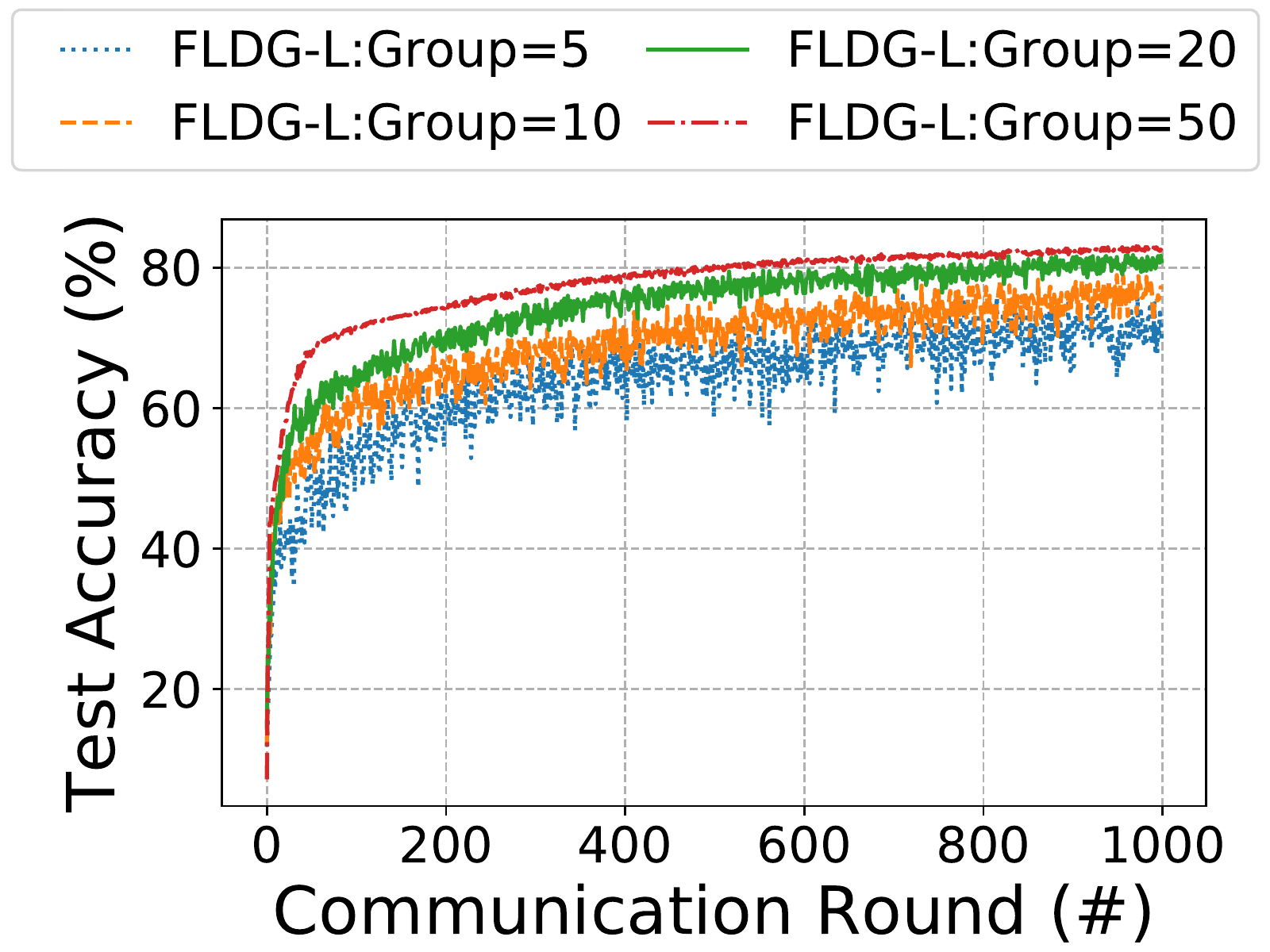}}
\subfigure[Case 3 of Fashion-MNIST]{
\includegraphics[width=0.23\linewidth]{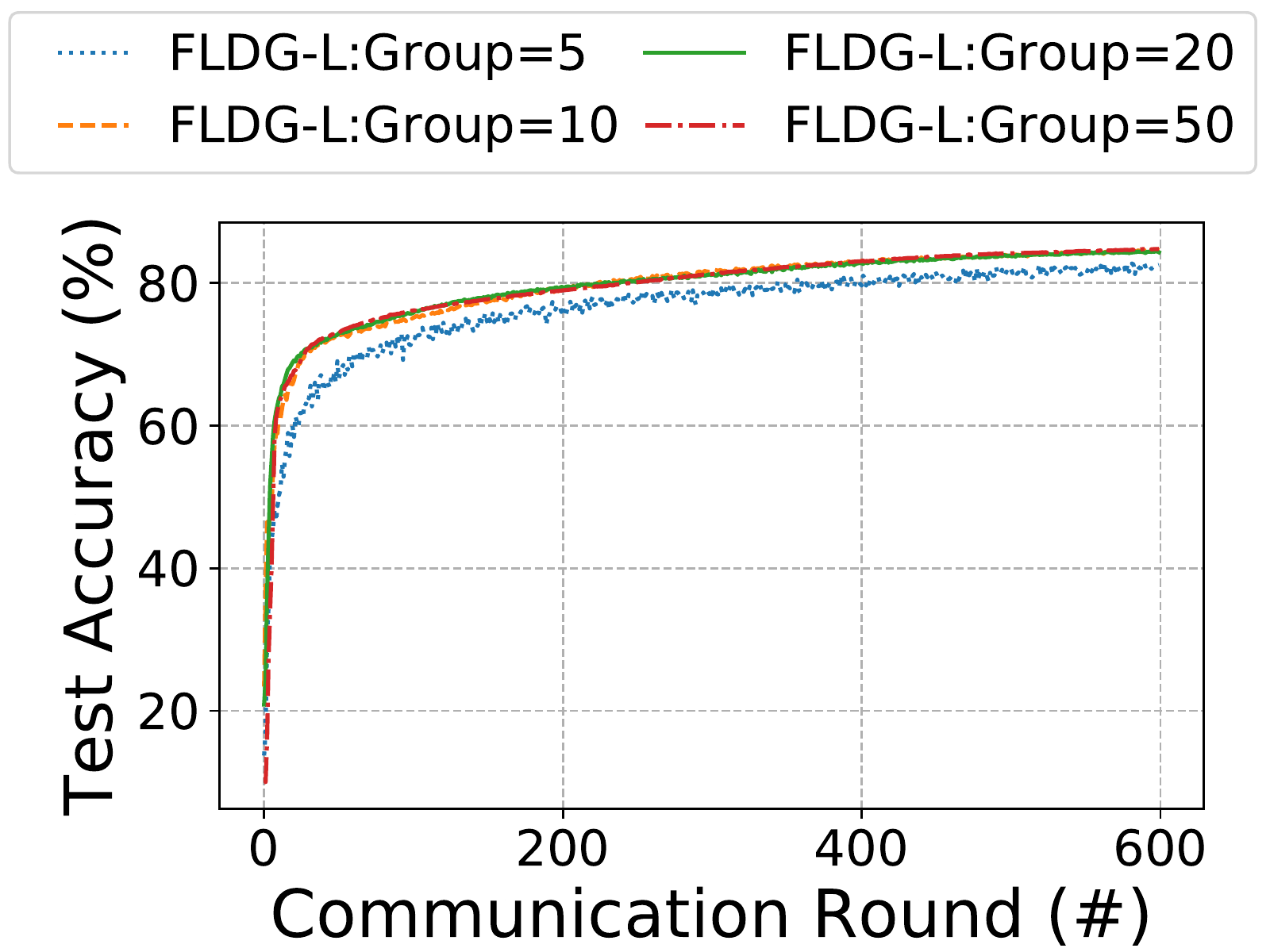}}
\subfigure[Case 4 of Fashion-MNIST]{
\includegraphics[width=0.23\linewidth]{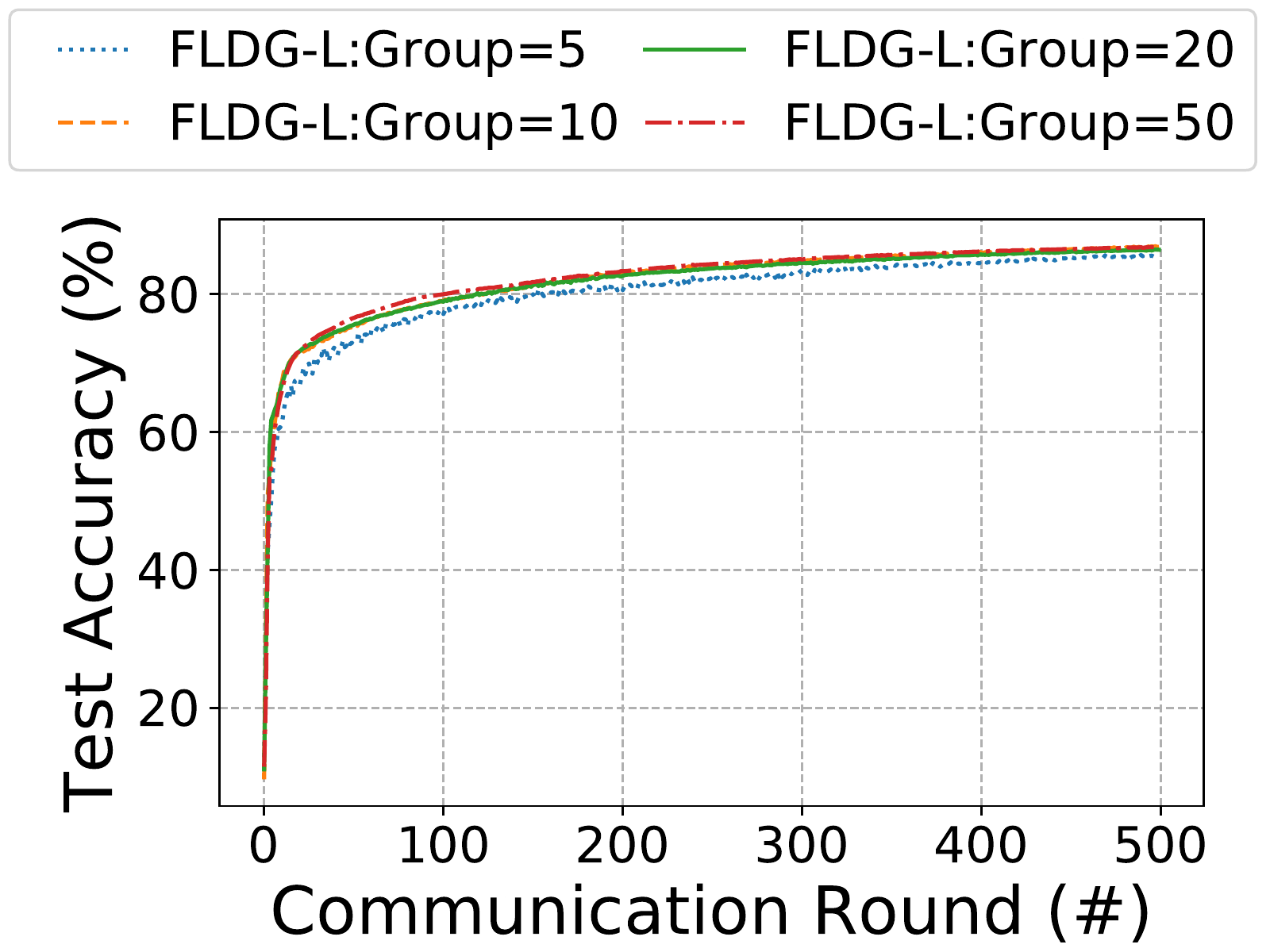}}

\subfigure[Case 1 of CIFAR-10]{
\includegraphics[width=0.23\linewidth]{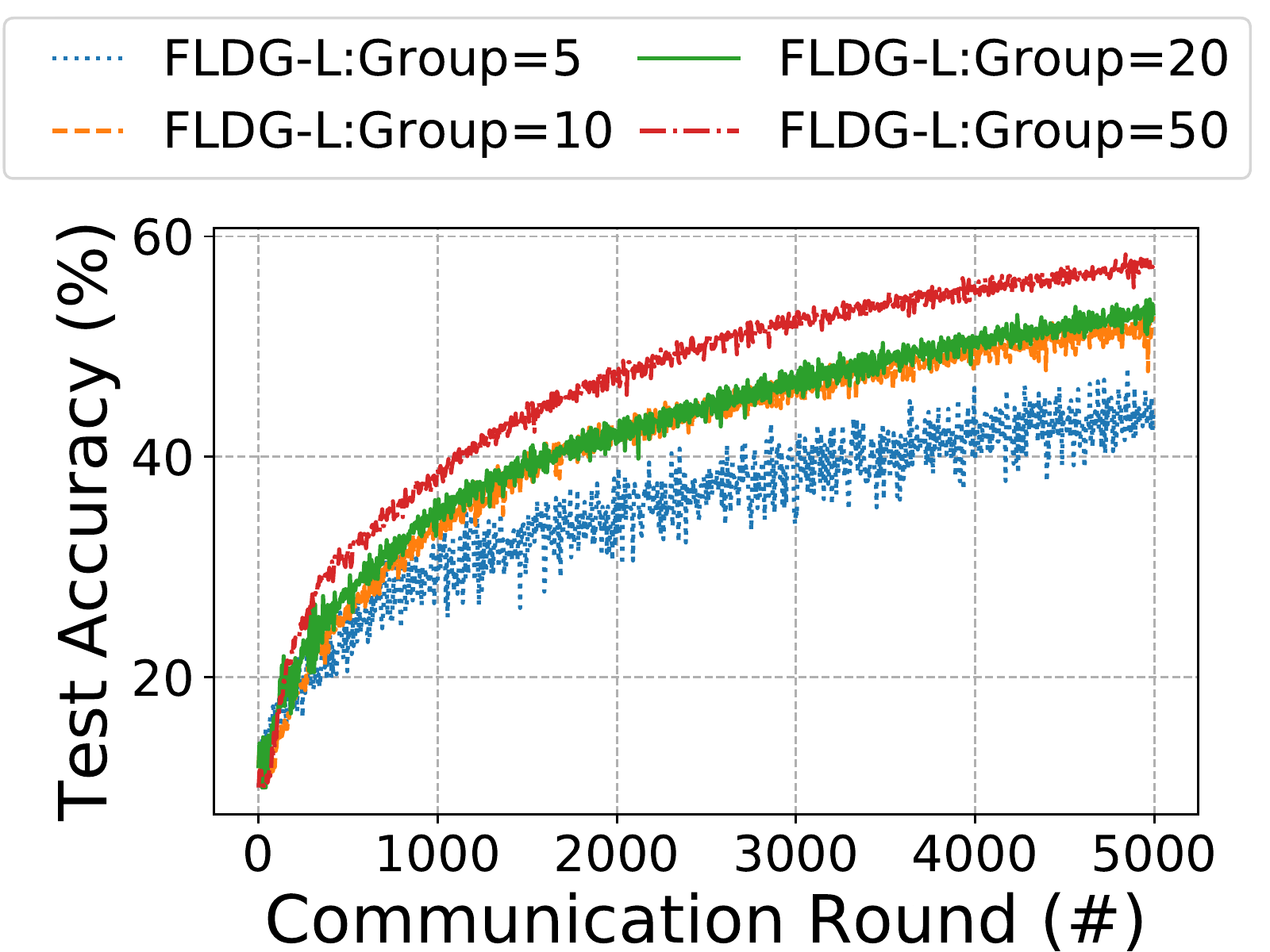}}
\subfigure[Case 2 of CIFAR-10]{
\label{fig:flnum3}
\includegraphics[width=0.23\linewidth]{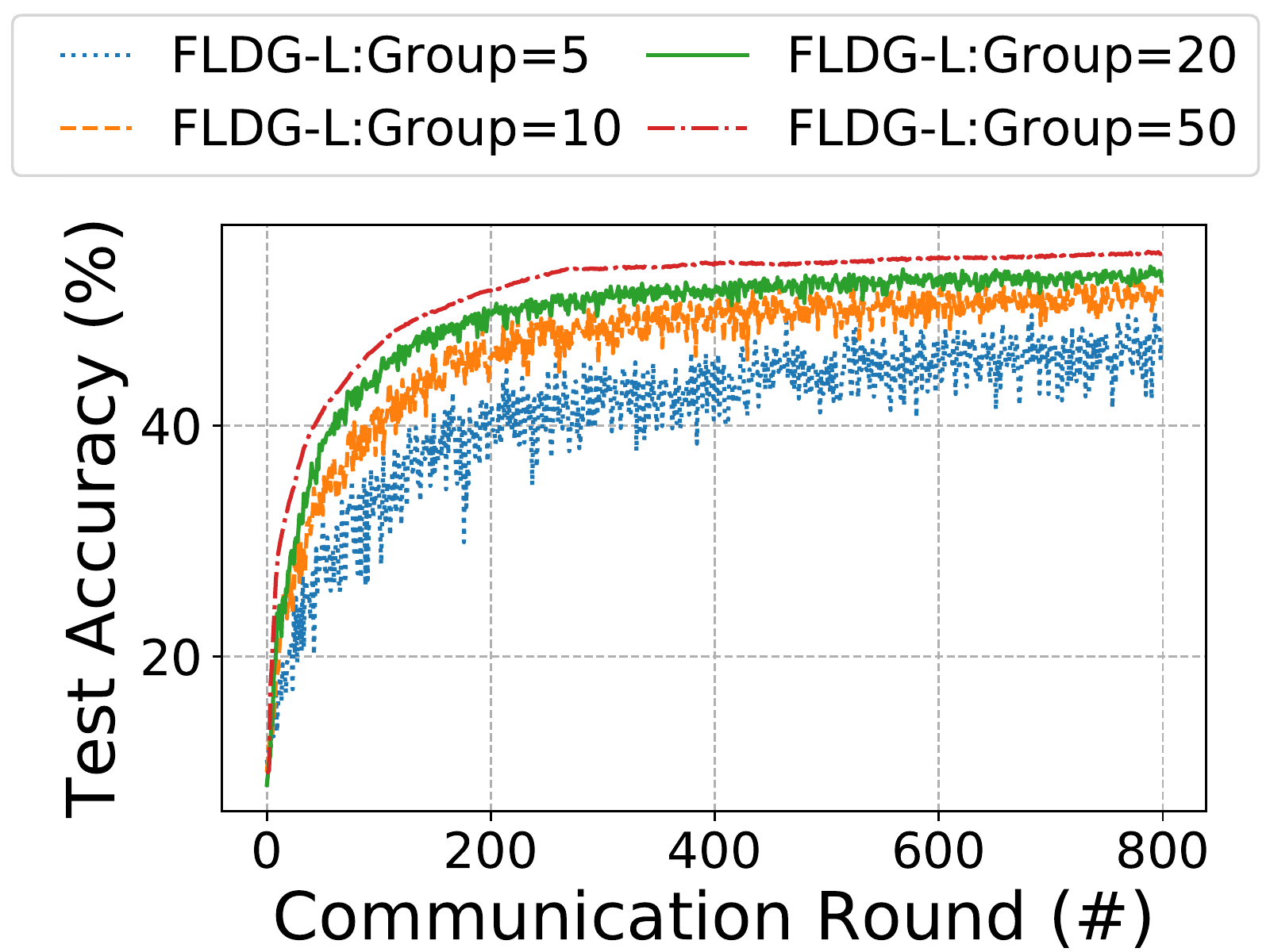}}
\subfigure[Case 3 of CIFAR-10]{
\includegraphics[width=0.23\linewidth]{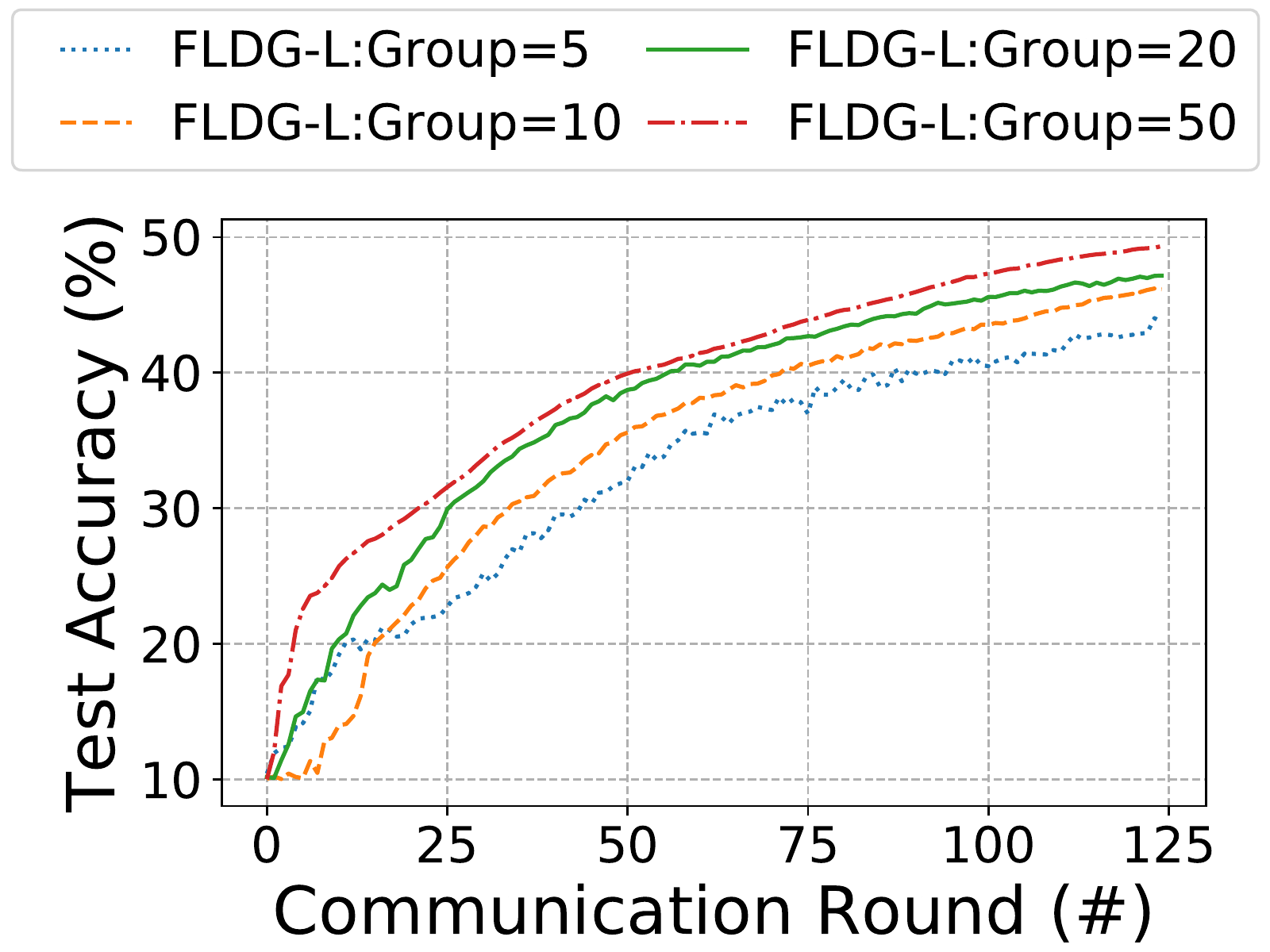}}
\subfigure[Case 4 of CIFAR-10]{
\includegraphics[width=0.23\linewidth]{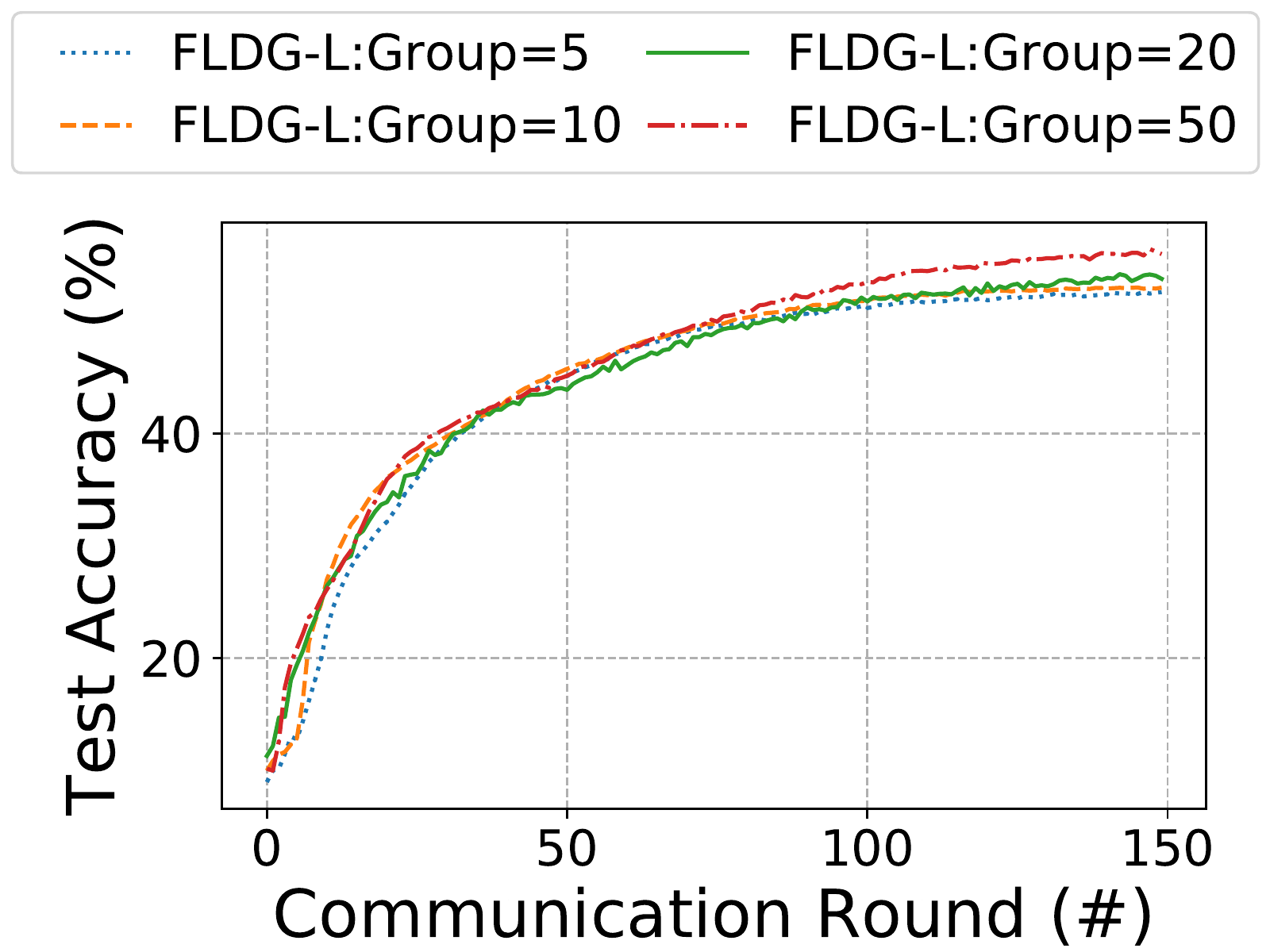}}
\vspace{-0.1in}
\caption{Test accuracy comparison w.r.t. different number of groups using FLDG-L}
\label{fig:num2}
\end{figure*}

\subsection{Impact of Output Dimension}

\begin{figure*}[h]
\centering
\subfigure[Case 1 of MNIST]{
\label{fig:mcase1} 
\includegraphics[width=0.23\linewidth]{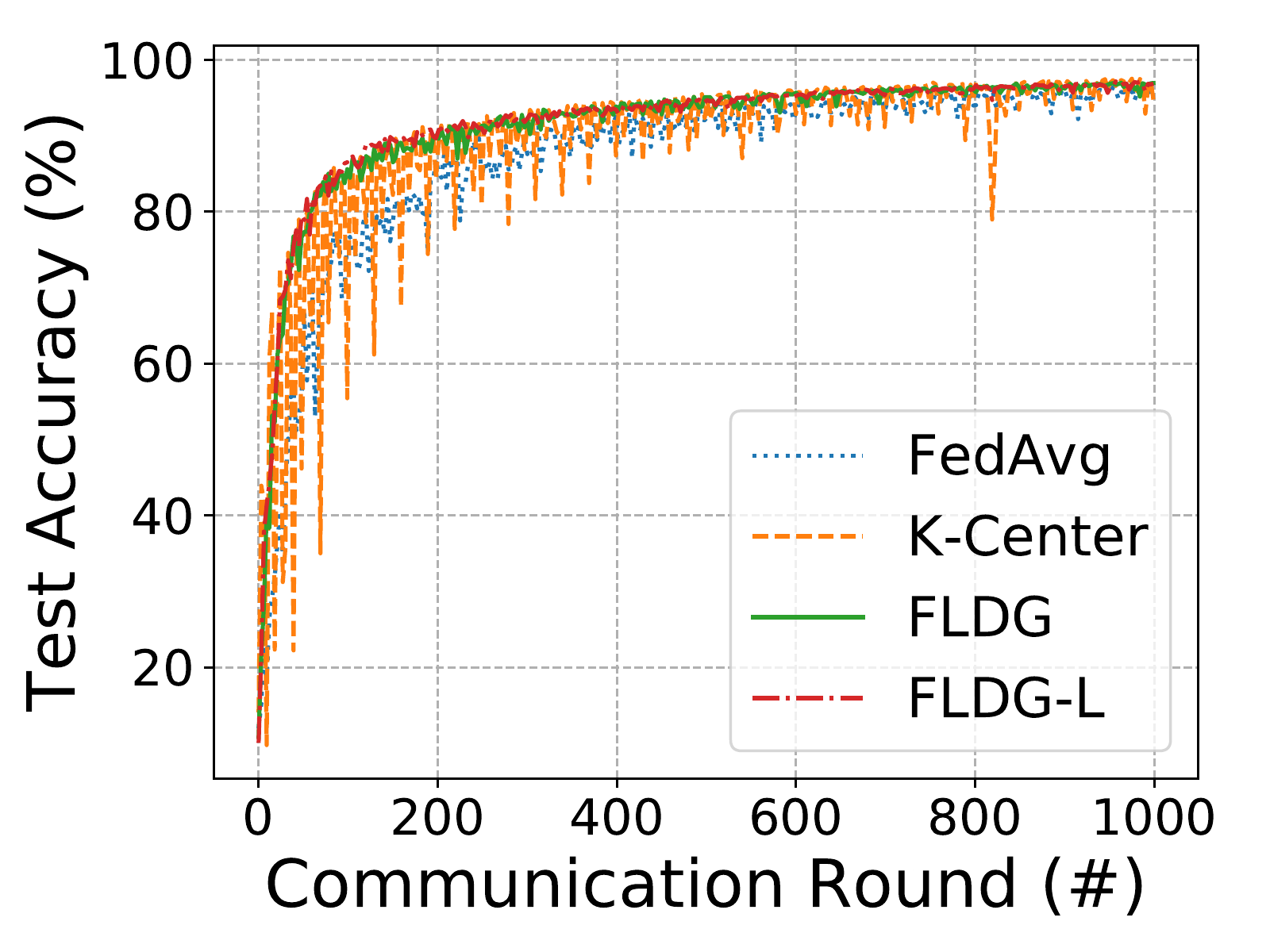}}
\subfigure[Case 2 of MNIST]{
\label{fig:mcase2}
\includegraphics[width=0.23\linewidth]{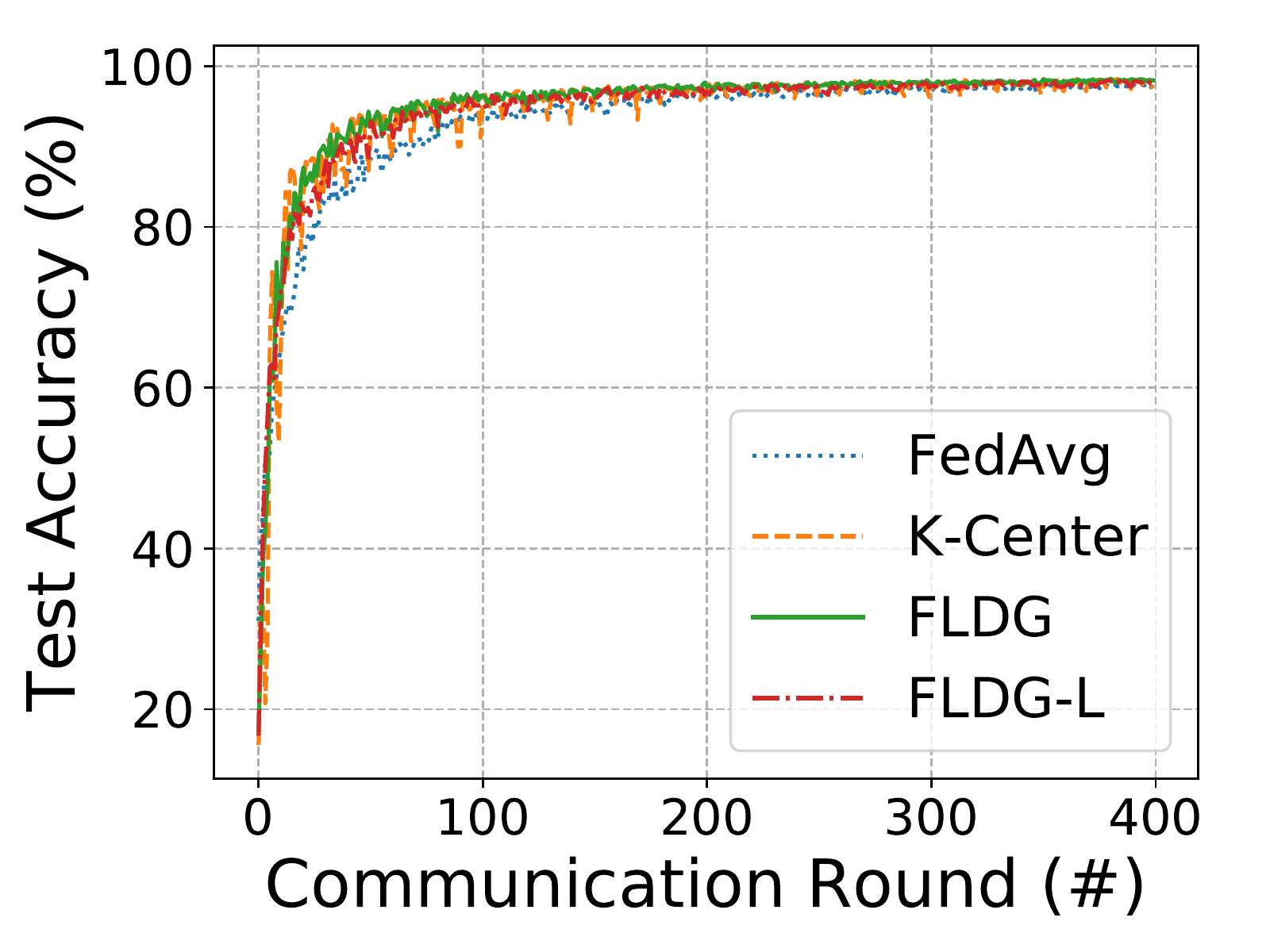}}
\subfigure[Case 3 of MNIST]{
\label{fig:mcase3} 
\includegraphics[width=0.23\linewidth]{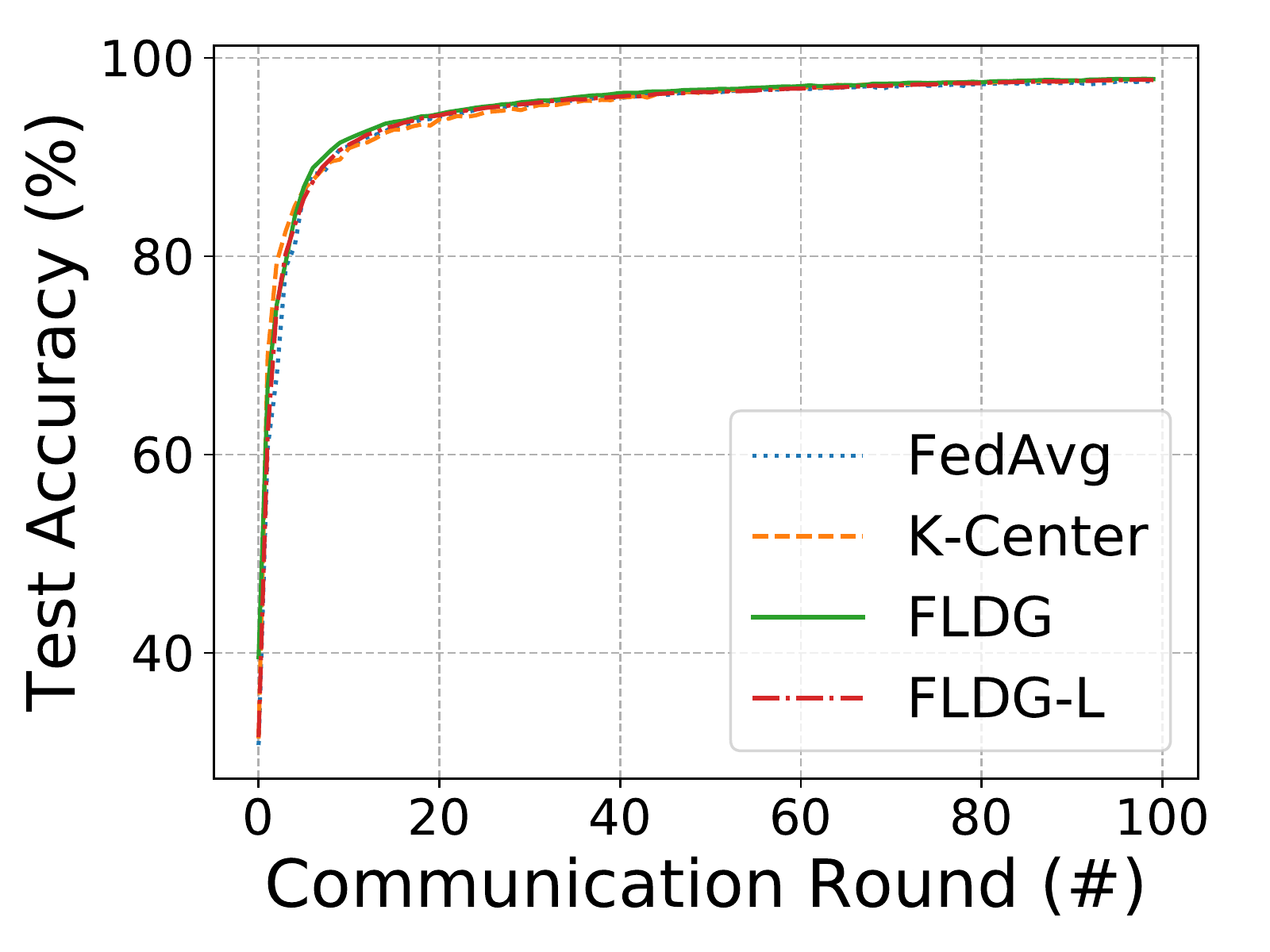}}
\subfigure[Case 4 of MNIST]{
\label{fig:mcase4}
\includegraphics[width=0.23\linewidth]{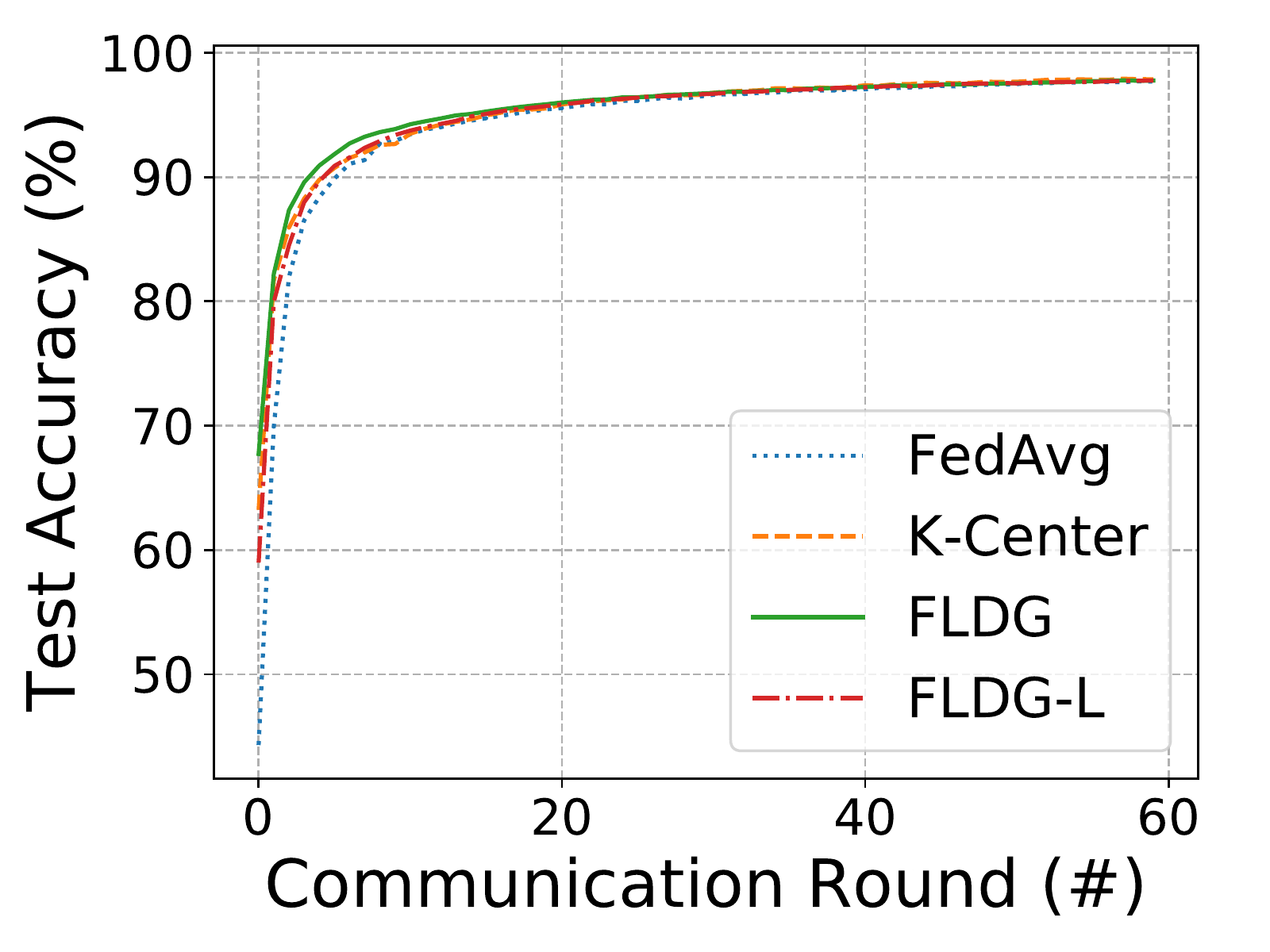}}
\vspace{-0.1in}

\subfigure[Case 1 of Fashion-MNIST]{
\label{fig:fcase1} 
\includegraphics[width=0.23\linewidth]{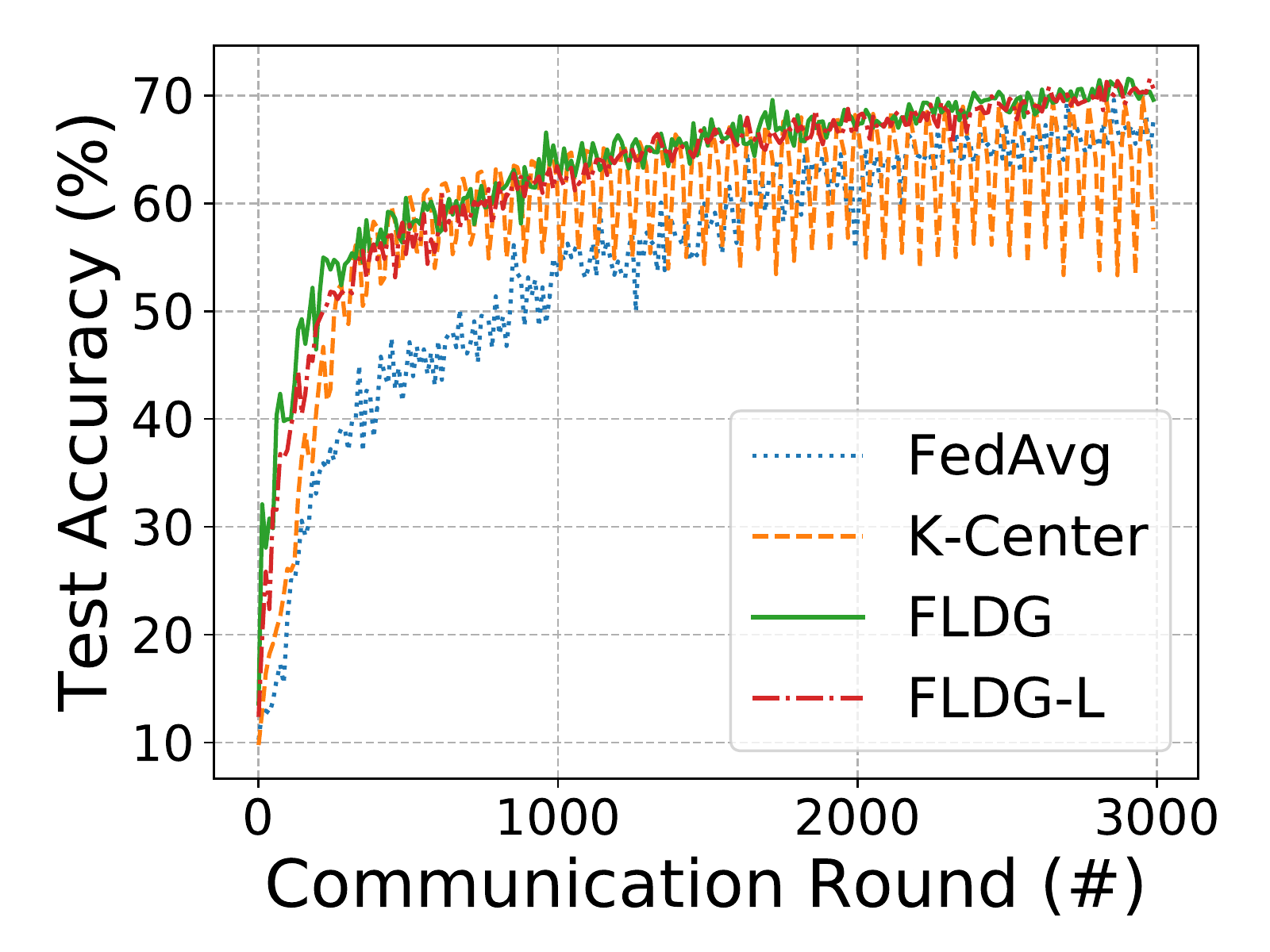}}
\subfigure[Case 2 of Fashion-MNIST]{
\label{fig:fcase2}
\includegraphics[width=0.23\linewidth]{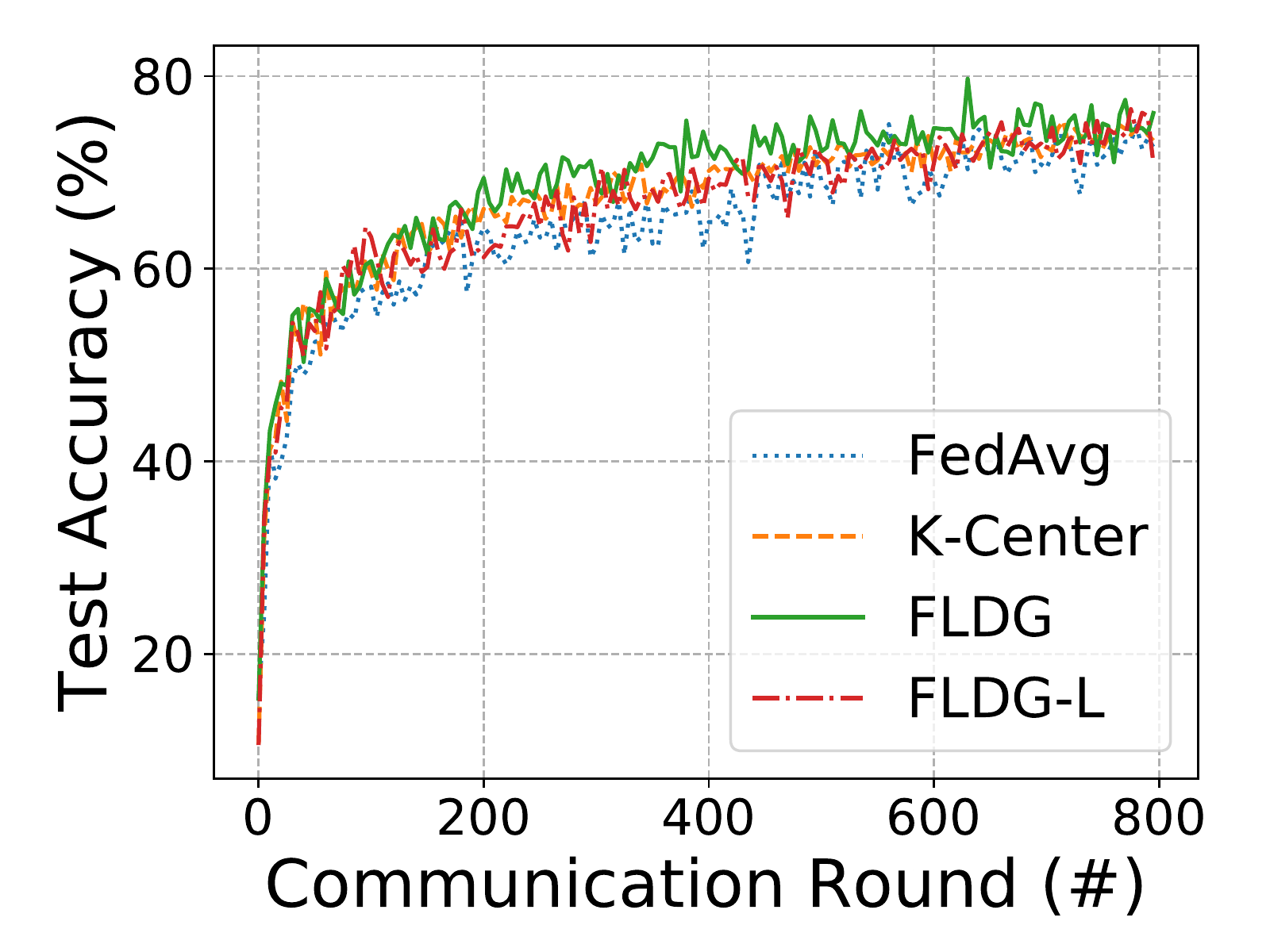}}
\subfigure[Case 3 of Fashion-MNIST]{
\label{fig:fcase3} 
\includegraphics[width=0.23\linewidth]{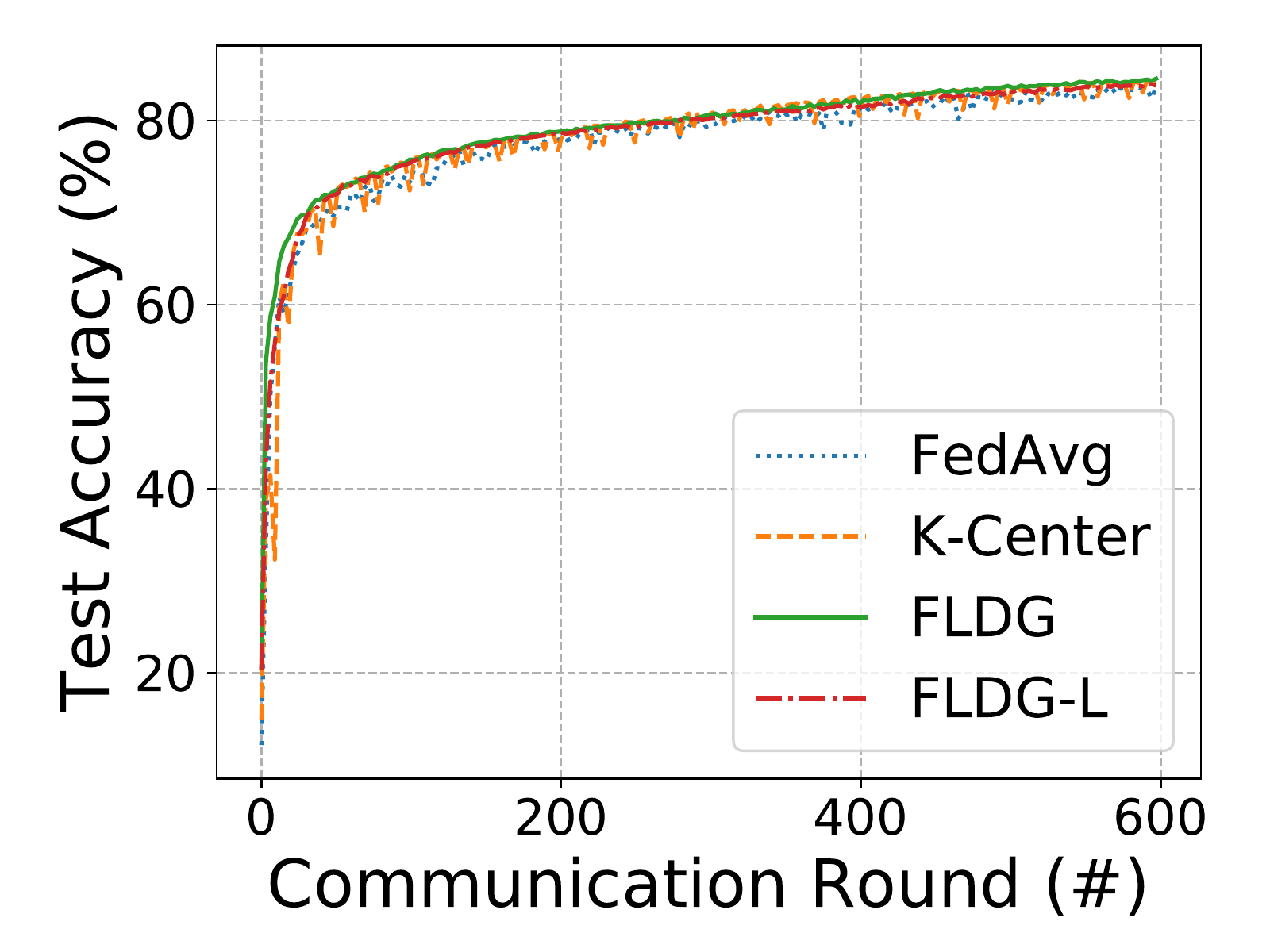}}
\subfigure[Case 4 of Fashion-MNIST]{
\label{fig:fcase4}
\includegraphics[width=0.23\linewidth]{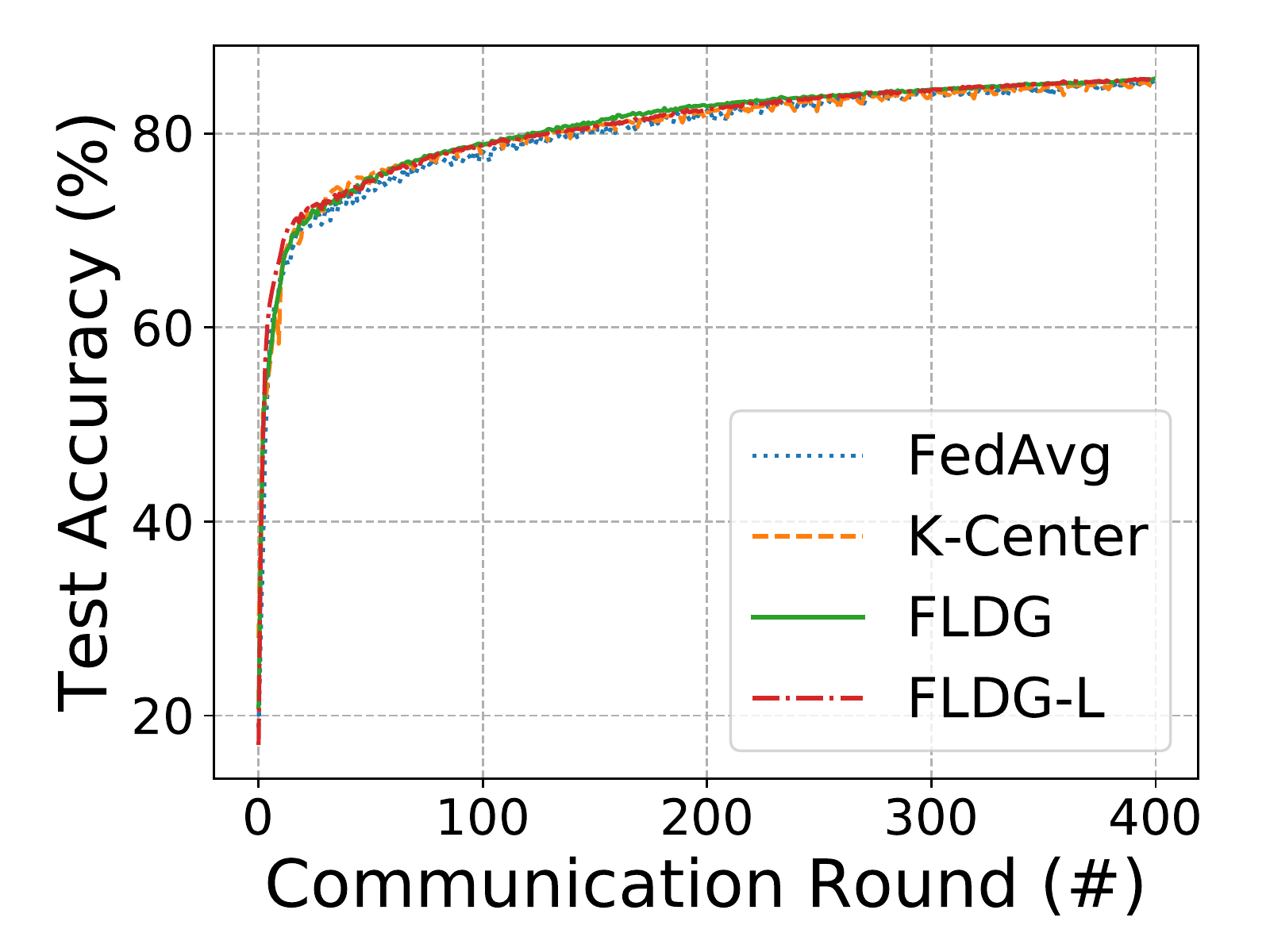}}
 \vspace{-0.1in}

\subfigure[Case 1 of CIFAR-10]{
\label{fig:ccase1} 
\includegraphics[width=0.23\linewidth]{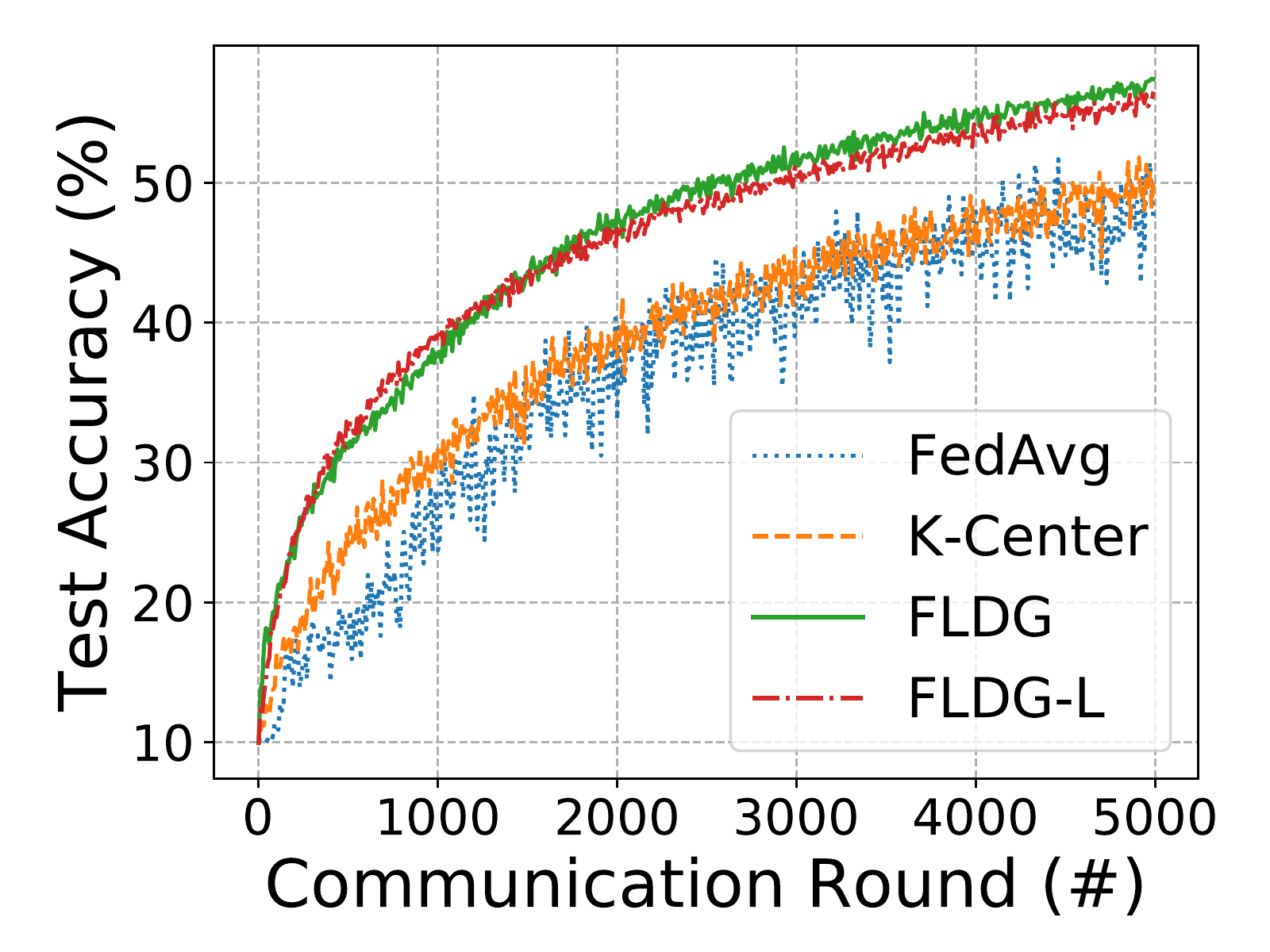}}
\subfigure[Case 2 of CIFAR-10]{
\label{fig:ccase2}
\includegraphics[width=0.23\linewidth]{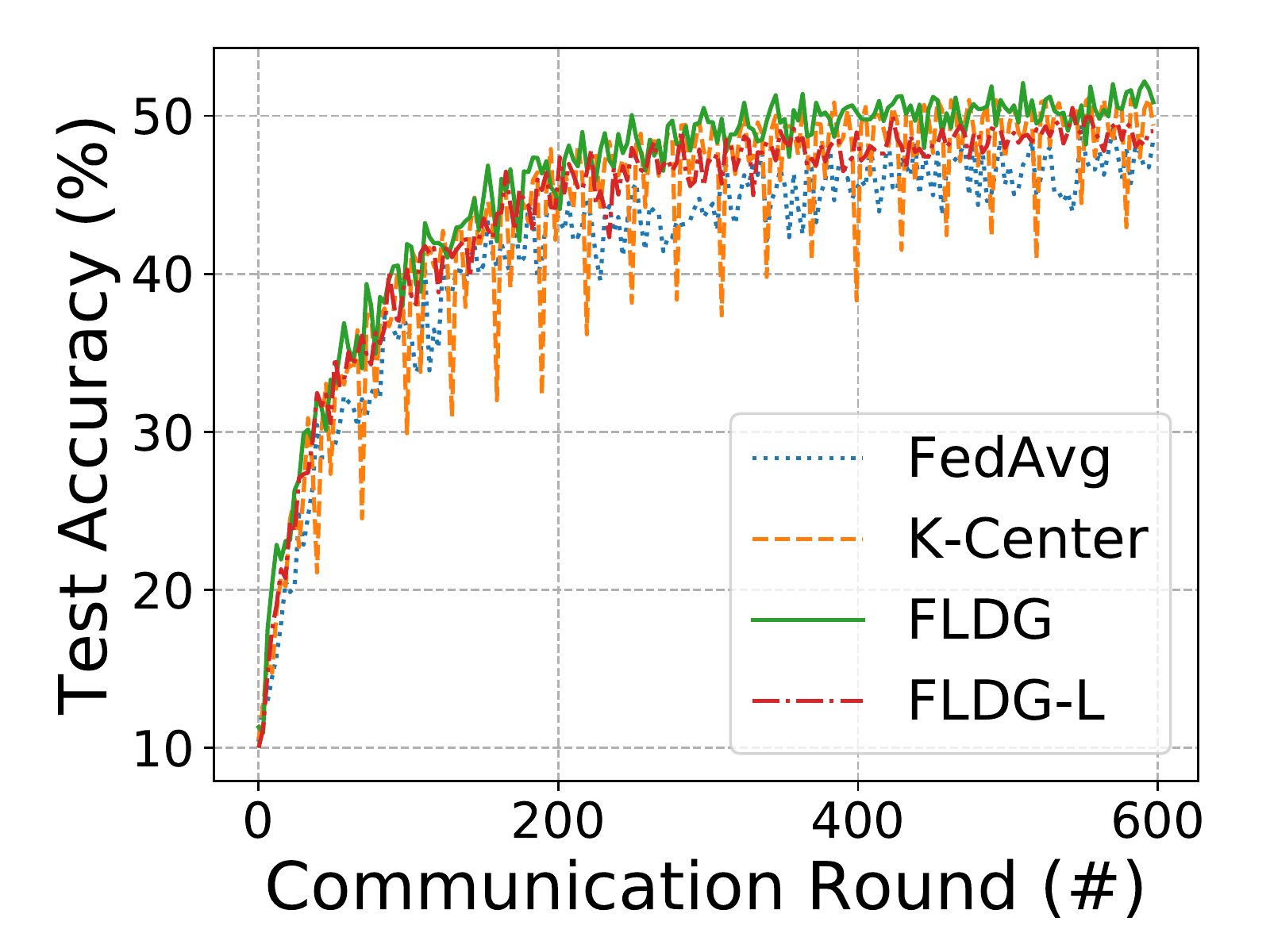}}
\subfigure[Case 3 of CIFAR-10]{
\label{fig:ccase3} 
\includegraphics[width=0.23\linewidth]{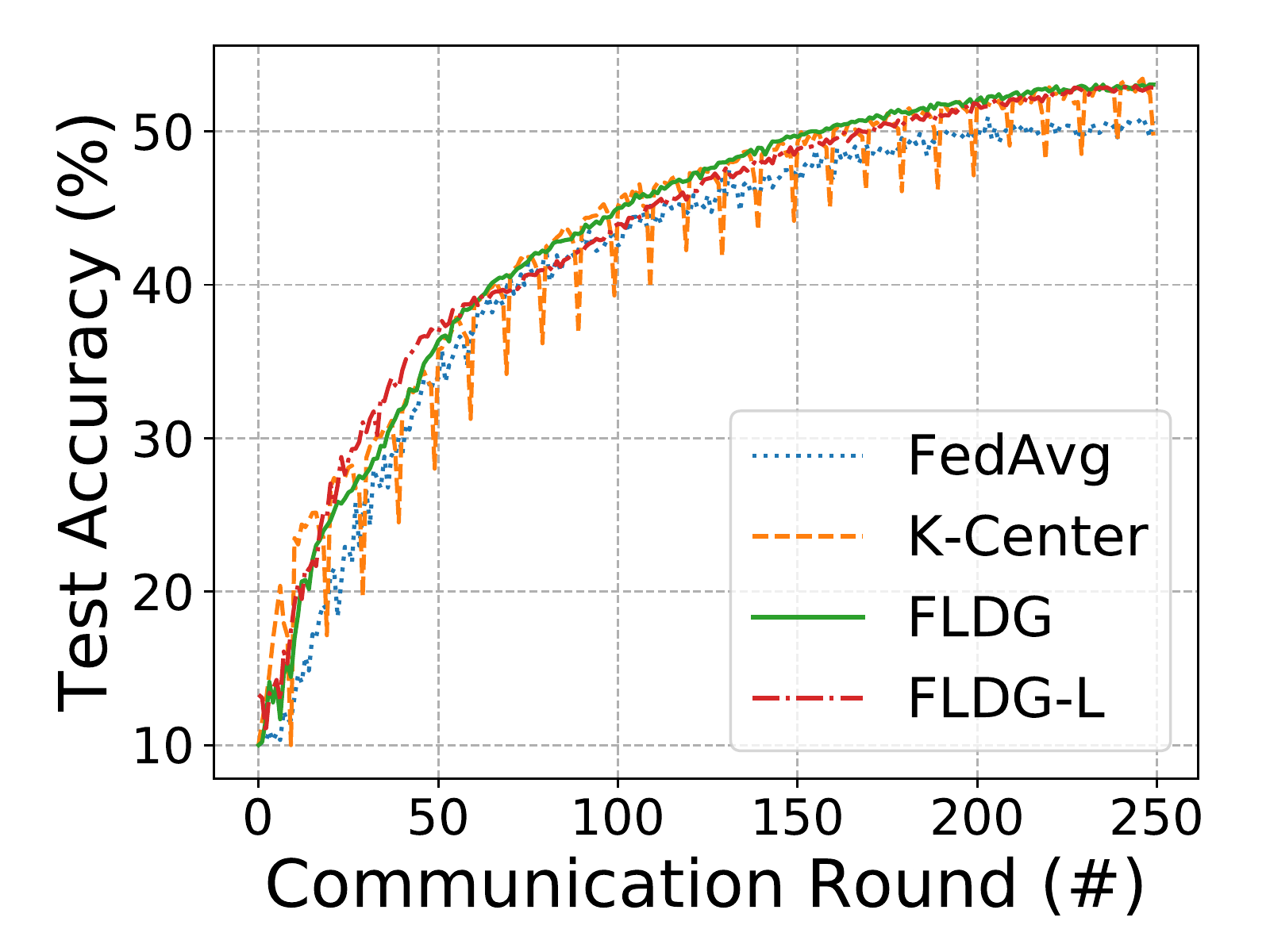}}
\subfigure[Case 4 of CIFAR-10]{
\label{fig:ccase4}
\includegraphics[width=0.23\linewidth]{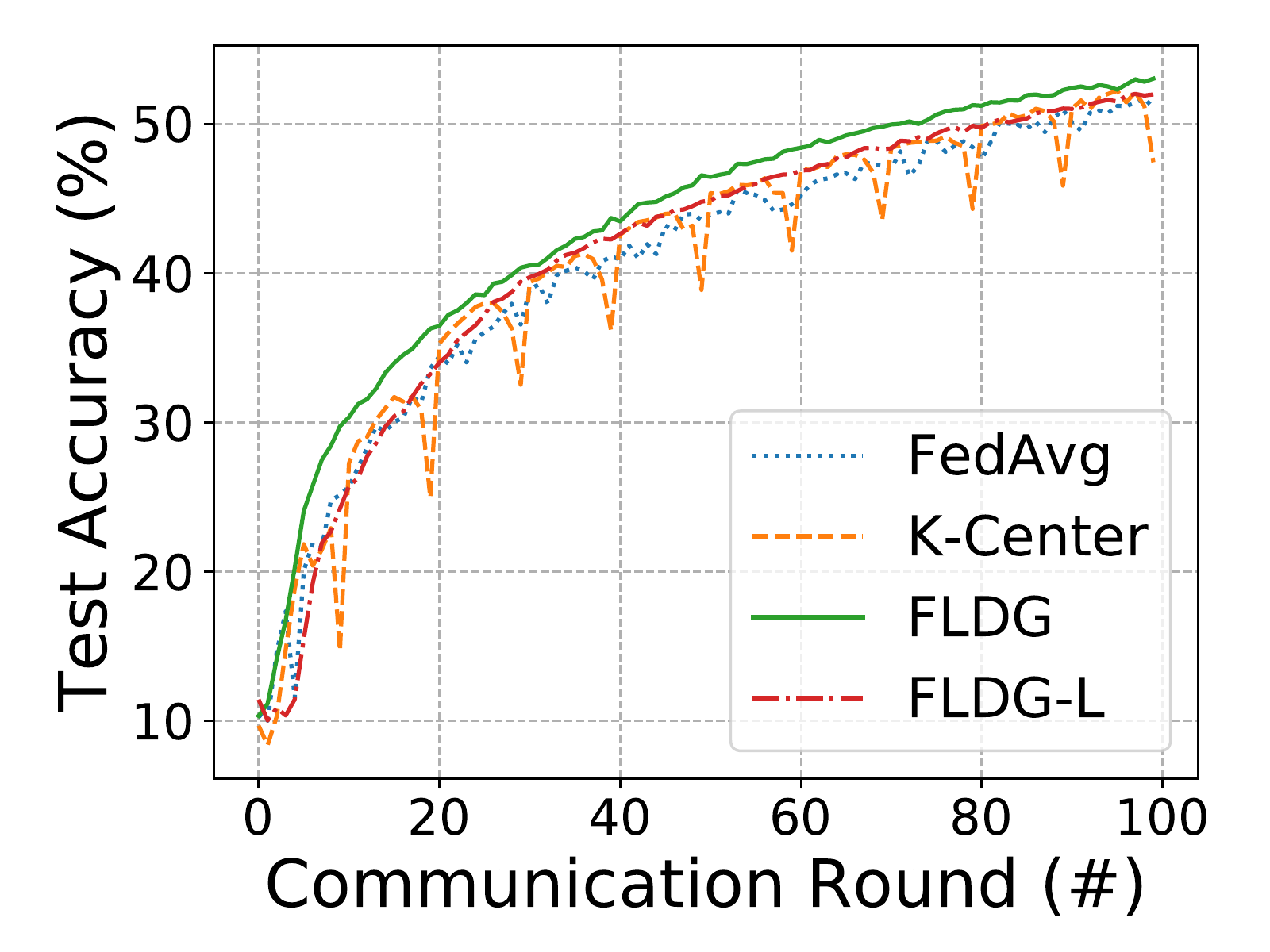}} 
\vspace{-0.1in}
\caption{Test Accuracy comparison of four approaches w.r.t. different non-IID cases}
\label{fig:mcase}
\end{figure*}

The output dimension (i.e., $h$) of p-stable LSH functions plays an important role in non-IID FL.
It determines both the accuracy of device grouping and communication overhead between the cloud and IoT devices.
Since FLDG does not use the LSH functions, we only investigate the impact of output dimension on FLDG-L here. 
Figure~\ref{fig:code} shows the inference performance of the three benchmarks using LSH functions with different output dimensions.
We tried to investigate the impact of output dimension from 1 to 1280 with a step of 1. Figure~\ref{fig:code} only shows four typical cases of them, since when the output dimension is larger than 4, the difference of test accuracy becomes small. 
From Figure~\ref{fig:code}, we can find that larger output dimension will lead to a better test accuracy. 
This is because larger output dimensions of LSH functions enable more accurate similarity comparisons for device grouping. 
When the output dimension $h$ is larger than $4$ and smaller than the feature map dimension $1280$, FLDG-L can achieve almost the best performance.  
Note that when $h=5$, the data privacy can be maximally protected without degrading test accuracy while the communication overhead is negligible. 
Therefore, we fixed $h = 5$ in the following experiments.

\subsection{Impact of The Number of Device Groups}

We also investigated the impact of different number of device groups on our FLDG and FLDG-L approaches. 
Figures~\ref{fig:num} and~\ref{fig:num2} show the trend of test accuracy along with the increment of device group number for the three benchmarks. 
From Figures~\ref{fig:num} and~\ref{fig:num2}, we can observe that when the number of device groups increases, both FLDG and FLDG-L converge faster and have higher test accuracy. 
For example, for the case of  FLDG-L on CIFAR-10 dataset in Figure~\ref{fig:flnum3}, when the number of groups increases from 5 to 50, the accuracy can be improved from 47.5\% to 54.8\% in the $800^{th}$ round.
In this case, to achieve 40\% test accuracy, FLDG-L with 5 groups needs around  200 rounds, while FLDG-L with 50 groups requires less than 40 rounds. 
This is mainly because more device groups can make IoT devices more finely clustered, thus making the model update closer to the optimal direction in each round of aggregation.
Moreover, both FLDG and FLDG-L take one IoT device from each group to participate in each round of gradient aggregation. 
Consequently, more groups lead to more IoT devices participating in the gradient aggregation, which in turn reduces the number of training rounds to achieve expected test accuracy.

\subsection{Performance Comparison with State-of-the-Art}\label{sec:sota}

To show the effectiveness of  FLDG and FLDG-L for non-IID scenarios, we used the state-of-the-art approaches, i.e., vanilla FL (FedAvg)~\cite{communication} and K-Center~\cite{active} as baselines for the comparison. 
Figure~\ref{fig:mcase} presents the test accuracy results of the four approaches for the four non-IID cases defined in Table~\ref{tab:dist}.
From Figure~\ref{fig:mcase} we can find that for all the four non-IID cases, FLDG and FLDG-L greatly outperform the two baselines in terms of convergence rate and test accuracy.
For example, in Figure~\ref{fig:mcase1} FLDG achieves better accuracy than FedAvg and  K-Center by 13.2\% and 46.3\% in the $100^{th}$ round, respectively. 
Similarly, in the same round FLDG-L  outperforms the two baselines by 11.8\% and 44.6\%, respectively.
For all the cases, the performance of FLDG-L is similar to that of FLDG. 
In other words, although FLDG-L is more secure than FLDG, the performance of FLDG-L can still be guaranteed with negligible LSH hashing overhead.
 The reason why our approaches are superior is mainly because our device grouping method can reduce the disadvantages of weight divergence during the FL training process.
Therefore, the model optimization direction of each aggregation round becomes more consistent with the optimization direction of the overall training. 
Note that in most cases the performance of K-Center fluctuates greatly.
This is because here K-Center conducts device clustering based on the model similarity in every 10 rounds, where the first round has the best performance while the last round gives the worst performance.
Moreover, we can find that for each benchmark, the four methods in Case 1 have the largest difference, but in Case 4 the four methods have similar performance. 
This is because that Case 1 has the most skewed non-IID device data while Case 4 has the 
least skewed non-IID device data. 
In other words, Case 4 tends to be more IID.

\section{Conclusion}\label{sec:conclusion}

Although FL techniques have been increasingly investigated in IoT domains to enable collaborative learning, they still suffer from low inference performance in non-IID scenarios.
To address this problem, this paper proposed two novel FL methods (i.e., FLDG and FLDG-L) based on device grouping, which can significantly mitigate the disadvantages of weight divergence during the FL model training.
By grouping IoT devices based on the feature maps of their raw data using the K-Means algorithm and selecting one IoT device from each group for the model aggregation per round, the proposed FLDG method can effectively stabilize the convergence of the global model. 
To further reduce the risk of privacy exposure of feature maps on the cloud, FLDG-L encodes extracted feature maps using the LSH function and sends their hash values to the cloud for device grouping.
Experimental results demonstrate that our approaches can effectively speed up the global model convergence rate while achieving higher accuracy.

in

\end{document}